\documentclass[journal]{IEEEtran}
\usepackage{hyperref}
\usepackage{url}
\usepackage{ragged2e}
\usepackage{algorithm} % algorithmic
\usepackage{algpseudocode}
\usepackage{mathtools}
\usepackage{subfigure}
\usepackage{amsmath,cases}
\usepackage{subeqnarray}
\usepackage{amsmath,amsfonts,bm}
\usepackage{caption}
\usepackage{xcolor}
\usepackage{enumitem}
\usepackage{kantlipsum}
\allowdisplaybreaks

\begin{document}

\title{Long-term Fairness For Real-time Decision Making: \\A Constrained Online Optimization Approach}

\author{Ruijie~Du,
        Deepan Muthirayan,
        Pramod P. Khargonekar,% \IEEEmembership{Life~Fellow,~IEEE,}
        and~Yanning~Shen%, \IEEEmembership{Member,~IEEE}
        %and~Jane~Doe,~\IEEEmembership{Life~Fellow,~IEEE}% <-this % stops a space
%\thanks{Thanks .............}% <-this % stops a space
%\thanks{..........}% <-this % stops a space
}

%\markboth{IEEE TRANSACTIONS ON NEURAL NETWORKS AND LEARNING SYSTEMS} %Journal of \LaTeX\ Class Files,~Vol.~14, No.~8, August~2015%
\makeatletter
\newcommand{\mathleft}{\@fleqntrue\@mathmargin0pt}
\newcommand{\mathcenter}{\@fleqnfalse}
\makeatother
\newcommand\shen[1]{{\color{blue}{\textbf{[[Shen: {\em#1}]]}}}}

\newcommand{\rd}[1]{\textcolor{red}{#1}}

\newcommand{\ys}[1]{\textcolor{blue}{#1}}

\newcommand{\dm}[1]{\textcolor{teal}{#1}}
\newcommand{\purple}[1]{\textcolor{purple}{#1}}

% Reference to an equation, lower-case.
\def\eqref#1{equation~\ref{#1}}
% Reference to an equation, upper case
\def\Eqref#1{Equation~\ref{#1}}
% Figure reference, lower-case.
\def\figref#1{figure~\ref{#1}}
% Figure reference, capital. For start of sentence
\def\Figref#1{Figure~\ref{#1}}
\def\twofigref#1#2{figures \ref{#1} and \ref{#2}}
\def\quadfigref#1#2#3#4{figures \ref{#1}, \ref{#2}, \ref{#3} and \ref{#4}}

\maketitle

\begin{abstract}
Machine learning (ML) has demonstrated remarkable capabilities across many real-world systems, from predictive modeling to intelligent automation.
However, the widespread integration of machine learning also makes it necessary to ensure machine learning-driven decision-making systems do not violate ethical principles and values of society in which they operate.
As ML-driven decisions proliferate, particularly in cases involving sensitive attributes such as gender, race, and age, to name a few, the need for equity and impartiality has emerged as a fundamental concern.
In situations demanding real-time decision-making, fairness objectives become more nuanced and complex: instantaneous fairness to ensure equity in every time slot, and long-term fairness to ensure fairness over a period of time.
There is a growing awareness that real-world systems that
operate over long periods and require fairness over different timelines. 
%To extend fairness into dynamic scenarios, recent research delves into the long-term fairness implications across various applications. 
%Consequently, understanding long-term fairness becomes increasingly crucial due to its adaptability and applicability.
However, existing approaches mainly address dynamic costs with time-invariant fairness constraints, often disregarding the challenges posed by time-varying fairness constraints.
To bridge this gap, this work introduces a framework for ensuring long-term fairness within dynamic decision-making systems characterized by time-varying fairness constraints. {We formulate the decision problem with fairness constraints over a period as a constrained online optimization problem.
A novel online algorithm, named LoTFair, is presented that solves the problem `on the fly’. We prove that LoTFair can make overall fairness violations negligible while maintaining the performance over the long run}.
\end{abstract}


% \fi
%\IEEEpeerreviewmaketitle

\section{Introduction}

In recent years, machine learning (ML) systems have demonstrated remarkable capabilities across a range of real-world systems, from predictive modeling to intelligent automation.
However, it has been shown that the sole pursuit of learning performance may lead to unfair results toward underrepresented groups. % It has been shown that the sole pursuit of learning performance may lead to results that are unfair towards the underrepresented groups. 
For example, applications to loan approvals \cite{fair_loan} and college admissions \cite{fair_college} highlight the importance of fairness to avoid discrimination. % For instance, in settings such as loan approvals \cite{fair_loan} or college admissions \cite{fair_college}, fairness must be carefully taken into account in order to ensure the absence of discrimination.
The widespread adoption of ML also raises the need to understand the social and ethical responsibility of ML-driven decision-making systems. 
Specifically, ensuring that ML-driven decisions are equitable and unbiased, especially when involving sensitive attributes such as gender, race, and age, has emerged as a fundamental and important problem.

%The significance of fairness-aware machine learning applies to a diverse array of real-world systems, each presenting unique challenges and opportunities. These systems cover areas as diverse as transportation networks \cite{Yan_Howe_2020}, wireless networks \cite{long-fair-dynamic-resource-allocation}, finance\cite{fair_loan}, \cite{fair_college}, and energy management \cite{guo2020optimal}, \cite{guo2022optimisation}.
Fairness-aware machine learning is important in a diverse array of real-world systems, including  transportation networks \cite{Yan_Howe_2020}, wireless networks \cite{long-fair-dynamic-resource-allocation}, finance\cite{fair_loan}, \cite{fair_college}, and energy management systems \cite{guo2020optimal}, \cite{guo2022optimisation}.
%In applications that require real-time decision-making, a multitude of fairness objectives have been used, including the instantaneous fairness objective, which aims to ensure fairness at every time slot, and the long-term fairness objective which aims to ensure fairness over a time horizon \cite{fairness-is-not-static}. 
{In applications that require real-time decision-making, two fairness concepts naturally arise: instantaneous fairness which aims to ensure fairness at every time slot, and long-term fairness which maintains fairness over a time horizon. Various fairness objectives are formulated from the aspects of instantaneous fairness and long-term fairness,}
while most existing studies concentrate on instantaneous fairness \cite{corbett2017algorithmic}. There is a growing realization that real-world systems operating over a period require fairness to be satisfied over an extended timeline \cite{zhang2021fairness}. 
%This perspective acknowledges that issues of fairness may persist over time . 
But achieving long-term fairness is challenging because long-term dynamics are hard to assess and the extended nature of the timeline does not align with the traditional supervised ML framework that uses fixed data sets \cite{fairness-is-not-static}, \cite{hu2022achieving}.
%
%To extend fairness to dynamic settings, 
Some recent works have explored the long-term fairness in certain applications 
\cite{fair_loan},\cite{fair_college},\cite{fairness-is-not-static}, \cite{hu2022achieving},\cite{weber2022enforcing}, \cite{zhang2021fairness}, \cite{yin2023long}. 
%{To achieve long-term fairness in different systems, most approaches focus on the problems with dynamic costs but time-invariant fairness constraints and do not consider the problem with time-varying fairness constraints. } % \shen{so many but in the previous sentence, re-write it.}
We observe that most of these approaches focus on dynamic costs and time-invariant fairness constraints.
However, the fairness constraint might vary with time due to the system dynamics and evolution of decisions \cite{zhang2021fairness}, as observed in many real world systems \cite{Yan_Howe_2020},\cite{P2P-energy-transaction-mechanisms}.
%There is a growing need to understand long-term fairness due to its flexibility and applicability to real-world systems \cite{Yan_Howe_2020},\cite{P2P-energy-transaction-mechanisms}.

Motivated by these considerations, in this work, we introduce a novel framework for long-term fairness of decision-making in dynamic systems with time-varying fairness constraints.
%In this work, we investigate the problem of long-term fairness for decision-making, formulated as a constrained online optimization problem in which the overall utility is optimized in the presence of the long-term fairness constraint. 
{The decision problem with long-term fairness is formulated as a constrained online optimization problem.}
We present a novel online algorithm and prove that the algorithm achieves sub-linear dynamic regret and sub-linear accumulated unfairness under certain assumptions. Experimental results on the classification problem and the energy management problem in a time-varying peer-to-peer electricity market are presented to validate the effectiveness of the proposed framework.

The paper is organized as follows: \ref{prob_formulation} introduces the problem formulation of the decision-making system. Section \ref{Methodology} proposes the online algorithm. Section \ref{bi-classify} introduces the online logistic classification and Section \ref{P2P} introduces the energy management problem in the P2P electricity market. Section \ref{analysis} introduces performance metrics and provides proof that the algorithm can achieve sub-linear dynamic regret and sub-linear accumulated unfairness under convexity assumptions. The numerical tests are presented in Section \ref{all-experiment} to evaluate the performance of the proposed method.

\section{Problem Formulation} \label{prob_formulation}
{Machine learning has been increasingly incorporated within decision-making systems, especially in ecosystems and cyber-physical systems. Such increasing popularity has led to a growing need for ensuring the fairness of ML-based decision-making systems \cite{robert2020icis}. Most approaches for fairness focus on instantaneous fairness, while ignoring long-term effects. 
%\shen{add motivations of while long-term fairness is important}
However, in applications requiring real-time decision-making, more general long-term fairness criteria need to be developed, and extending the instantaneous approach directly to the long-term setting may not achieve long-term fairness.}% long-term fairness needs to be taken into account, we formulate it as an online optimization problem with task-specific long-term fairness constraints.

{The problem we are solving is a real-time decision problem, where the decisions have to be adapted online to the changing dynamics of the system.
In this online setting, our focus is to ensure the sequence of decisions made online is fair in the long term.}
%Consider that the time is discrete and indexed by $t$. 
At each time slot indexed by $t$, an online learner learns to
%\shen{avoid using the word controller, we are writing a learning paper} 
makes a decision $\mathbf{x}_t$, 
% \shen{sometimes you call $\mathbf{x}_t$ decision, sometimes you call it action...}
and incurs a loss function $f_t(\cdot)$: $\mathbb{R}^n \rightarrow \mathbb{R}$, where the decision $\mathbf{x}_t\in \mathcal{X}_t \subseteq \mathcal{X}\in \mathbb{R}^n$, with $\mathcal{X}$ known and fixed. 
{Note that $f_t(\cdot)$ and $\mathcal{X}_t$ are driven by the unknown dynamics in various applications.}
In addition, there are two groups $\mathcal{D}_z$, with $z\in\{0,1\}$ denoting the sensitive attribute of individuals in each group. 
%\shen{at this point, it is better not to mention anything convex, as it is a ML paper, and almost nothing is convex. You can introduce convexity at theoretical analysis}
%The learner incurs a loss $f_t(\mathbf{x}_t)$. 
The goal is to ensure fairness by achieving equitable outcomes for different sensitive groups.

%Specifically, %penalty \shen{but why is this a penalty function? it is never used as a penalty function in this manuscript...}
A time-varying function $g_t(\cdot):\mathbb{R}^n \rightarrow \mathbb{R}$ of the decision $\mathbf{x}_t$ can characterize the bias of the model, %, which is driven by the unknown dynamics in various applications and computed from specific fairness metrics. 
%\shen{why? give an example} \\
% In $\kappa$-fairness scenarios
%\shen{what is $\kappa$-fairness? not defined yet, if it was defined in other literature, add citation}, 
\begin{align*}
    g_t(\mathbf{x}_t) \coloneqq \Omega(\mathcal{D}_0,\mathbf{x}_t) - \Omega(\mathcal{D}_1,\mathbf{x}_t)
\end{align*}
where $\Omega(\mathcal{D}_0,\mathbf{x}_t)$ denotes the performance metric reflecting the performance on group $\mathcal{D}_0$ with the decision $\mathbf{x}_t$.
{Since $\mathcal{X}_t$ is not a fixed set, the time-varying $g_t(\mathbf{x}_t)$ can vary arbitrarily or even adversarially from slot to slot.}
Ideally, if $g_t(\cdot)=0$ at every $t$, which implies that the decisions are perfectly fair at each time slot, the learner can also guarantee fairness in the long run, $\sum_{t=1}^T g_t(\mathbf{x}_t) =0$. However, achieving $g_t(\cdot)=0$ is very restrictive and sometimes infeasible for the learner. Most instantaneous approaches relax the equality constraint, $g_t(\cdot)=0$, to a general inequality constraint $g^{\kappa}_t(\mathbf{x}_t) \leq 0$ where
\begin{align*}
    g_t^{\kappa}(\mathbf{x}_t) \coloneqq \big(\Omega(\mathcal{D}_0,\mathbf{x}_t) - \Omega(\mathcal{D}_1,\mathbf{x}_t)\big)^2 - \kappa. %\leq 0
\end{align*}
The relaxed instantaneous constraint, $g^{\kappa}_t(\mathbf{x}_t) \leq 0$, bounds the learner to make the violation caused by the decision within an acceptable range, controlled by $\kappa$, at each time slot $t$. 
{
The problem can be formulated as
\begin{align}
    &\min_{\mathbf{x}_t\in \mathcal{X}_t, \forall t \in \{1,\cdots,T \}}  \sum_{t=1}^T f_t(\mathbf{x}_t) 
    \qquad \text{ s.t. } [\mathbf{x}_t]_{t=1}^T \in \mathcal{G}_T^{I}  % , \qquad \forall t \in \{1,\cdots,T \}
    \label{instantaneous_general}
\end{align}
% \shen{the constraint is not correct, the left hand side is a set, if it is a subset of right hand side, then should use $\subset$, if it is element wise, the should be each xt is a element of Gt}
where $T$ is the time horizon, $[\mathbf{x}_t]_{t=1}^T = [\mathbf{x}_1, \mathbf{x}_2, \cdots, \mathbf{x}_T]$ denotes the {decision trajectory}, which contains the sequence of instantaneous decisions over the time horizon $T$, where $\mathcal{G}_T^{I}$ denotes the relaxed instantaneous fairness constraints over the time horizon which is defined as
\begin{align} \label{instantaneous-set}
    \mathcal{G}_T^{I} = \left\{ [\mathbf{x}_t ]_{t=1}^T \bigg|  g_t^{\kappa}(\mathbf{x}_t) \leq 0, \forall t \in \{1,\cdots,T \}  \right\}.
\end{align}}

While the relaxation can avoid infeasibility, it can lead to violation of fairness in the long run: aggregation of $g_t(\mathbf{x}_t)$ over $T$ can grow linearly, $\sum_{t=1}^T g_t(\mathbf{x}_t) = \mathcal{O}(\kappa T)$, which trivially does not satisfy fairness. To address this problem, we introduce long-term fairness, which allows fairness violations at individual time slots, while ensuring fairness in the long run: $$ \lim_{T \rightarrow \infty} \frac{1}{T}\sum_{t=1}^T g_t(\cdot) = 0.$$
{
{The long-term fairness constraint can be formulated as %To address this problem, the long-term constraint is defined as
}
\begin{align} \label{long-term-set}
     \mathcal{G}_T^{L} = \left\{ [\mathbf{x}_t ]_{t=1}^T \Bigg| \sum_{t=1}^T g_t(\mathbf{x}_t) = 0 \right\},
\end{align}
where %$g_t(\cdot)$ defined as 
 % \shen{I am confused, but our long-term also cannot obtain equal sign?}
\begin{align*}
    g_t(\mathbf{x}_t) \coloneqq \Omega(\mathcal{D}_0,\mathbf{x}_t) - \Omega(\mathcal{D}_1,\mathbf{x}_t).
\end{align*}
In the long-term approach, we adopt the long-term formulation as the fairness constraint, and our goal is to find a sequence of decisions %\shen{soultion? you mean decision?} 
that minimizes the aggregate loss and ensures the fairness constraints are satisfied in the long run:
\begin{align}
    & \min_{\mathbf{x}_t\in \mathcal{X}_t, \forall t \in \{1,\cdots,T \}}  \sum_{t=1}^T f_t(\mathbf{x}_t)
    \qquad \text{ s.t. } [\mathbf{x}_t ]_{t=1}^T \in \mathcal{G}_T^{L}.
    \label{long_general}
\end{align}
}
{Note that enforcing \eqref{long-term-set} is different from restricting $\mathbf{x}_t$ in every time slot to satisfy $g_t(\mathbf{x}_t)= 0$. The learner is tolerable of instantaneous fairness violation, e.g., $g_t(\mathbf{x}_t)\leq 0$ or $g_{t}(\mathbf{x}_{t}) \geq 0$, but will ensure the overall violation will become negligible over the long run.}
%{where $\Omega(\mathcal{D}_0,\mathbf{x}_t)$ denotes the performance metric \rd{reflecting the performance on group $\mathcal{D}_0$ with the decision $\mathbf{x}_t$}.
%It returns a positive value only when the difference between the fairness metrics of two different groups exceeds the threshold $\kappa$, namely $\kappa$-fairness.
%}

%An instantaneous time-varying constraint $g^{\kappa}_t(\mathbf{x}_t) \leq 0$ can be introduced to bound the disparity caused by the decision $\mathbf{x}_t$ between two groups by $\kappa>0$.
%\shen{why? needs more intuition}
%\blue{[an example of equal opportunity(EO) -> demographic parity(DP)]}
%race and gender. 
%\shen{ I guess you mean male or female, white or black, but this sentence is ambiguous, it makes as if the two groups are race and gender. As there are more than two races or genders, be clear and to the point}
% ,$Y$ denotes the 
% \blue{Although $g_t^{\kappa}(\mathbf{x}_t)$ can be revealed before the decisions are made, it might lead to huge increases in loss function or even cannot be satisfied if the constraint $g_t^{\kappa} \leq 0$ is very strict($\kappa$ is close to 0).} \shen{Did not follow this}
%Therefore, the goal is to find online solutions that minimize the aggregate loss and ensure the fairness constraints are satisfied.
For example, in ecosystems such as bank loans and college admissions, the learner decides whether or not to approve a stream of applications \cite{fairness-is-not-static}. In this case, $\mathbf{x}_t$ denotes the set of binary decisions for individuals at $t$.
The sensitive attributes can denote the properties of individuals such as male/female and white/non-white.
%, \shen{what is the utility function in this example?} 
In the instantaneous approach, the learner minimizes the cost function subject to the instantaneous demographic parity (DP) constraints at every time slot \cite{dwork2012fairness}, \cite{yin2023long}. %subjects to instantaneous fairness constraints at every time slot: 
% The instantaneous fairness constraint as $g^{\kappa}_t(\mathbf{x}_t) \leq 0$, which bound the disparity between two groups caused by the decision $\mathbf{x}_t$.
%Specifically for the college example, 
At each time slot $t$, decisions should be made such that individuals from different sensitive groups have similar probabilities of being approved,
%\shen{check this sentence, there seems to be some grammar mistakes}: 
{i.e., $g_t^{\kappa}(\mathbf{x}_t) \coloneqq \left( P(q_t(d\in \mathcal{D}_0,\mathbf{x}_t)=1) - P(q_t(d\in \mathcal{D}_1,\mathbf{x}_t)=1)\right)^2-\kappa
\leq0$, where $q_t(d\in \mathcal{D}_0,\mathbf{x}_t)$ denotes the predicted outcome of an individual belonging to sensitive group $\mathcal{D}_0$ based on current decision $\mathbf{x}_t$, $\Omega(\mathcal{D}_i,\mathbf{x}_t) = P(q_t(d\in \mathcal{D}_i,\mathbf{x}_t)=1)$ denotes the probability of being approved for individuals $\mathcal{D}_i$}
%\shen{what does it mean by predicted as the positive class on group S0?Do you mean xt belongs to S0?}
and $\kappa \in [0,1]$.
%\shen{but this is still not correct, the right hand side has something to do with T, but the left-hand side does not}
%\begin{align*}
%    \min_{\mathbf{x}_t\in \mathcal{X}}  f_t(\mathbf{x}_t) 
%    \qquad \text{ s.t. } g_t^{\kappa}(\mathbf{x}_t)\leq 0, \qquad \forall t \in \{1,\cdots,T \}
%\end{align*}
%defined as \blue{}\shen{?}\shen{since it is a new definition, maybe better to define the kappa fairness as an equation and refer to it later}.
%To consider long-term fairness, one intuitive attempt would be aggregating the instantaneous fairness violation as the constraint, $\sum_{t=1}^T g_t^{\kappa}(\mathbf{x}_t) \leq 0$. 
%\rd{However, instantaneous fairness does not take account of long-term effects, and} such direct aggregation cannot ensure fairness in the long term. In the college admission example, the constraint $\sum_{t=1}^T g_t^{\kappa}(\mathbf{x}_t) \leq 0$ with $g_t^{\kappa}(\mathbf{x}_t) \coloneqq \left(\Omega(\mathcal{D}_0,\mathbf{x}_t) - \Omega(\mathcal{D}_1,\mathbf{x}_t)\right)^2-\kappa =\left(P(q_t(d\in \mathcal{D}_0,\mathbf{x}_t)=1) - P(q_t(d\in \mathcal{D}_1,\mathbf{x}_t)=1)\right)^2-\kappa$ cannot ensure that the average DP goes to zero over time:
In the long-term approach, the long-term fairness is also subject to DP constraint, $g_t(\mathbf{x}_t) \coloneqq \Omega(\mathcal{D}_0,\mathbf{x}_t) - \Omega(\mathcal{D}_1,\mathbf{x}_t) = P(q_t(d\in \mathcal{D}_0,\mathbf{x}_t)=1) - P(q_t(d\in \mathcal{D}_1,\mathbf{x}_t)=1)$.

\section{Method} \label{Methodology}
%\blue{
%If the action $\mathbf{x}_t$ is selected from a convex set $\mathcal{X}\in \mathrm{R}^n$, and the loss function $f_t(\cdot)$ is convex, the online optimization problem of \eqref{objective} is an online convex optimization(OCO) problem.}
%To solve the optimization problem online with the long-term fairness constraint in (\ref{long_general})
%\shen{but why is \eqref{objective} an OCO problem? It is never shown that the constraints are convex sets?} 
%and achieving sub-linear dynamic regret and fairness on average, 
%we introduce our method in this section and analyze its performance in Section \ref{Method}.
%The present section proposes an online algorithm that makes decision $\mathbf{x}_t$ on the fly.
In this section we present the online algorithm for the problem (\ref{long_general}).
%The \eqref{long_general} can be rewritten as
%\begin{align} %\shen{ the more strict and more lose is not clear} \label{1line-objective}
%    \min_{\mathbf{x}_t\in \mathcal{X}_t, \forall t}  \sum_{t=1}^T f_t(\mathbf{x}_t)
%    \qquad \text{s.t. } ~~ \sum_{t=1}^T g_t(\mathbf{x}_t) = 0 
%\end{align}
%\shen{is the constraint the same as (2), then we should use whatever constraint set you have defined.}
Note that long-term fairness constraint $[\mathbf{x}_t ]_{t=1}^T \in \mathcal{G}_T^{L}$ (\ref{long-term-set}) can be equivalently written as two inequality constraints. Therefore, problem (\ref{long_general}) is equivalent to %\shen{but why is the following problem online? it seems batch}
\begin{align} \label{objective}
        \min_{\mathbf{x}_t\in \mathcal{X}_t, \forall t}  \sum_{t=1}^T f_t(\mathbf{x}_t) 
   ~ \text{ s.t.} \sum_{t=1}^T g_t(\mathbf{x}_t)\leq 0,  \sum_{t=1}^T \!\!-g_t(\mathbf{x}_t)\leq 0 .
\end{align}
Hence, the instantaneous problem at time $t$ can be written as
\begin{align} \label{one-frame}
        &\min_{\mathbf{x}_t\in \mathcal{X}_t} f_t(\mathbf{x}_t)
   \qquad \text{ s.t. } ~~   g_t(\mathbf{x}_t)\leq 0, ~~  -g_t(\mathbf{x}_t)\leq 0 .
\end{align}

Let $\lambda_{1,t},\lambda_{2,t} \geq 0$ denote the Lagrange multipliers associated with the time-varying constraints. The online Lagrangian of \eqref{one-frame} can be written as
\begin{align} \label{online_lagrangian}
    \mathcal{L}_t(\mathbf{x}_t,\lambda_{1,t},\lambda_{2,t}) \!=\!  f_{t}(\mathbf{x}_t) + \lambda_{1,t} g_t(\mathbf{x}_t) + \lambda_{2,t} (-g_t(\mathbf{x}_t)).
\end{align}
%For the online Lagrangian \eqref{online_lagrangian}% \shen{what do you mean by solving lagrangian? }
%, we utilize the modified online saddle point (MOSP) approach: takes a descent step in the primal domain and a dual ascent step at each time slot in a Gauss-Seidel manner \cite{chen2017online}. 
Given the previous primal iterate $\mathbf{x}_{t-1}$ and current dual iterates $\lambda_{1,t},\lambda_{2,t}$ at each time slot $t$, %we want take a descent step as Gauss-Seidel method. 
instead of taking the exact gradient step, the current decision $\mathbf{x}_t$ can be obtained by solving the following optimization problem:
\begin{align} \label{primal_update}
    \mathbf{x}_t = \min_{\mathbf{x} \in \mathcal{X}_t} &\nabla f_{t-1}(\mathbf{x}_{t-1})^\top(\mathbf{x}-\mathbf{x}_{t-1}) + \lambda_{1,t} g_{t-1}(\mathbf{x}) \nonumber \\
    & + \lambda_{2,t} (-g_{t-1}(\mathbf{x})) + \frac{||\mathbf{x}-\mathbf{x}_{t-1} ||^2}{2\alpha} \notag \\
    = \min_{\mathbf{x}\in \mathcal{X}_t} &\nabla f_{t-1}(\mathbf{x}_{t-1})^\top(\mathbf{x}-\mathbf{x}_{t-1}) \nonumber \\
    & + \bm{\lambda}_{t}^\top \mathbf{\bar{g}}_{t-1}(\mathbf{x}) + \frac{||\mathbf{x}-\mathbf{x}_{t-1} ||^2}{2\alpha}
\end{align}
where $\alpha>0$ is the primal step size, $\bm{\lambda}_t \coloneqq [\lambda_{1,t}, \lambda_{2,t}]^\top$, and $\mathbf{\bar{g}}_{t}(\mathbf{x}) \coloneqq [g_t(\mathbf{x}),-g_t(\mathbf{x})]^\top$. %$\nabla f_{t-1}(\mathbf{x}_{t-1})$ is the gradient of primal objective $f_{t-1}(\mathbf{x})$ at $\mathbf{x}_{t-1}$. 
\Eqref{primal_update} tries to minimize the fair constraint violation by taking a modified descent step. %$\frac{||\mathbf{x}-\mathbf{x}_{t-1} ||^2}{2\alpha}$ is a proximal term.
After the decision $\mathbf{x}_t$ is made, $f_t(\mathbf{x}_t)$ and $g_t(\mathbf{x}_t)$ are revealed. Then the dual variables can be updated as % \shen{why?need more details here}
\begin{align}
    \lambda_{1,t+1} &= \max(0,\lambda_{1,t} + \mu \nabla_{\lambda_{1,t}} \mathcal{L}_t(\mathbf{x}_t,\lambda_{1,t},\lambda_{2,t})) \nonumber \\
    &=\max(0,\lambda_{1,t} + \mu g_t(\mathbf{x}_t)) \label{1_update}\\
    \lambda_{2,t+1} &= \max(0,\lambda_{1,t} + \mu \nabla_{\lambda_{2,t}} \mathcal{L}_t(\mathbf{x}_t,\lambda_{1,t},\lambda_{2,t})) \nonumber \\
    &= \max(0,\lambda_{2,t} - \mu g_t(\mathbf{x}_t)) \label{2_update}
\end{align}
%$\nabla_{\lambda_1} \mathcal{L}_t(\mathbf{x},\lambda_1,\lambda_2) = g_t(\mathbf{x}_t), \nabla_{\lambda_2} \mathcal{L}_t(\mathbf{x},\lambda_1,\lambda_2) = -g_t(\mathbf{x}_t)$ are the gradients of \eqref{online_lagrangian} with respect to $\lambda_1,\lambda_2$ at $\bm{\lambda} = \bm{\lambda}_t$.
where $\mu>0$ is the dual step size, $\lambda_{1,t}, \lambda_{2,t}\geq 0$. $\nabla_{\lambda_{1,t}} \mathcal{L}_t(\mathbf{x}_t,\lambda_{1,t},\lambda_{2,t}))=g_t(\mathbf{x}_t)$ and $\nabla_{\lambda_{2,t}} \mathcal{L}_t(\mathbf{x}_t,\lambda_{1,t},\lambda_{2,t})=-g_t(\mathbf{x}_t)$ are the gradients of \eqref{online_lagrangian} with respect to $\lambda_{1,t}$ and $\lambda_{2,t}$. 
Note that updating $\bm{\lambda}_t$ and making decision $\mathbf{x}_t$ %\shen{do you mean xt? check glboally}
only require the previous time slot's information, $f_{t-1}(\mathbf{x}_{t-1})$ and $g_{t-1}(\mathbf{x}_{t-1})$. 
The overall algorithm is summarized as Algorithm \ref{algo:two-side}.

\noindent
% $L_t(\mathbf{x},\lambda) = f_t(\mathbf{x}) + \lambda_1^\top g_t(\mathbf{x}) + {\lambda_2}^\top g_t^{-}(\mathbf{x}_t) $ % = f_t(\mathbf{x}) + (\lambda_1-\lambda_2)^\top g_t(\mathbf{x})
\\
%To ensure $\sum_{t=1}^\top g_t(\mathbf{x}_t)\leq 0$ in long term, $\lambda$ updates as: $\mathbf{\lambda_{t+1}} = \max(0,\mathbf{\lambda_t} + \mu g_t(\mathbf{x}_t))$.\\
%Let $g_t^{neg}(\mathbf{x}_t)=-g_t(\mathbf{x}_t)$. To ensure $\sum_{t=1}^\top g_t^{neg}(\mathbf{x}_t)\leq 0$, $\lambda^{neg}$ updates as: $\mathbf{\lambda_{t+1}}^{neg} = \max(0,\mathbf{\lambda_t}^{neg} + \mu^{neg} g_t^{neg}(\mathbf{x}_t))$.\\
%$\lambda = \mathbf{\lambda_{t+1}} - \mathbf{\lambda_{t+1}}^{neg} = \max(0,\mathbf{\lambda_t} + \mu g_t(\mathbf{x}_t))  - \max(0,\mathbf{\lambda_t}^{neg} + \mu^{neg} g_t^{neg}(\mathbf{x}_t))$

\begin{algorithm} 
\caption{\underline{Lo}ng-\underline{t}erm \underline{Fair}ness-aware Online Learning Algorithm (LoTFair)}\label{algo:two-side}
\begin{algorithmic}[1]
\For{$t=1,2,\dots$}
    \State $\bm{\lambda}_t = [\lambda_{1,t}, \lambda_{2,t}]^\top$, $\mathbf{\bar{g}}_{t}(\mathbf{x})=[g_t(\mathbf{x}),-g_t(\mathbf{x})]^\top$
    \State observe constraint $g_{t-1}(\mathbf{x}_{t-1})$
    \State $\mathbf{x}_t=\min_{\mathbf{x}\in \mathcal{X}_t} \nabla f_{t-1}(\mathbf{x}_{t-1})^\top(\mathbf{x}-\mathbf{x}_{t-1}) + \bm{\lambda}_{t}^\top \mathbf{\bar{g}}_{t-1}(\mathbf{x}) + \frac{||\mathbf{x}-\mathbf{x}_{t-1} ||^2}{2\alpha}$
    \State observe current cost $f_{t}(\mathbf{x}_t)$ and constraint $g_{t}(\mathbf{x}_t)$
    \State $\lambda_{1,t+1} = \max(0,\lambda_{1,t} + \mu g_t(\mathbf{x}_t))$ 
    \State $\lambda_{2,t+1} = \max(0,\lambda_{2,t} - \mu g_t(\mathbf{x}_t))$ % \max(0,\lambda_{2,t} - \mu g_t(\mathbf{x}_t))
\EndFor
\end{algorithmic}
\end{algorithm}

%\begin{algorithm}
%\caption{Online saddle-point method}\label{one-side}
%\begin{algorithmic}[1]
%\Require $n \geq 0$
%\Ensure $y = x^n$
%\For{$t=1,2,\dots$}
%    \State $\mathbf{x}_t=\min_{\mathbf{x}} \nabla f_{t-1}(\mathbf{x}_{t-1})^\top(\mathbf{x}-\mathbf{x}_{t-1}) + \mathbf{\lambda_t}^\top g_{t-1}(\mathbf{x}) + \frac{||\mathbf{x}-\mathbf{x}_{t-1} ||^2}{2\alpha}$
%    \State observe current cost $f_{t}(\mathbf{x}_t)$ and constraint $g_{t}(\mathbf{x}_t)$
%    \State $\bm{\lambda}_{t+1} = \max(0,\bm{\lambda}_t + \mu g_t(\mathbf{x}_t))$ 
%\EndFor
%\end{algorithmic}
%\end{algorithm}

%\textbf{Long-term short-term online learning?}

\section{Practical Applications} \label{bi-classify}
%\shen{I still feel this section would be better to appear earlier. } \dm{Deepan: I agree with this point. It would be better to move this earler.}
Algorithm \ref{algo:two-side} can benefit various applications requiring making decisions `on the fly' with long-term fairness. %We will elaborate on two of them in further detail.
In this section, we demonstrate LoTFair on two practical real-time decision problems.

\subsection{Classification}
% In classical machine learning, when considering a 
Consider the classification task, where the objective is to minimize a loss function(e.g., cross-entropy) that reflects the {discrepency between the predicted label and the ground truth label}. 
%And logistic regression has emerged as a versatile and widely utilized technique for classification tasks.
%However, as the ethical implications of automated decision-making get more attention, a fairness concern has arisen.
%Purely minimizing a fairness-agnostic loss function may lead to discriminatory results towards under-represented groups, due to the lack of awareness of the discrepancy in group-wise performance.%is unable to control the distribution of errors across different subgroups. 
{Nonetheless, as mentioned earlier, solely optimizing the learning performance can compromise fairness, especially for sensitive and underrepresented groups.}
%The rapid generation, \rd{change and evolution} of data require online classification which brings complex challenges to the fusion of online classification and fairness. 
{While data streams in and decisions need to be made `on the fly', the potential for biased outcomes grows, perpetuating social inequalities and reinforcing existing disparities.}
To ensure fairness, the learner needs to mitigate bias and create an equitable logistic classifier for subgroups. {Enforcing instantaneous fairness constraint strictly, $g_t(\cdot) = 0$, at every time slot is too strict and can sometimes lead to negative consequences in the long run. For example, in the context of loan approval, if all the applicants from one subgroup happen to not meet the qualification standard, while the applicants from the other group are all qualified at the time slot $t$, enforcing the instantaneous fairness constraint will lead to the classifier to either approve un-qualified applicants or reject qualified ones, which is harmful to the overall global qualification decision in the long run.
%insufficient since such constraints may not adapt to the potential changing factors. In the context of loan approval cases, ensuring only instantaneous fairness will ignore the feature distribution shifts of applicants in different subgroups over time and such classification could lead to negative consequences in the long run.
} 

Let the classification problem be the binary classification problem: At each time slot $t$, there are $N$ individuals to be classified into two classes. 
Each individual $i$ entails feature vector $\mathbf{d}^i$. And $y^i \in \{-1,1\}$ denotes the associated label. %: $y^i=1$ represents the individual $i$ is classified as a positive case.
%\rd{Update definition of $\mathbf{x}_t$, not only decision, but also model parameters.}

{In this case, the online learner is learning % to decide 
the parameter $\mathbf{x}_t$ of the online classifier, which predicts the label based on the feature vector, $\mathbf{d}^i$. For example, when logistic regression is employed, the learner predicts the probability of individual $i$ as
\begin{align*}
    \hat{y}^i = \frac{e^{\mathbf{x}_t^\top \mathbf{d}^i}}{1+ e^{\mathbf{x}_t^\top \mathbf{d}^i}}. % \hat{y}
\end{align*}
}
%where $\hat{y}^i$ represents the probability of being predicted as positive. %``income exceeds \$50,000" and the bank will approve the application

%A common approach is adding an instantaneous constraint to ensure the biases across different subgroups are acceptable. 

We adopt the standard loss function for the performance which is the cross-entropy loss \cite{de2005tutorial,mannor2005cross}.
\begin{align} \label{cross-entropy}
    f_t(\mathbf{x}_t) = -\sum_{d^i \in \mathbf{D}_t}  \frac{y^i+1}{2} \log \hat{y}^i + \frac{1-y^i}{2} \log (1-\hat{y}^i). 
\end{align}
%where $y^i$ is the ground truth label for individual $i$. 

%Hence the set $\mathcal{G}_T^{I}$ is defined as
%$\mathcal{G}_T^{I} = \left\{ [\mathbf{x}_t ]_{t=1}^T \bigg|  g_t^{\kappa}(\mathbf{x}_t) = \left( DP(\mathcal{D}_0) - DP(\mathcal{D}_1) \right)^2 - \kappa \leq 0, \forall t \in \{1,\cdots,T \}  \right\}$
%\\

%$\mathcal{G}_T^{L} = \left\{ [\mathbf{x}_t ]_{t=1}^T \Bigg| \sum_{t=1}^T g_t(\mathbf{x}_t) = \sum_{t=1}^T \left( DP(\mathcal{D}_0) - DP(\mathcal{D}_1) \right) = 0 \right\}$
\noindent
{Hence the set $\mathcal{G}_T^{L}$ is defined as
\begin{align} \label{long-fair-constraint-log}
    \mathcal{G}_T^{L} =\left\{ [\mathbf{x}_t ]_{t=1}^T \Bigg|  \sum_{t=1}^T \left( \Omega(\mathcal{D}_0,\mathbf{x}_t) - \Omega(\mathcal{D}_1,\mathbf{x}_t) \right) = 0 \right\} %\sum_{t=1}^T g_t(\mathbf{x}_t) = DP(\mathcal{D}_0) - DP(\mathcal{D}_1)
\end{align}
where $g_t(\mathbf{x}_t)=\Omega(\mathcal{D}_0,\mathbf{x}_t) - \Omega(\mathcal{D}_1,\mathbf{x}_t)$, $\Omega(\mathcal{D}_0,\mathbf{x}_t) = \frac{1}{|\mathcal{D}_0|} \sum_{i \in \mathcal{D}_0} \hat{y}^i $ %\shen{as mentioned in our meeting, this needs to be changed, it is not DP}
represents the average probability that the group $\mathcal{D}_0$ is predicted as 1 \cite{chuang2021fair,han2023retiring}. The $g_t(\mathbf{x}_t)=0$ means the classifier satisfies demographic parity at time slot $t$: different subgroups have equal probability of being assigned to the positive predicted class.} %\shen{I'm a bit confused, but when learner use Algorithm 1, $\theta$ instead of should be updated?}

The overall online classification problem can be formulated as \eqref{long_general}, which the task-specific loss function defined in \eqref{cross-entropy} and the constraint set $\mathcal{X}_t$ as $\mathbb{R}^n$, $\mathcal{G}_T^{L}$ as \eqref{long-fair-constraint-log}. %viewed as the following:
%\begin{align}
%    \min_{\mathbf{x}_t} \sum_{t=1}^T f_t(\mathbf{x}_t) ~~ \text{s.t.} \qquad [\mathbf{x}_t]_{t=1}^T \in \mathcal{G}_T^{L} \label{objective:log}
%\end{align}
The numerical experiments on the ``Adult" dataset are presented in Section \ref{adult-exp}.

\subsection{Peer-to-peer Electricity Market} \label{P2P}
Another example is resource allocation in networks or energy management in power grids.
In this section, we discuss the peer-to-peer(P2P) electricity market \cite{PARK20173},\cite{P2P-CBM-reveiw},\cite{P2P-elec-distribution}. We formulate the energy management problem in the peer-to-peer(P2P) electricity market as an online optimization problem like (\ref{long_general}). Besides the objective function and long-term fairness constraints, there are additional constraints such as the physical limits of the system. %inherited from the physical limits of the system.

Specifically, consider the P2P network with $N$ nodes that are responsive to grid conditions such as energy prices, local energy supply and demand \cite{P2P-elec-distribution}.
%A node in the power grid also represents a client in the P2P market
All nodes are divided into 2 groups $\mathcal{D}_0$ and $\mathcal{D}_1$ based on their sensitive attributes. At time $t$, each client $i$ corresponds to node $i$, has demand $d^i_t$ and supply $s^i_t$. {In this P2P network, a client is a prosumer who can sell or buy energy with other clients.}
%Assume $d_i^t/s_i^t$ is predicted 100\% correctly based on client's historical features such as status of Wet Appliance, heating ventilation and Air conditions(HAVC), solar panels, etc.\\
%The consumers whose $s^i_t-d^{i}_t < 0$ need to buy the power from producer with cheaper price or they have to buy the electricity from utilities. Producer i has redundant power as $s^i_t-d^i)t \geq 0$ at time t.\\
Each client has surplus or deficit as given by $h^i_t=s^i_t-d^{i}_t$. The client $i$ is considered as a producer at time $t$ if $h^i_t \geq 0$; otherwise client $i$ is considered as a consumer since $h^i_t \leq 0$.
The decision $\mathbf{X}_t$ is an $N \times N$ matrix with the $(i,j)$th entry representing the trade between node $i$ and $j$. Specifically, if $\mathbf{X}_t^{ij}>0$, it denotes client $i$ sells $\mathbf{X}_t^{ij}$ units of energy to client $j$; otherwise, client $i$ buys energy from client $j$.
{After P2P trading, for consumer $i \in w^c_t$, it has deficit $\Delta^{i,-}_t=s^i_t-d^i_t - \sum_{j \in N} \mathbf{X}_t^{ij} = h^i_t - \sum_{j \in N} \mathbf{X}_t^{ij}$.}
Assuming the price of unit energy traded in P2P electricity market $e^{ij}_t, \forall i,j \in N$ is cheaper than the price of utility company $p$, consumers will buy electricity from producers in the P2P market first then buy electricity from utility companies. 

%\blue{[define and express the task-specific fairness.]} 
The goal is to minimize the total cost to satisfy the supply-demand balance for all consumers. Therefore, the learner tends to satisfy consumers with lower costs first due to their convenient locations or advanced power transmission system, and the advantageous group would potentially have a higher satisfaction rate in P2P market transactions. 
{In this scenario, enforcing the instantaneous fairness constraint at each time slot may lead to a significant increase in cost. Due to the complex dynamics of power systems, the feasible solution of the problem subject to the instantaneous fairness constraint sometimes may make the power system unstable (e.g., exceeding transmission line capacity), even if the constraint is relaxed as \eqref{instantaneous-set}.} % \shen{cannt follow}
To ensure fairness in the long run, each group should have a similar satisfaction rate on average. 
% Constraint: each group has similar (dis)satisfaction rate at any time. 
Hence, the set $\mathcal{G}_T^{I}$ is formulated as 
\begin{align*}
   \mathcal{G}_T^{I} = \left\{ [\mathbf{X}_t ]_{t=1}^T \bigg|  g_t^{\kappa}(\mathbf{X}_t) % =  \left( \frac{1}{|\mathcal{D}_0|} \sum_{i \in \mathcal{D}_0 \cap w^c_t} \frac{\Delta^{i,-}_t}{d^i_t} - \frac{1}{|\mathcal{D}_1|} \sum_{i \in \mathcal{D}_1 \cap w^c_t} \frac{\Delta^{i,-}_t}{d^i_t} \right)^2 - \kappa 
   \leq 0  \right\}
\end{align*}
where $$ g_t^{\kappa}(\mathbf{X}_t) =  \left( \frac{1}{|\mathcal{D}_0|} \sum_{i \in \mathcal{D}_0 \cap w^c_t} \frac{\Delta^{i,-}_t}{d^i_t} - \frac{1}{|\mathcal{D}_1|} \sum_{i \in \mathcal{D}_1 \cap w^c_t} \frac{\Delta^{i,-}_t}{d^i_t} \right)^2 - \kappa.$$
The set $\mathcal{G}_T^{L}$ is formulated as 
\begin{align*}
    \mathcal{G}_T^{L} = \left\{ [\mathbf{X}_t ]_{t=1}^T \bigg| \sum_{t=1}^T g_t(\mathbf{X}_t) %=  \sum_{t=1}^T \left(\frac{1}{|\mathcal{D}_0|} \sum_{i \in \mathcal{D}_0 \cap w^c_t} \frac{\Delta^{i,-}_t}{d^i_t} - \frac{1}{|\mathcal{D}_1|} \sum_{i \in \mathcal{D}_1 \cap w^c_t} \frac{\Delta^{i,-}_t}{d^i_t} \right) 
    = 0 \right\}.
    %g_t(\mathbf{X}_t) = .
\end{align*}
where $$  g_t(\mathbf{X}_t) =  \left(\frac{1}{|\mathcal{D}_0|} \sum_{i \in \mathcal{D}_0 \cap w^c_t} \frac{\Delta^{i,-}_t}{d^i_t} - \frac{1}{|\mathcal{D}_1|} \sum_{i \in \mathcal{D}_1 \cap w^c_t} \frac{\Delta^{i,-}_t}{d^i_t} \right) .$$
\\
And the cost function $f_t(\mathbf{X}_t)$ is
\begin{align*}
    f_t(\mathbf{X}_t) = \frac{1}{|w_t^c|}  \big(& - \sum_{i \in w_t^c} \left( (h_t^i -\sum_{j \in N} \mathbf{X}_t^{ij})p + \sum_{j \in N}e^{ij}_t \mathbf{X}_t^{ij} \right) \\
    &-\sum_{j \in w_t^c} \sum_{i \in N} \gamma \mathbf{D}^{ji} \mathbf{X}_t^{ji} \big)% \blue{\mathbf{1}_{w_t^c}^\top M^{D} \mathbf{X}_t \mathbf{1}}
\end{align*}
{where $p$ is the unit price of energy trading with the utility company, $e^{ij}$ denotes the trading price between client $i$ and $j$ in the P2P electricity market, $\mathbf{D}$ denotes the time-invariant power transfer distance matrix can be obtained from the grid (Appendix \ref{flow-constraint})
% \shen{I don't understand this notation? Is it a scalor, why use so complicated notation?does it actually change with the value or x?}
and $\gamma$ denotes the unit price of the transmission line utilization fee.}
The sum term $-\sum_{j \in w_t^c} \sum_{i \in N} \gamma \mathbf{D}^{ji} \mathbf{X}_t^{ji}$ represents the total transmission line utilization fee.
The objective function attempts to minimize the total cost to purchase enough energy to satisfy the supply-demand balance for all consumers. 
The overall online energy management % or resource allocation
problem can be viewed as \eqref{long_general} with the corresponding $ f_t(\mathbf{X}_t)$, $\mathcal{G}_T^{L}$, and $\mathcal{X}_t$ is shown as

%\begin{subequations}
%\begin{equation}
%  \min_{\mathbf{X}_t \in \mathcal{X}_t, \forall t} \sum_{t=1}^T f_t(\mathbf{X}_t)
%  ~~ \text{s.t.} \qquad [\mathbf{X}_t]_{t=1}^T \in \mathcal{G}_T^{L}
%\end{equation}
%where
%  \[
%   \mathcal{X}_t = \left\{\begin{array}{lr}
%         \mathbf{X} \in \mathbb{R}^{N \times N}, & \text{for } 0\leq n\leq 1\\
%         0 \leq \mathbf{X}_t^{ij} \leq h^i_t ,\forall i \in w^p_t, \forall j \in \{1,\cdots,n\}, \forall t , & \text{for } 0\leq n\leq 1\\
%        h^i_t \leq \mathbf{X}_t^{ij} \leq 0 ,\forall i \in w^c_t, \forall j \in \{1,\cdots,n\}, \forall t, & \text{for } 0\leq n\leq 1
%        \end{array}\right\}
%  \]\shen{I tried to help, but you need to first fix the t index.. }
\mathleft
\begin{equation} \small
%\[
  \hfill \left\{\begin{array}{lrr r}
         \mathbf{X}_t \in \mathbb{R}^{N \times N} &  (13\text{a})\\
         0 \leq \mathbf{X}_t^{ij} \leq h^i_t ,  \forall i \in w^p_t,  \forall j \in \{1,\cdots,n\}  &  (13\text{b})\\
         h^i_t \leq \mathbf{X}_t^{ij} \leq 0 ,  \forall i \in w^c_t, \forall j \in \{1,\cdots,n\}  &  (13\text{c})\\
         0 \leq \sum_{j \in N_i} \mathbf{X}_t^{ij} \leq h^i_t ,  \forall i \in w^p_t &  (13\text{d})\\
         h^i_t \leq \sum_{j \in N_i} \mathbf{X}_t^{ij} \leq 0,  \forall i \in w^c_t &  (13\text{e})\\
         \mathbf{X}_t + \mathbf{X}_t^\top = \mathbf{0}&  (13\text{f})\\
         -\rho^l_{\max} \leq \rho^l(\mathbf{X}_t) \leq \rho^l_{\max},   \forall l \in \{1,\cdots,N_l\} &  (13\text{g}) \\
         g_t^{\kappa}(\mathbf{X}_t) \leq {\tau} &  (13\text{h})\\
        \end{array} 
        \right\}.
%\]
\label{X_t-set}
\end{equation}

\mathcenter

% \end{subequations}

%\section{other application}
% \Eqref{con_a} and \eqref{con_b} are the long-term fairness constraints.  
\noindent
Hence, $\mathcal{X}_t$ is the physical constraints of the P2P electricity market.
(13b) and (13c) guarantee that producers only sell energy and consumers only buy energy. 
And (13d) and (13e) ensure that producers do not sell more than their redundancy and consumers do not buy more than their deficit. Thus, after P2P trading, for consumer $i \in w^c_t$, it has deficit $\Delta^{i,-}_t \leq 0$. % =s^i_t-d^i_t - \sum_{j \in N} \mathbf{X}_t^{ij} = h^i_t - \sum_{j \in N} \mathbf{X}_t^{ij}
%$\mathbf{X}_t + \mathbf{X}_t^\top = \mathbf{0}$
(13f) represents the equal trading volume and offsetting purchase and sell. 
% $-\rho^l_{\max} \leq \rho^l(\mathbf{X}_t) \leq \rho^l_{\max}$
(13g) denotes the flow constraint, which restricts the energy flow from %avoids the energy flow 
exceeding transmission line capacity (details in Appendix \ref{flow-constraint}).
% $g_t^{\kappa}(\mathbf{X}_t) <= \tau$
(13h) prevents very unfair outcomes, $\tau$ is a larger constant than $\kappa$, $\tau > \kappa$.

%The numerical experiments on synthetic data and real dataset are presented in Section \ref{experiments}.

\section{Performance Analysis} \label{analysis}
In this section, two metrics are introduced to measure the performance and fairness of online decisions: dynamic regret and dynamic fairness \cite{mokhtari2016online}, \cite{chen2017online}. We rigorously prove that dynamic regret and dynamic fairness are both sub-linear.
\subsection{Performance Metrics}
%Firstly, we introduce dynamic regret which measures the performance:
Dynamic regret is formally defined as:
%\shen{but there are no constraints in the following equation?}
\begin{align} \label{regret}
    \mathbf{\mathcal{R}}_T \coloneqq \sum_{t=1}^T f_t(\mathbf{x}_t) - \sum_{t=1}^T f_t(\mathbf{x}_t^*) % Reg_T
\end{align}
where $ [\mathbf{x}_t^* ]_{t=1}^T = [\mathbf{x}_1^*,\cdots,\mathbf{x}_t^*,\cdots,\mathbf{x}_T^*]$ % ,
%are obtained as: % the optimal solution of \eqref{objective}. \\
and
\begin{align}
    \mathbf{x}_t^* = \arg \min_{\mathbf{x}_t \in \mathcal{X}_t} f_t(\mathbf{x}_t) ~~ \text{ s.t. } g_t(\mathbf{x}_t) = 0 .\label{dynamic_x*}
\end{align}

%To evaluate the feasibility of online decisions, dynamic fairness is introduced to measure the accumulated fairness violation of constraints, which is defined as:
The dynamic fairness of the sequence of online decisions is determined by the cumulative violation of the fairness:
\begin{align} \label{dynamic_fair}
    \mathbf{\mathcal{F}}_T \coloneqq  \sum_{t=1}^T g_t(\mathbf{x}_t). % Fit_T
\end{align}
% The definition in \eqref{dynamic_fair} \rd{allows} that the instantaneous fair violation(e.g. $g_t(\mathbf{x}_t)\geq 0$) can be compensated by future decisions with opposite violations(e.g. $g_{t+1}(\mathbf{x}_t)\leq 0$). %\shen{I cannot see why this claim is true} 
% It provides a more flexible decision set at each time slot, which could potentially sacrifice much less performance drop on the fundamental objective function.\shen{can't follow this sentence} 
%\rd{Over the long term, the overall violation becomes negligible.}
%\rd{In the cases of cyber-physical systems, such as energy management in power grids and resource allocation in networks, since there usually exist other types of constraints, such as transmission limitations for power grids \cite{8782819}, introducing long-term constraint offers the flexibility to avoid significant increases in $f_t(\mathbf{x}_t)$ compared with meeting the instantaneous constraint at each time slot.}

An ideal algorithm should achieve both sub-linear dynamic regret and sub-linear dynamic fairness on the long-term average, i.e., $\lim_{T\rightarrow \infty}\frac{\mathbf{\mathcal{R}}_T}{T} =0$ and $\lim_{T\rightarrow \infty}\frac{\mathbf{\mathcal{F}}_T}{T} =0$.
%\shen{YS revised till here}

\subsection{Performance Bounds}
We assume that the following conditions are satisfied:

\noindent\textit{Assumption 1}: For every $t$, the cost function $f_t(\mathbf{x})$ and the time-varying fairness constraint $g_t(\mathbf{x})$ are convex.\\
\textit{Assumption 2}: For every $t$, $f_t(\mathbf{x})$ has bounded gradient on $\mathcal{X}$, $|| \nabla f_t(\mathbf{x}) ||\leq G, \forall \mathbf{x} \in \mathcal{X}$; and ${g}_t(\mathbf{x})$ is bounded, $||g_t(\mathbf{x})||\leq M$. \\%, $||\mathbf{\bar{g}}_t(\mathbf{x})|| \leq \sqrt{2}M$.\\
\textit{Assumption 3}: The radius of the convex feasible set $\mathcal{X}$ is bounded, $||\mathbf{x}-\mathbf{y}||\leq R,\forall \mathbf{x},\mathbf{y}\in \mathcal{X}$. And the decision $\mathbf{x}_t$ is selected from a convex set $\mathcal{X}_t \subseteq \mathcal{X}\in \mathbb{R}^n$ \\% \blue{The radius of $g_t(\mathbf{x})$ is bounded, $||g_t(\mathbf{x})-g_t(y) || \leq S$.}\\
\textit{Assumption 4}: There exists a constant $\epsilon>0$, and two interior points $\mathbf{\tilde{\mathbf{x}}}_t$ and $\tilde{\tilde{\mathbf{x}}}_t$ such that $g_t(\tilde{\mathbf{x}}_t) \leq -\epsilon$ and $g_t(\tilde{\tilde{\mathbf{x}}}_t) \geq \epsilon,\forall t$. \\ 
\textit{Assumption 5}: The constant $\epsilon$ in Assumption 4 satisfies $\epsilon > \bar{V}(g)$, which is the point-wise maximum variation of the time-varying constraints and is defined as \\ % \max(0,g_{t+1}(\mathbf{x}) - g_t(\mathbf{x}))
\begin{align*}
    \bar{V}(g) = \max_{t\in T} \max_{\mathbf{x} \in \mathcal{X}} | g_{t+1}(\mathbf{x}) - g_t(\mathbf{x}) |.
\end{align*}

% Since the decision $\mathbf{x}_t$ is selected from a convex set and the loss function $f_t(\cdot)$ is convex, 
Based on assumptions 1 and 3, the online optimization problem (\ref{objective}) is an online convex optimization(OCO) problem.
Firstly, the following theorem states the upper bound and lower bound of dynamic fairness for Algorithm \ref{algo:two-side}.\\
\textbf{Theorem 1:} Under assumptions 1-5, using  Algorithm \ref{algo:two-side} with dual variable initialization as $\bm{\lambda}_1=\mathbf{0}$, the ${\lambda}_{1,t}$ and ${\lambda}_{2,t}$ are bounded by
\begin{align}
    {\lambda}_{1,t}, {\lambda}_{2,t} \leq \bar{\lambda} \coloneqq 2 \mu M + \frac{2GR+\frac{R^2}{2\alpha} + \mu M^2}{\epsilon-\bar{V}(g)} \label{bar_lambda}
\end{align}
and the dynamic fairness of \eqref{dynamic_fair} is bounded by
{
\begin{align} \label{theorem1}
    %-2M - \frac{\frac{2GR}{\mu}+\frac{R^2}{2\mu \alpha} + M^2}{\epsilon-\bar{V}(g)} \leq
    |\mathbf{\mathcal{F}}_T| \leq 2M + \frac{\frac{2GR}{\mu}+\frac{R^2}{2\mu \alpha} + M^2}{\epsilon-\bar{V}(g)}. % \sum_{t=1}^Tg_t(\mathbf{x}_t)
\end{align} }
\textit{Proof of {Theorem 1}: see Appendix \ref{proof_theorem1}.}

Under the conditions on the time-varying constraint, Theorem 1 shows that $\mathbf{\mathcal{F}}_T$ depends on primal and dual stepsizes. $|\mathbf{\mathcal{F}}_T|$ is in the order of $\mathcal{O}(\frac{1}{\mu})$ with fixed primal stepsize $\alpha$. Thus, a larger dual stepsize is beneficial for the long-term fairness constraint.

We further bound the dynamic regret of \eqref{regret} in the next theorem.
\\
\textbf{Theorem 2:} Under assumptions 1-5, using  Algorithm \ref{algo:two-side} with dual variable initialization as $\bm{\lambda}_1=\mathbf{0}$, the dynamic regret of \eqref{regret} is upper bounded by
{
\begin{align} \label{theorem2}
    \mathbf{\mathcal{R}}_T =& \sum_{t=1}^Tf_t(\mathbf{x}_t) - \sum_{t=1}^T f_t(\mathbf{x}_t^*) \nonumber \\ 
    \leq& \frac{R}{\alpha}V(\{ \mathbf{x}_t^*\}_{t=1}^T) +\frac{R^2}{2\alpha} + | \bar{\lambda} | V(\{\mathbf{\bar{g}}_t\}_{t=1}^T) \nonumber \\
    & + \mu M^2 T + \frac{\alpha G^2 T}{2} + \frac{\mu M^2}{2}.
\end{align}
}
$V(\{ \mathbf{x}_t^*\}_{t=1}^T)$ and $ V(\{\mathbf{\bar{g}}_t\}_{t=1}^T)$ denote the accumulated variations of the minimizers $\mathbf{x}_t^*$ and the fairness constraints at every time slot, which are defined as follows:
\begin{align}
    V(\{ \mathbf{x}_t^*\}_{t=1}^T) &\coloneqq \sum_{t=1}^T V(\mathbf{x}_t^*) = \sum_{t=1}^T ||\mathbf{x}_t^* - \mathbf{x}_{t-1}^*||, \label{var_x} \\
    V(\{\mathbf{\bar{g}}_t\}_{t=1}^T) &\coloneqq \sum_{t=1}^T  V(\mathbf{\bar{g}}_t) = \sum_{t=1}^T \max_{\mathbf{x}} ||\mathbf{\bar{g}}_{t+1}(\mathbf{x}) - \mathbf{\bar{g}}_{t}(\mathbf{x}) ||. \label{var_g}
\end{align}
\textit{Proof of {Theorem 2}: see Appendix \ref{proof_theorem2}.}

It can be observed that the dynamic regret $\mathbf{\mathcal{R}}_T$ is bounded by the primal stepsize, dual stepsize and accumulated variations of instantaneous minimizers $V(\{ \mathbf{x}_t^*\}_{t=1}^T)$ and time-varying constraints $V(\{\mathbf{\bar{g}}_t\}_{t=1}^T)$. The upper bound could be small while choosing appropriate primal and dual stepsizes.

Under assumptions 1-5 and without knowledge of variations($ V(\{ \mathbf{x}_t^*\}_{t=1}^T)$ and $V(\{\mathbf{\bar{g}}_t\}_{t=1}^T)$), according to Theorems 1-2, if the primal and dual stepsizes are chosen as $\alpha=\mu = \mathcal{O}(T^{-\frac{1}{3}})$, then the dynamic fairness and regret can be bounded as
\begin{align*}
    |\mathbf{\mathcal{F}}_T| & = \mathcal{O}(T^{\frac{2}{3}})\\
    \mathbf{\mathcal{R}}_T & = \mathcal{O}( \max \big\{ V(\{ \mathbf{x}_t^*\}_{t=1}^T) T^{\frac{1}{3}}, V(\{\mathbf{\bar{g}}_t\}_{t=1}^T) T^{\frac{1}{3}}, T^{\frac{2}{3}}  \big\} ).
\end{align*}
If the variations, $V(\{ \mathbf{x}_t^*\}_{t=1}^T)$ and  $ V(\{\mathbf{\bar{g}}_t\}_{t=1}^T)$, are known, the primal and dual stepsizes can be chosen as 
\begin{align*}
    \alpha = \mu = \sqrt{\frac{ \max \big\{ V(\{ \mathbf{x}_t^*\}_{t=1}^T),V(\{\mathbf{\bar{g}}_t\}_{t=1}^T) \big\} }{T}},
\end{align*}
then the regret can be bounded by
\begin{align*}
    \mathbf{\mathcal{R}}_T & = \mathcal{O} \left(  \max \left\{ \sqrt{V(\{ \mathbf{x}_t^*\}_{t=1}^T) T}, \sqrt{V(\{\mathbf{\bar{g}}_t\}_{t=1}^T) T} \right\} \right),
\end{align*}
and the dynamic fairness is bounded as
\begin{align*}
    \mathbf{\mathcal{F}}_T & = \mathcal{O} \left(  \max \left\{ \frac{T}{V(\{ \mathbf{x}_t^*\}_{t=1}^T)}, \frac{T}{V(\{\mathbf{\bar{g}}_t\}_{t=1}^T)} \right\} \right).
\end{align*}
{\textit{Proof: Appendix \ref{bound_proof}}}
%\blue{choose $\alpha = \mu = \mathcal{O}(T^{-\frac{1}{3}})$, dynamic regret and cumulative fair violation increases sub-linearly.}

Although the dynamic benchmark $\mathbf{x}_t^*$ for regret in (\ref{regret}) is got through \eqref{dynamic_x*},
%which only requires instantaneous information, i.e., $f_t(\mathbf{x})$ and $g_t(\mathbf{x}_t)$, 
$[\mathbf{x}_t^*]_{t=1}^T$ is not the optimal solution to the problem (\ref{long_general}). % \shen{which problem?}. 
Instead, the optimal solutions can be obtained as the offline optimal solutions of \eqref{long_general}, denoted as $[\mathbf{x}_t^{\text{off}}]_{t=1}^T$. %, which requires the information over the entire time horizon. %However, solving \eqref{long_general} 
%In another word, the per-slot minimizer $\mathbf{x}_t^*$ only requires one-slot information($f_t(\mathbf{x})$ and $g_t(\mathbf{x}_t)$), the offline solution $\mathbf{x}_t^{\text{off}}$ comes from offline computation with all information over $T$ slots. 
{However, solving the problem (\ref{long_general}) offline requires information {about $f_t(\cdot)$ and $g_t(\cdot)$} over the entire time horizon, which is not practical in real scenarios. 
And at time slot $t$, previous decisions $[\mathbf{x}_0,\cdots,\mathbf{x}_{t-1}]$ cannot be changed after receiving latest information, $f_{t}(\mathbf{x}_{t})$. Instead, the problem needs to be solved online: at each time slot $t$, the decision $\mathbf{x}_{t}$ needs to be made only according to current and inherited information.}
In the implementations, especially the \textbf{time-varying} setting, solving \eqref{dynamic_x*} frequently fails, so we use $\mathbf{x}_t^{\text{off}}$ as the dynamic benchmark for some experiments {where the corresponding regret, named as offline dynamic regret,} is computed as
\begin{align*}
    \mathbf{\mathcal{R}}^{\text{off}}_T = \sum_{t=1}^T f_t(\mathbf{x}_t) - \sum_{t=1}^T f_t(\mathbf{x}_t^{\text{off}}).
\end{align*}
where the superscript denotes the dynamic benchmark is $\mathbf{x}_t^{\text{off}}$. 
{Since $ \mathbf{\mathcal{R}}^{\text{off}}_T = \mathbf{\mathcal{R}}_T + \sum_{t=1}^T f_t(\mathbf{x}_t^*) - \sum_{t=1}^T f_t(\mathbf{x}_t^{\text{off}}) \geq \mathbf{\mathcal{R}}_T $, the upperbound of $\mathbf{\mathcal{R}}^{\text{off}}_T$ is larger than the upperbound shown in \eqref{theorem2}. 
We discuss further the upperbound of the offline dynamic regret in Appendix \ref{offline_upperbound}.} % \shen{why do we care about the upper bound? needs clarification}

In this section, we present that dynamic regret and dynamic fair increase sub-linearly under Assumptions 1-5 while choosing suitable primal and dual stepsizes. %We will test our method on 2 applications introduced in Section \ref{bi-classify} and Section \ref{P2P}.
Later, we show the effectiveness of LoTFair in controlling unfairness while optimizing performance on the two applications in Section \ref{all-experiment}.

\section{Experiments} \label{all-experiment}
{This section tests the proposed algorithm in the  two applications as described in Section \ref{bi-classify}.}
\iffalse
\noindent\textbf{Evaluation Metrics:} 
To demonstrate LoTFair, we present the time-average cost, dynamic regret, and dynamic fairness to evaluate if the learner could deliver accurate decisions while addressing the fairness concerns. We further introduce averaged absolute fair violation($A^2FV$)
\begin{align*}
    A^2FV=\frac{1}{T}\sum_{t=1}^T |g_t(\mathbf{x}_t)|
\end{align*}
\rd{which captures the average magnitude of fair violations.}
\fi
\subsection{Classification} \label{adult-exp}
\noindent
\textbf{Datasets and Settings:}
In this section, we evaluate the proposed algorithm using a logistic binary classifier 
and conduct tests on the ``Adult'' dataset, which contains anonymous data on adult individuals in the United States. The objective is to predict whether an individual's annual income exceeds \$50,000 by analyzing the various features representing demographic, educational and employment-related information. The income information is often seen as a critical factor for certain decisions: like loan applications, a bank may approve an application based on the individual's income exceeding \$50,000.
The ``Adult" dataset poses fairness challenges, as some features can be considered sensitive attributes, potentially leading to biased decisions across different demographic groups, such as gender, race, and age.
In this simulation, we assume that at each time slot $t$, $N$ individuals apply for the loan. Each individual $i$ corresponds to feature $d^i$. $y^i \in \{-1,1\}$ is associated with its label: $y^i=1$ denotes $i$'s income exceeds \$50,000 and the bank will approve the loan application of $i$. All individuals are divided into 2 groups $\mathcal{D}_0$ and $\mathcal{D}_1$ based on their sensitive attributes (e.g., race and gender).
%The goal is to minimize the total loss (cost) to make correct predictions for all individuals. 

\noindent
\noindent\textbf{Methods:} we compare the results of the proposed method, LoTFair, and 2 baselines: %3 different approaches:
{
\begin{itemize}[leftmargin=*]
\item\textbf{Long-term:} LoTFair (Algorithm \ref{algo:two-side}). %Solve \eqref{long_general} via
\item\textbf{Instantaneous:} $\mathbf{x}_t =\arg \min_{x} f_t(\mathbf{x}) ~~ \text{ s.t. }  g_t^{\kappa}(\mathbf{x})\leq 0$.  % , \qquad \forall t \in \{1,\cdots,T \}$\\
%\blue{\textbf{ \textbullet Offline:} Solve the problem in \eqref{long_general} offline. ----------}
\item\textbf{Stochastic Gradient Descent (SGD):} training by stochastic gradient descent without fairness constraint: $\mathbf{x}_{t+1} = \mathbf{x}_t - \alpha \nabla_{\mathbf{x}}f_t(\mathbf{x}_t)$. %\shen{I don't understand, this should be stochastic gradient descent, right?}
\end{itemize}
}
% \textbf{ \textbullet Offline:} Solve the problem in \eqref{long_general} offline.
%\shen{are these baselines the same as the baselines in the next experiment? If so, unify the name. If not, why?}

\noindent\textbf{Experimental results:} In this experiment, gender is treated as the sensitive attribute: $\mathcal{D}_0$ denotes the female individuals and $\mathcal{D}_1$ denotes the male individuals. And for the instantaneous approach, %we replace the equality constraint with inequality constraint 
$g_t^{\kappa}(\mathbf{x})=\left( \Omega(\mathcal{D}_0,\mathbf{x}_t) - \Omega(\mathcal{D}_1,\mathbf{x}_t) \right)^2-0.04^2\leq 0$ {where $\Omega(\cdot)$ is defined in \eqref{long-fair-constraint-log}.
Since $g_t(\mathbf{x})=0$ is more strict than $g_t^{\kappa}(\mathbf{x})\leq 0$, the instantaneous approach is the relaxed version of \eqref{dynamic_x*}: the solution of the instantaneous approach will have an equal or smaller cost value, $f_t(\mathbf{x})$, than the solution of \eqref{dynamic_x*}. In the implementation, it is unable to obtain the solution of \eqref{dynamic_x*} so we simulate the $\mathbf{x}_t^*$ from the relaxed approach.} %\shen{clarify $\Omega$}

\begin{figure*}
    \centering
    \vspace{-4mm}
    \captionsetup{justification=centering}
    \subfigure[Time-averaged cost]{ \label{log:cost}
    \includegraphics[width=0.4\textwidth]{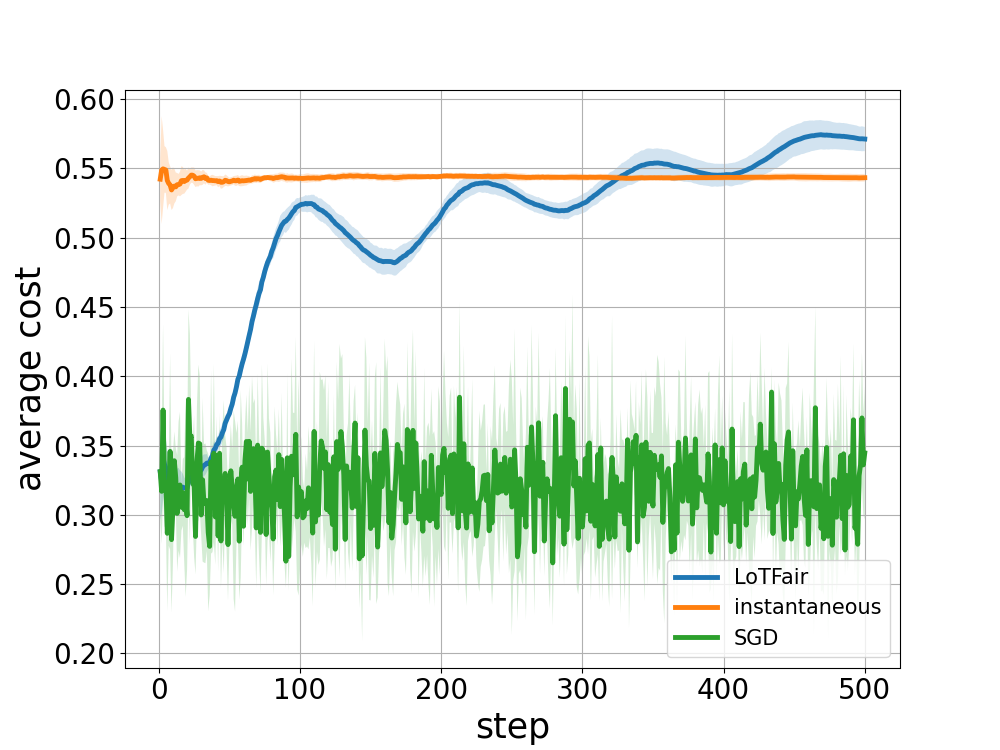}} 
    \quad
    \subfigure[Time-averaged dynamic regret]{ \label{log:regret}
    \includegraphics[width=0.4\textwidth]{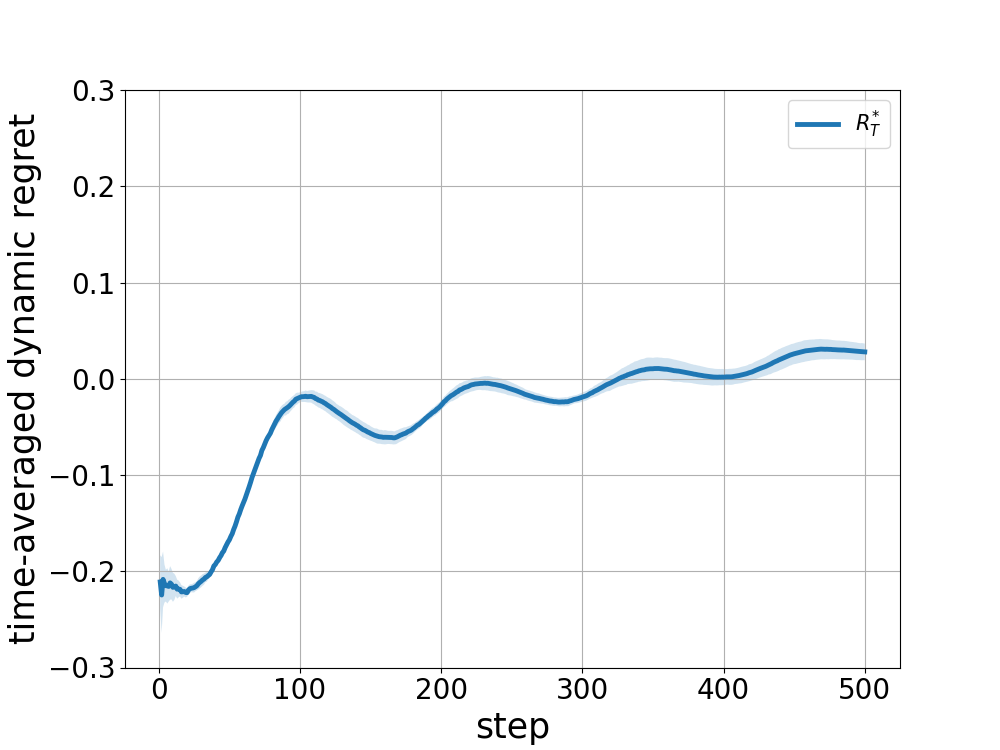}} % some mistakes
    \caption{Results of (a) time-averaged costs and (b) dynamic regrets on Adult dataset.}\label{log:cost&regret}
\end{figure*}
\iffalse
\begin{figure*}
    \centering
    \captionsetup{justification=centering}
    \subfigure[$\mathbf{\mathcal{F}}_T$]{ \label{log:dynamic_fair}
    \includegraphics[width=0.33\textwidth]{log/dynamic_fair.png}} 
    \quad
    \subfigure[$A^2FV$]{ \label{log:A2FV}
    \includegraphics[width=0.33\textwidth]{log/A2FV.png}}
    \caption{Experimental results of Adult dataset, dynamic fairness and $A^2FV$. 
    }\label{log:dynamic_fair&A2FV}
\end{figure*}
\fi
\begin{figure}
    \centering
    \vspace{-4mm}\includegraphics[width=0.4\textwidth]{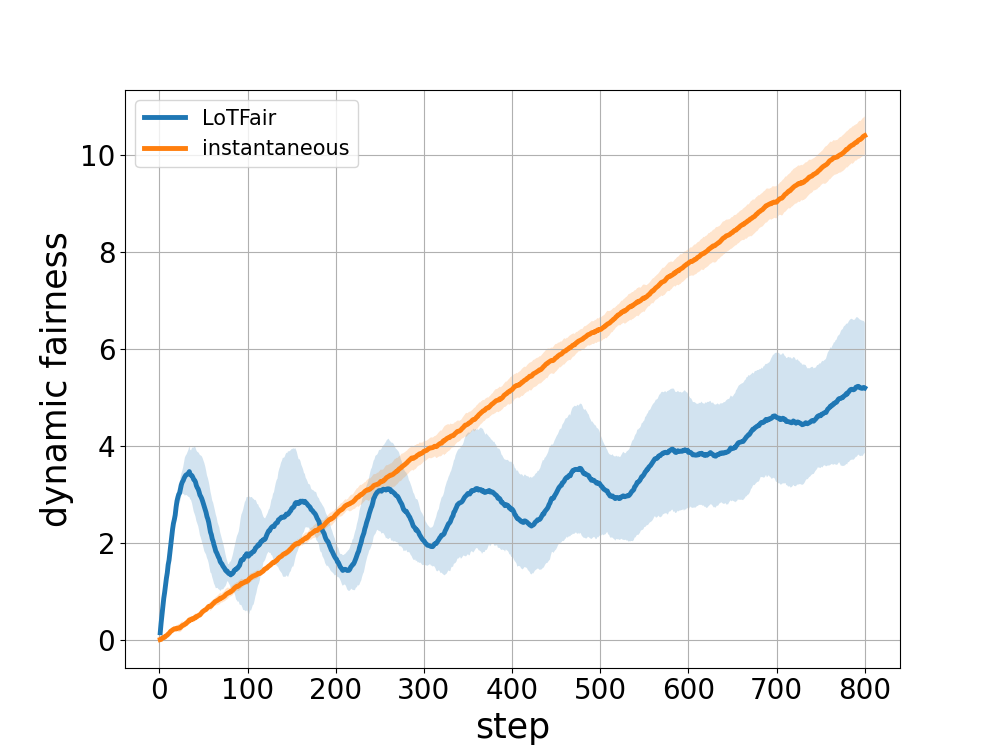}
    \caption{Dynamic fairness results on Adult dataset.}
    \label{log:dynamic_fair}
\end{figure}

%The performance of the proposed method is evaluated. 
With time horizon $T=300$ and decreasing stepsize $\mu$, and constant stepsize $\alpha$ is shown in \figref{log:cost&regret}. In this case, LoTFair has similar performance to the instantaneous optimal approach and the value of dynamic regret oscillates around 0. It matches the analysis of dynamic regret (\textbf{Theorem 2}) as \eqref{theorem2}.

The performance of fairness is shown in \figref{log:dynamic_fair}. It can be observed that $\mathbf{\mathcal{F}}_T$ eventually becomes a sine wave centered at a constant value(around 3 in \figref{log:dynamic_fair}). The amplitude and frequency of the oscillation increase since we have decreasing stepsize $\mu$. It also shows that LoTFair allows instantaneous violations but ensures long-term fairness by adapting the decisions over the time horizon. The result matches the analysis of dynamic fair (\textbf{Theorem 1}) as \eqref{theorem1}.
\subsection{P2P Electricity Market}\label{experiments}
In this section, we test the proposed algorithm on the P2P electricity market application in the 14-bus grid system. 
% Each node is considered as a client in the market. Three settings are considered for generating supply and demand.

\noindent
\textbf{Datasets and Settings:}
The experiments are carried out on synthetic data and real data from California ISO Open Access Same-time Information System (OASIS).
14 nodes are divided into 2 groups by their locations in the power grids. In the synthetic data, we assume that the average trading price within the same group is a little cheaper than the average trading price between two groups.

\noindent\textbf{1. Random setting:} %The supply and demand of each client at every $t$ are randomly generated according to predefined random distributions.
% For the nodes that have the ability to generate electricity, the supply generated by solar power follows uniform distributions, the supply generated by wind power follows normal distributions. 
The supply and demand of each client at every $t$ are randomly generated according to predefined time-invariant distributions. The supply comes from two renewable energy sources: solar power and wind power. More specifically, solar power generation follows a normal distribution and wind power generation follows a uniform distribution.

\noindent\textbf{2. Time-varying setting:} The supply and demand of each client at every $t$ are generated as cosine functions, e.g., $s_t^i \coloneqq a^i+b^i cos(\pi t) + c_t$ where $a^i$ and $b^i$ are two constants for client $i$ and $c_t\sim \mathcal{N}(0,\sigma^{2})$ is a small noise centered at zero. This setting of supply and demand satisfies the assumptions in section \ref{analysis}. 

\noindent\textbf{3. OASIS:} The supply and demand of each client at every $t$ are pre-processed real data from OASIS. OASIS contains hourly total demand and supply. At each time $t$, based on the Dirichlet distribution, three distributions for demand, wind power supply and solar power supply are randomly generated. Then assign demand and supply to each node according to distributions.
% Although $g_t^{\kappa}(\mathbf{x}_t)$ can be revealed before the decisions are made, it might lead to huge increases in loss function or even cannot be satisfied if the constraint $g_t^{\kappa} \leq 0$ is very strict($\kappa$ is close to 0).\textbf{}
%The objective function attempts to minimize the total cost to purchase enough energy to satisfy the supply-demand balance for all consumers at time $t$, the constraints ensure the system stability and instantaneous fairness.
%\blue{Offline for dynamic regret}

\noindent
\textbf{Methods:} in each setting, we compare the results of the proposed method, LoTFair, and 2 baselines:
%\shen{change below to itemize environment, similar to classification}
%\indent\textbf{ \textbullet Long-term:} Solve \eqref{long_general} via Algorithm \ref{algo:two-side}. \\
%\indent\textbf{ \textbullet Instantaneous:} Solve the problem in \eqref{instantaneous_general} at each time slot: $\mathbf{x}_t =\arg \min_{\mathbf{x} \in \mathcal{X}_t} f_t(\mathbf{x}) ~~ \text{ s.t. }  g_t^{\kappa}(\mathbf{x}) \leq 0$. \\ % , \qquad \forall t \in \{1,\cdots,T \}$\\
%\indent\textbf{ \textbullet Offline:} Solve the problem in \eqref{long_general} offline.
\begin{itemize}[leftmargin=*]
    \item \textbf{Long-term:} LoTFair (Algorithm \ref{algo:two-side}).
    \item \textbf{Instantaneous:} $\mathbf{X}_t =\arg \min_{\mathbf{X} \in \mathcal{X}_t} f_t(\mathbf{X}) ~\text{ s.t.}  g_t^{\kappa}(\mathbf{X}) \leq 0$.
    \item \textbf{Offline:} Solve \eqref{long_general} offline.
\end{itemize}

\noindent\textbf{Experimental results} Below we will present the experimental results for different settings.
\subsubsection{\bf Random Setting}

\begin{figure*} 
    \centering
    \vspace{-4mm}
    \captionsetup{justification=centering}
    \subfigure[Time-averaged cost]{ \label{random:cost}
    \includegraphics[width=0.4\textwidth]{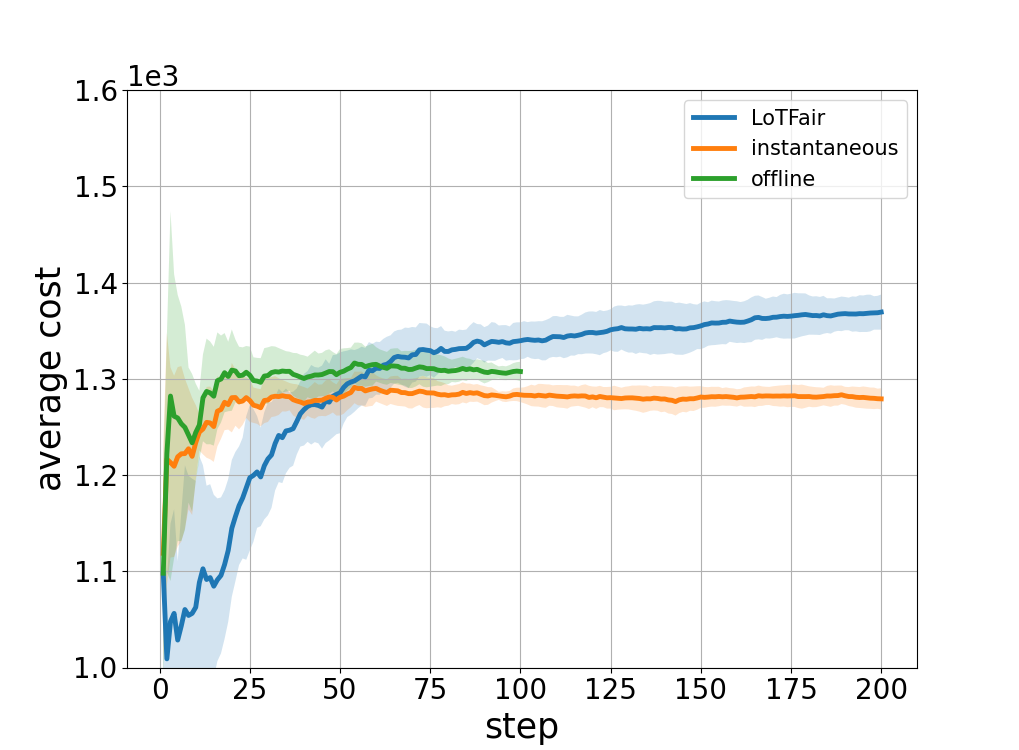}} %\columnwidth
    \quad\quad~~~~~~~~~
    \subfigure[Time-averaged dynamic regret]{ \label{random:regret}
    \includegraphics[width=0.4\textwidth]{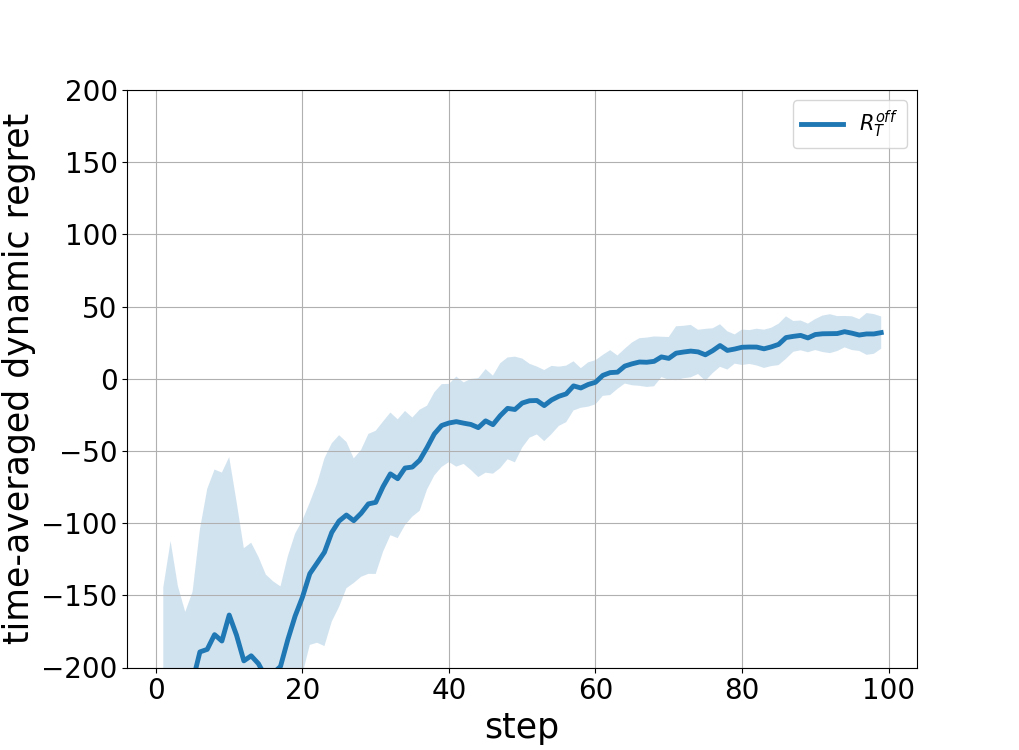}} % some mistakes
    \caption{Results of (a) time-averaged costs and (b) dynamic regrets on random setting.}\label{random:cost&regret} %$\mu=1e^4, \alpha = 0.5$.
\end{figure*}

The performance of the proposed method is evaluated on randomly generated synthetic data.
With time horizon $T=200$ and constant stepsizes, $\mu$ and $\alpha$, the results are shown as \figref{random:cost&regret} and \figref{random:dynamic_fair}. LoTFair can achieve both sub-linear regret and dynamic fairness on the long-term average. 
The average cost of LoTFair is slightly higher than the instantaneous solution. 
{The solution of the instantaneous approach benefits from the relaxation ($\kappa>0$) but then $\mathbf{\mathcal{F}}_T$ increases linearly as shown in \figref{random:dynamic_fair}.}
It is worth mentioning that the dynamic regret may not be sub-linear as \textit{Assumption 5} may not hold. However, that LoTFair still can achieve sublinear dynamic fairness with acceptable loss, shown as \figref{random:cost&regret} and \figref{random:dynamic_fair}. \Figref{random:dynamic_fair} also confirms that the instantaneous approach can not ensure fairness in the long term. %In \figref{random:dynamic_A2FV}, it shows that our method allows violations in each time slot but enforces the long-term fairness by adapting the decisions to the environment dynamics.

\iffalse
\begin{figure*} 
    \centering
    \captionsetup{justification=centering}
    \subfigure[$\mathbf{\mathcal{F}}_T$]{ \label{random:dynamic_fair}
    \includegraphics[width=0.33\textwidth]{figures/random_dynamic_fair.png}} 
    \quad
    \subfigure[$A^2FV$]{ \label{random:dynamic_A2FV}
    \includegraphics[width=0.33\textwidth]{figures/random_A2FV.png}} % some mistakes
    \caption{Experimental results of random setting, dynamic fairness and $A^2FV$, $\mu=1e^4, \alpha = 0.5$.}\label{random:dynamic_fair&A2FV}
\end{figure*}
\fi

\begin{figure}
    \centering
    \vspace{-4mm}\includegraphics[width=0.4\textwidth]{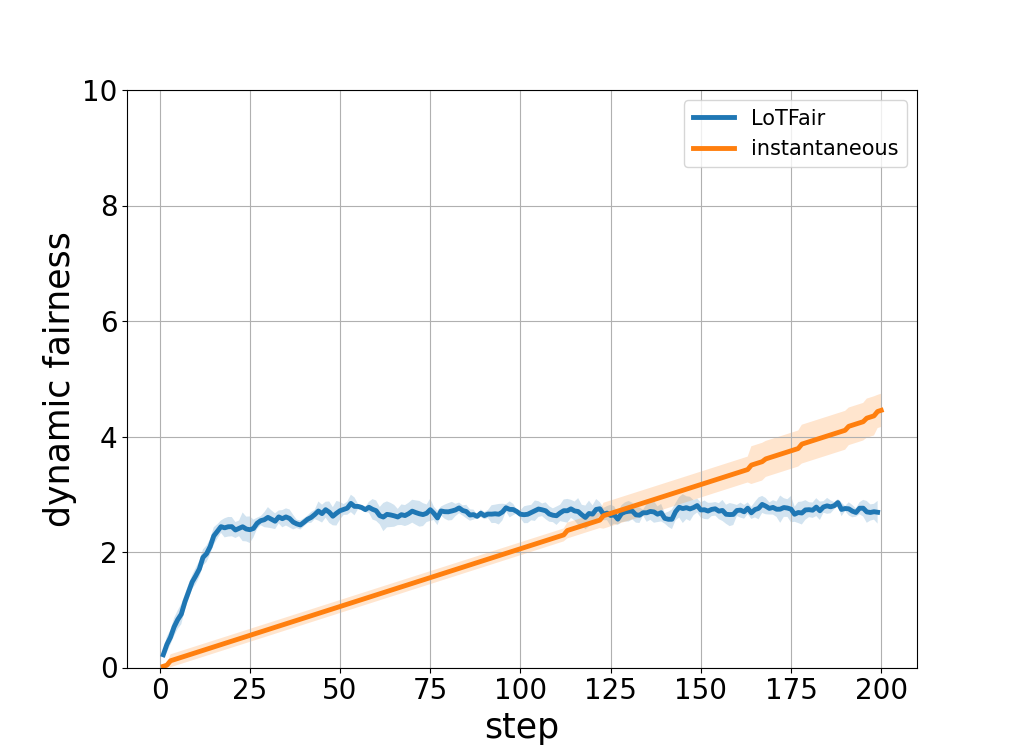}
    \caption{Dynamic fairness results on random setting.} % , $\mu=1e^4, \alpha = 0.5$
    \label{random:dynamic_fair}
\end{figure}

\subsubsection{\bf Time-varying Setting}
% Instantaneous method fails frequently with the fairness constraint when supply and demand are generated by cosine functions plus noise.
\begin{figure*} 
    \centering
    \vspace{-4mm}
    \captionsetup{justification=centering}
    \subfigure[Time-averaged Cost]{ \label{time-varying:cost}
    \includegraphics[width=0.4\textwidth]{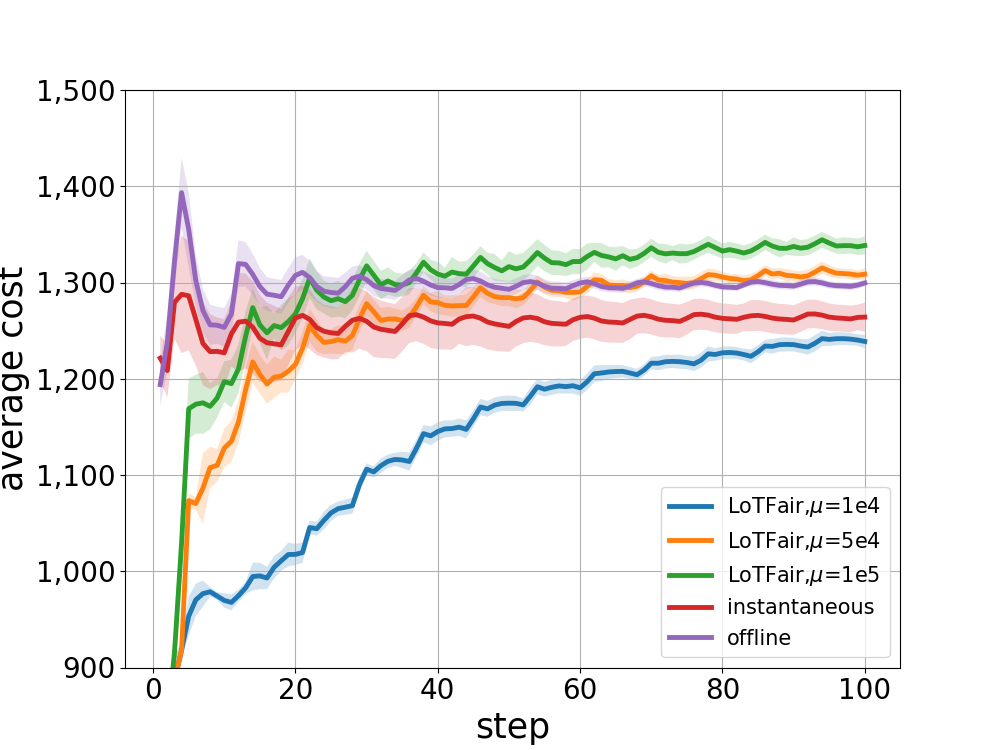}} 
    \quad\quad~~~~~~~~~
    \subfigure[Time-averaged dynamic regret($\mathbf{\mathcal{R}}^{\text{off}}_T$) ]{ \label{time-varying:regret}
    \includegraphics[width=0.4\textwidth]{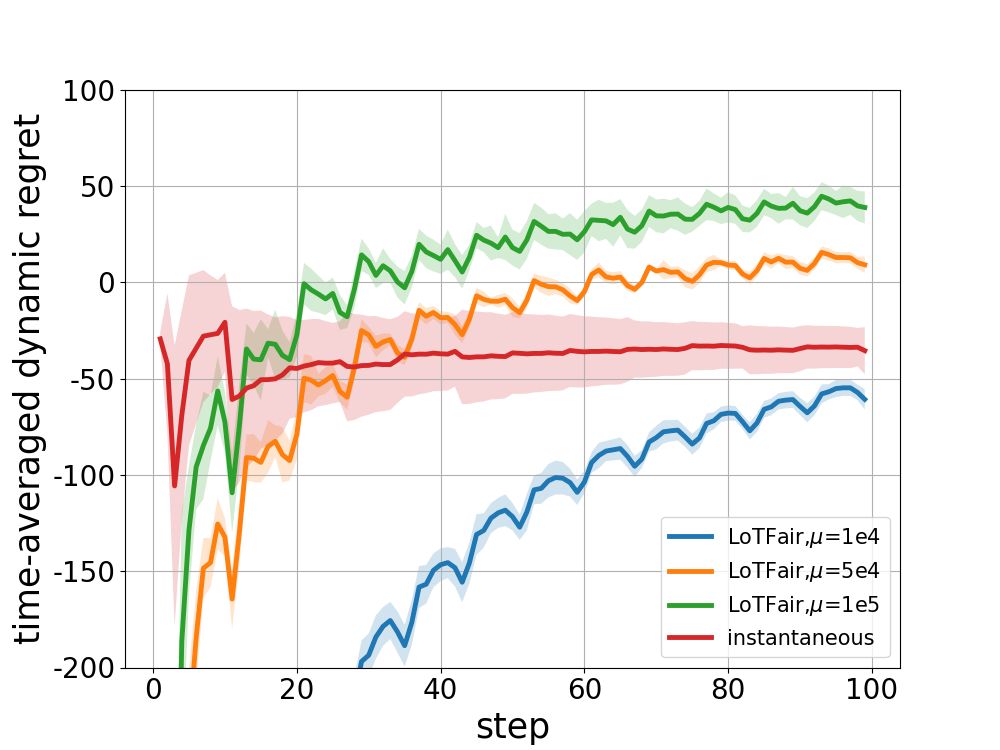}} % some mistakes
    \caption{Results of (a) time-averaged costs and (b) dynamic regrets on time-varying setting.} % , $\alpha = 0.5$
    \label{time-varying:cost&regret}
\end{figure*}
\iffalse
\begin{figure*} 
    \centering
    \captionsetup{justification=centering}
    \subfigure[$|\mathbf{\mathcal{F}}_T|$]{ \label{time-varying:dynamic_fair}
    \includegraphics[width=0.33\textwidth]{figures/TV_dynamic_fair.png}} 
    \quad
    \subfigure[$A^2FV$]{ \label{time-varying:A2FV}
    \includegraphics[width=0.33\textwidth]{figures/TV_A2FV.png}} % some mistakes
    \caption{Experimental results of time-varying setting dynamic fairness and $A^2FV$, $\alpha = 0.5$.}\label{time-varying:dynamic_fair&A2FV}
\end{figure*}
\fi

\begin{figure}
    \centering
    \vspace{-4mm}
    \captionsetup{justification=centering}
    \includegraphics[width=0.4\textwidth]{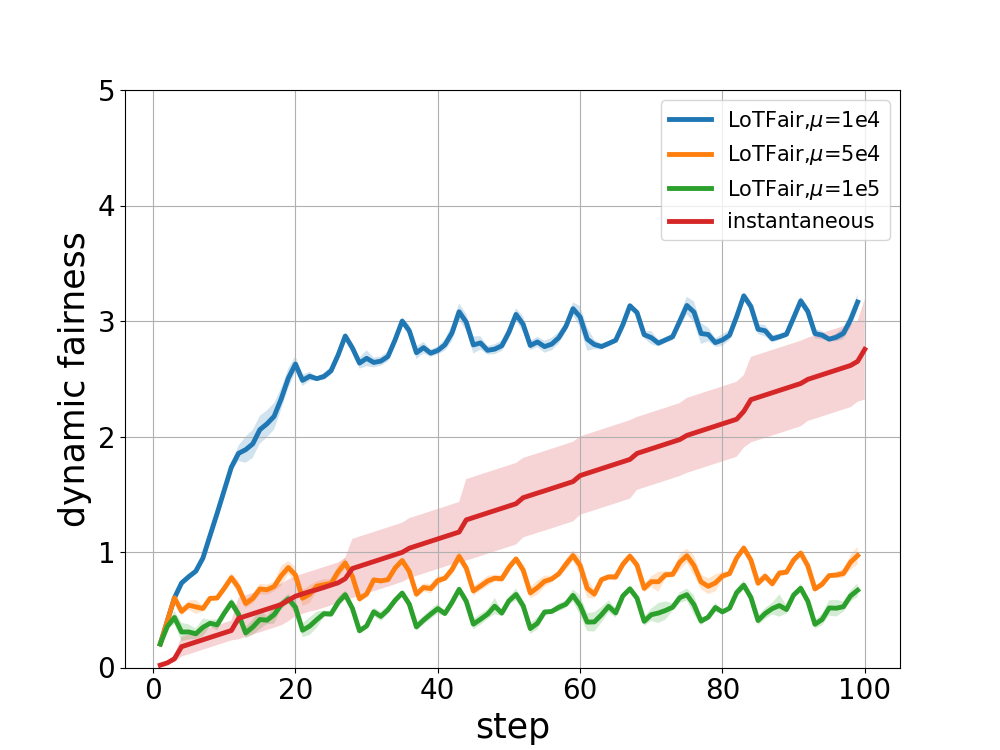}
    \caption{Dynamic fairness results on time-varying setting.} % , $\alpha = 0.5$
    \label{time-varying:dynamic_fair}
\end{figure}
In this section, we investigate how LoTFair performs when demand and supply are time-varying. %, e.g., $s_t^i \coloneqq a^i+b^i cos(\pi t) + c_t$ where $a^i$ and $b^i$ are two constants for client $i$ and $c_t\sim \mathcal{N}(0,\sigma^{2})$ is a small noise. This setting of supply and demand satisfies the assumptions in section \ref{analysis}. 
%{In this setting, the instantaneous approach that solves the problem (\ref{instantaneous_general}) fails frequently and usually cannot survive over 15 iterations.}
{In this setting, the instantaneous approach is infeasible in many instances.
To address this issue, we further relax the $g_t^{\kappa}(\mathbf{x}) \leq 0$ to $g_t^{\kappa}(\mathbf{x}) \leq \tau$, as (13h). }
The results for time horizon $T=100$ and constant step-sizes are shown in \figref{time-varying:cost&regret} and \figref{time-varying:dynamic_fair}. 
%It will cause more fairness violations, corresponding to small sharp increases of the ``instantaneous" curve in \figref{time-varying:dynamic_fair}.

%While increasing the value of $\mu$, the average cost increases and $|\mathbf{\mathcal{F}}_T|$ decreases. It can be concluded that the parameter $\mu$ controls the strength enforcing the long-term fairness. 
%If the stepsize is initialized with a small value and $\bm{\lambda}$ is initialized as $\mathbf{0}$, the cost could be better than the offline decisions in a short-term period due to the fair violations at the first few iterations. For better long-term fairness (smaller upper bound of $|\mathbf{\mathcal{F}}_T|$), the algorithm can sacrifice the utility by increasing dual stepsize. 
 %\blue{With fixed stepsize, $A^2FV$ also converges to a small value.}
{The average cost and $|\mathbf{\mathcal{F}}_T|$ increases and decreases respectively with $\mu$. Therefore, it follows that there is a trade-off in choosing the parameters $\mu$. 
Since the constraint is further relaxed in some time slots on the instantaneous approach, it can result in larger fairness violations, which is reflected in the small but sharp increases observed in \figref{time-varying:dynamic_fair}.}
The experimental results show that the sub-linear dynamic regret and sub-linear dynamic fairness are guaranteed in the long run with fixed stepsizes {while the dynamic fairness can increase linearly with the (relaxed) instantaneous approach.}
% The figures demonstrate the merits of our method: 
% \subsubsection{[Decreasing stepsize]} shown in next dataset
\subsubsection{\bf OASIS}
\begin{figure} 
    \centering
    \vspace{-4mm}
    \captionsetup{justification=centering}
    %\subfigure[Cost]{ 
    \includegraphics[width=0.4\textwidth]{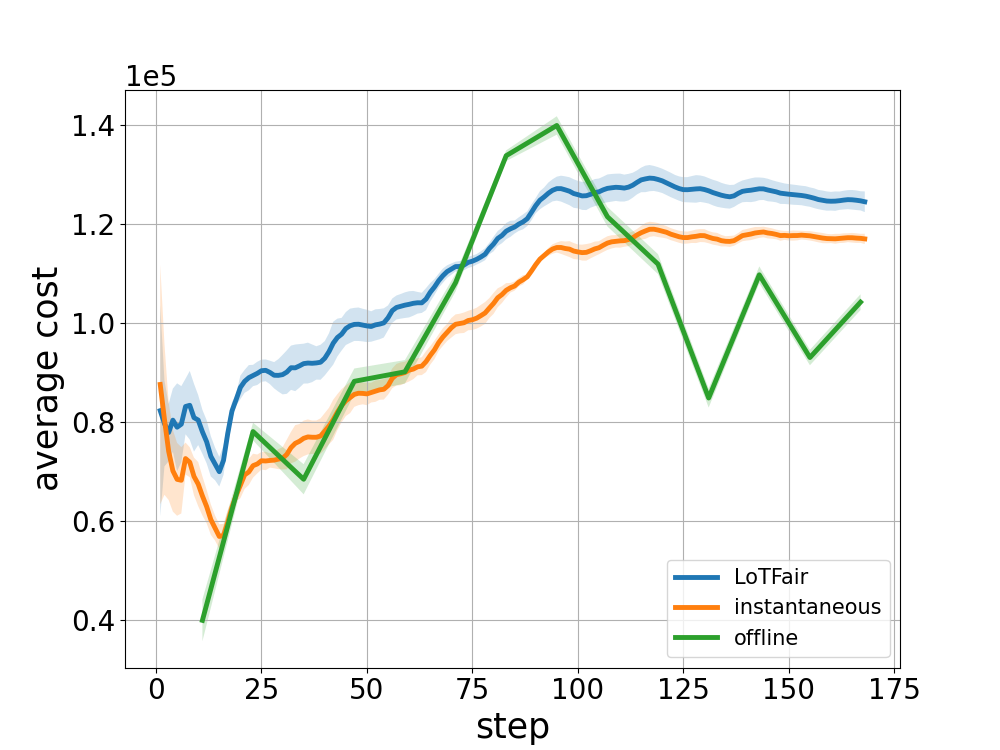} 
    %\quad
    %\subfigure[Regret]{ \label{OASIS:regret}
    %\includegraphics[width=4.5cm]{figures/OASIS_regret.png}} % some mistakes
    \caption{Results of time-averaged costs on OASIS dataset.}\label{OASIS:cost} %\label{OASIS:cost&regret}
\end{figure}

In addition to two synthetic data generated under 2 different settings, we also consider real-world data in simulating. We collect one week's demand, supply and price starting on 2023 January 1st at 7:00 am and ending on 2023 January 8th at 7:00 am from 14 nodes on California ISO Open Access Same-time Information System (OASIS). 14 nodes are randomly divided into 2 groups. The supply comes from 2 kinds of renewable energy sources(RES): solar power and wind power \cite{P2P-elec-distribution}. The data contains hourly information about the total demand, supply and unit price of each node buying energy. The total demand is much larger than the supply of RES since the energy market is still dominated by traditional energy, so we only count 15\% of it as the demand of the P2P electricity market. With the one-week time horizon and decreasing stepsizes, the results are shown as \figref{OASIS:cost} and \figref{OASIS:dynamic_fair}.
%Data from ''California ISO OASIS"

\iffalse
\begin{figure*} 
    \centering
    \captionsetup{justification=centering}
    \subfigure[$\mathbf{\mathcal{F}}_T$]{ \label{OASIS:dynamic_fair}
    \includegraphics[width=0.33\textwidth]{figures/OASIS_dynamic_fair.png}} 
    \quad
    \subfigure[$A^2FV$]{ \label{OASIS:A2FV}
    \includegraphics[width=0.33\textwidth]{figures/OASIS_A2FV.png}} % some mistakes
    \caption{Experimental results of OASIS, dynamic fairness and $A^2FV$. 
    % \shen{this figure does not look nice. Try to get a better one, e.g., by multiple runs}
    }\label{OASIS:dynamic_fair&A2FV}
\end{figure*}
\fi

\begin{figure}
    \centering
    \vspace{-4mm}
    \captionsetup{justification=centering}
    \includegraphics[width=0.4\textwidth]{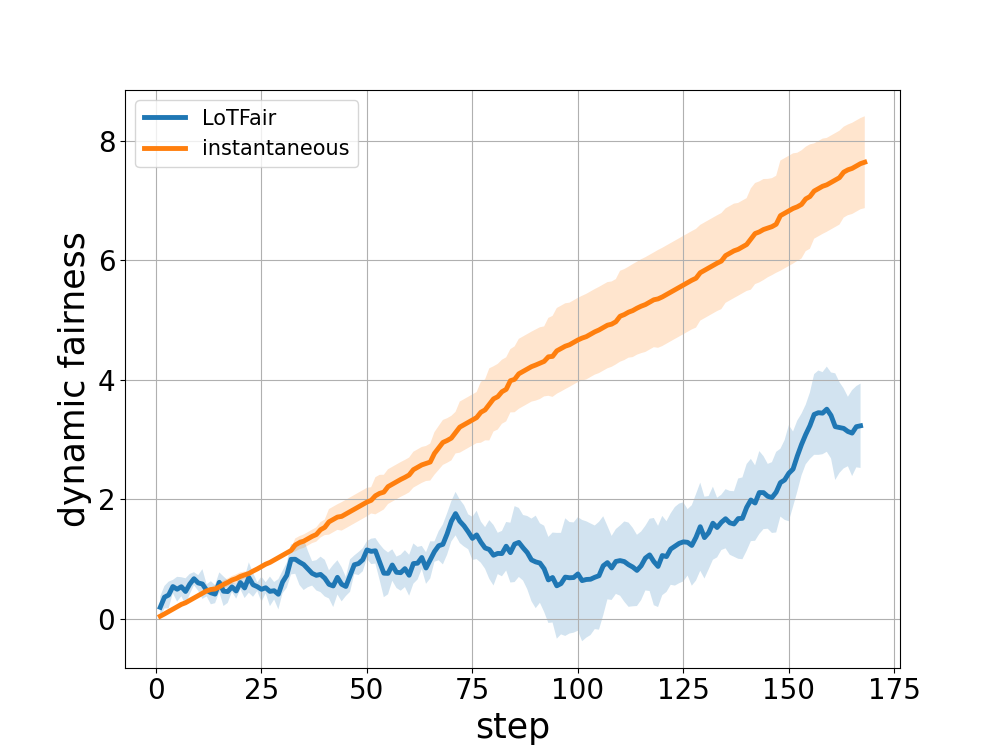}
    \caption{Dynamic fairness results on OASIS dataset.}
    \label{OASIS:dynamic_fair}
\end{figure}

% \blue{Rerunning with upper bounds 0.15, $\mu$ too small while t>120, cost decreases compared with last run. Tuning $\mu$ and $\alpha$.\\
% slightly increases the lower bound of $\mu$ does not change a lot.}.
{Since there is a huge decrease in wind power supply on the fourth day, it reflects as the sharp increase in the average cost in \figref{OASIS:cost}.}
%In \figref{OASIS:cost}, the sharp increase in average cost is due to a huge decrease in wind power supply during that period. 
It implies that even solving problem (\ref{long_general}) offline, the long-term fairness constraint, $\sum_{t=1}^T g_t(\mathbf{x}_t) = 0$, is still very restricted in some real scenarios. In order to achieve more fair results in dynamic decision-making systems with a lot of uncertainty, there is a growing need to understand long-term fairness due to its flexibility and applicability. 
In the \figref{OASIS:dynamic_fair}, $g_t(\mathbf{x}_t)$ oscillates around 0 in some periods {because the advantaged group is changing and could lead to arbitrary or even adversarial variations in the $g_t(\mathbf{x}_t)$.} Though the violation of each time slot is much huger than the solution of the instantaneous approach with $g^{\kappa}_t(\mathbf{x}_t)$, LoTFair still can achieve the sub-linear increase in dynamic fairness.

\section{Conclusion}
In practice, decision-making systems are usually operating in a dynamic manner and raising concerns about potential unfairness.
The sole pursuit of the overall performance may lead to unfair results.
To achieve fair results in decision making scenarios, such as energy management, there is a growing need to understand long-term fairness due to its flexibility and applicability. 
In this paper, we investigate the problem of long-term fairness for decision making where the overall utility is optimized in the presence of the long-term fairness constraint. 
The present paper proposes an online algorithm, LoTFair, that is proven to achieve sub-linear dynamic regret and sub-linear dynamic fairness. 
The experimental results on the classification task on the `Adult' dataset and the energy management case show that the proposed method could indeed guarantee long-term fairness without significant performance loss.



\bibliographystyle{IEEEtran}
\bibliography{main}

\begin{thebibliography}{10}
\providecommand{\url}[1]{#1}
\csname url@samestyle\endcsname
\providecommand{\newblock}{\relax}
\providecommand{\bibinfo}[2]{#2}
\providecommand{\BIBentrySTDinterwordspacing}{\spaceskip=0pt\relax}
\providecommand{\BIBentryALTinterwordstretchfactor}{4}
\providecommand{\BIBentryALTinterwordspacing}{\spaceskip=\fontdimen2\font plus
\BIBentryALTinterwordstretchfactor\fontdimen3\font minus \fontdimen4\font\relax}
\providecommand{\BIBforeignlanguage}[2]{{%
\expandafter\ifx\csname l@#1\endcsname\relax
\typeout{** WARNING: IEEEtran.bst: No hyphenation pattern has been}%
\typeout{** loaded for the language `#1'. Using the pattern for}%
\typeout{** the default language instead.}%
\else
\language=\csname l@#1\endcsname
\fi
#2}}
\providecommand{\BIBdecl}{\relax}
\BIBdecl

\bibitem{fair_loan}
E.~L. Lee, J.-K. Lou, W.-M. Chen, Y.-C. Chen, S.-D. Lin, Y.-S. Chiang, and K.-T. Chen, ``Fairness-aware loan recommendation for microfinance services,'' in \emph{Proceedings of the International Conference on Social Computing}, 2014, p. 1–4.

\bibitem{fair_college}
S.~Bird, K.~Kenthapadi, E.~Kiciman, and M.~Mitchell, ``Fairness-aware machine learning: Practical challenges and lessons learned,'' in \emph{Proceedings of the ACM International Conference on Web Search and Data Mining}, 2019, p. 834–835.

\bibitem{Yan_Howe_2020}
A.~Yan and B.~Howe, ``Fairness-aware demand prediction for new mobility,'' \emph{Proceedings of the AAAI Conference on Artificial Intelligence}, vol.~34, no.~01, pp. 1079--1087, Apr. 2020.

\bibitem{long-fair-dynamic-resource-allocation}
T.~Si~Salem, G.~Iosifidis, and G.~Neglia, ``Enabling long-term fairness in dynamic resource allocation,'' \emph{Proceedings of the ACM on Measurement and Analysis of Computing Systems}, vol.~6, no.~3, pp. 1--36, 2022.

\bibitem{guo2020optimal}
Z.~Guo, W.~Wei, L.~Chen, Z.~Wang, J.~P. Catal{\~a}o, and S.~Mei, ``Optimal energy management of a residential prosumer: A robust data-driven dynamic programming approach,'' \emph{IEEE Systems Journal}, vol.~16, no.~1, pp. 1548--1557, 2020.

\bibitem{guo2022optimisation}
Z.~Guo, W.~Wei, M.~Shahidehpour, Z.~Wang, and S.~Mei, ``Optimisation methods for dispatch and control of energy storage with renewable integration,'' \emph{IET Smart Grid}, vol.~5, no.~3, pp. 137--160, 2022.

\bibitem{corbett2017algorithmic}
S.~Corbett-Davies, E.~Pierson, A.~Feller, S.~Goel, and A.~Huq, ``Algorithmic decision making and the cost of fairness,'' in \emph{Proceedings of the 23rd acm sigkdd international conference on knowledge discovery and data mining}, 2017, pp. 797--806.

\bibitem{zhang2021fairness}
X.~Zhang and M.~Liu, ``Fairness in learning-based sequential decision algorithms: A survey,'' in \emph{Handbook of Reinforcement Learning and Control}.\hskip 1em plus 0.5em minus 0.4em\relax Springer, 2021, pp. 525--555.

\bibitem{fairness-is-not-static}
A.~D'Amour, H.~Srinivasan, J.~Atwood, P.~Baljekar, D.~Sculley, and Y.~Halpern, ``Fairness is not static: deeper understanding of long term fairness via simulation studies,'' in \emph{Proceedings of the 2020 Conference on Fairness, Accountability, and Transparency}, 2020, pp. 525--534.

\bibitem{hu2022achieving}
Y.~Hu and L.~Zhang, ``Achieving long-term fairness in sequential decision making,'' in \emph{Proceedings of the AAAI Conference on Artificial Intelligence}, vol.~36, no.~9, 2022, pp. 9549--9557.

\bibitem{weber2022enforcing}
A.~Weber, B.~Metevier, Y.~Brun, P.~S. Thomas, and B.~C. da~Silva, ``Enforcing delayed-impact fairness guarantees,'' 2022.

\bibitem{yin2023long}
T.~Yin, R.~Raab, M.~Liu, and Y.~Liu, ``Long-term fairness with unknown dynamics,'' \emph{arXiv preprint arXiv:2304.09362}, 2023.

\bibitem{P2P-energy-transaction-mechanisms}
E.~Oh and S.-Y. Son, ``Peer-to-peer energy transaction mechanisms considering fairness in smart energy communities,'' \emph{IEEE Access}, vol.~8, 2020.

\bibitem{robert2020icis}
L.~Robert, G.~Bansal, and C.~Lutge, ``Icis 2019 sighci workshop panel report: Human computer interaction challenges and opportunities for fair, trustworthy and ethical artificial intelligence,'' 2020.

\bibitem{dwork2012fairness}
C.~Dwork, M.~Hardt, T.~Pitassi, O.~Reingold, and R.~Zemel, ``Fairness through awareness,'' in \emph{Proceedings of the 3rd innovations in theoretical computer science conference}, 2012, pp. 214--226.

\bibitem{de2005tutorial}
P.-T. De~Boer, D.~P. Kroese, S.~Mannor, and R.~Y. Rubinstein, ``A tutorial on the cross-entropy method,'' \emph{Annals of operations research}, vol. 134, pp. 19--67, 2005.

\bibitem{mannor2005cross}
S.~Mannor, D.~Peleg, and R.~Rubinstein, ``The cross entropy method for classification,'' in \emph{Proceedings of the 22nd international conference on Machine learning}, 2005, pp. 561--568.

\bibitem{chuang2021fair}
C.-Y. Chuang and Y.~Mroueh, ``Fair mixup: Fairness via interpolation,'' \emph{arXiv preprint arXiv:2103.06503}, 2021.

\bibitem{han2023retiring}
X.~Han, Z.~Jiang, H.~Jin, Z.~Liu, N.~Zou, Q.~Wang, and X.~Hu, ``Retiring {$\Delta$DP} : New distribution-level metrics for demographic parity,'' \emph{Transactions on Machine Learning Research}, 2023.

\bibitem{PARK20173}
C.~Park and T.~Yong, ``Comparative review and discussion on p2p electricity trading,'' \emph{Energy Procedia}, vol. 128, pp. 3--9, 2017, international Scientific Conference “Environmental and Climate Technologies”, CONECT 2017, 10-12 May 2017, Riga, Latvia.

\bibitem{P2P-CBM-reveiw}
T.~Sousa, T.~Soares, P.~Pinson, F.~Moret, T.~Baroche, and E.~Sorin, ``Peer-to-peer and community-based markets: A comprehensive review,'' \emph{Renewable and Sustainable Energy Reviews}, vol. 104, pp. 367--378, 2019.

\bibitem{P2P-elec-distribution}
Y.~Liu, L.~Wu, and J.~Li, ``Peer-to-peer (p2p) electricity trading in distribution systems of the future,'' \emph{The Electricity Journal}, vol.~32, no.~4, pp. 2--6, 2019, special Issue on Strategies for a sustainable, reliable and resilient grid.

\bibitem{mokhtari2016online}
A.~Mokhtari, S.~Shahrampour, A.~Jadbabaie, and A.~Ribeiro, ``Online optimization in dynamic environments: Improved regret rates for strongly convex problems,'' in \emph{2016 IEEE 55th Conference on Decision and Control (CDC)}.\hskip 1em plus 0.5em minus 0.4em\relax IEEE, 2016, pp. 7195--7201.

\bibitem{chen2017online}
T.~Chen, Q.~Ling, and G.~B. Giannakis, ``An online convex optimization approach to proactive network resource allocation,'' \emph{IEEE Transactions on Signal Processing}, vol.~65, no.~24, pp. 6350--6364, 2017.

\bibitem{nesterov2013introductory}
Y.~Nesterov, \emph{Introductory Lectures on Convex Optimization: A Basic Course}.\hskip 1em plus 0.5em minus 0.4em\relax Springer Science \& Business Media, 2013, vol.~87.

\bibitem{liu2004role}
M.~Liu and G.~Gross, ``Role of distribution factors in congestion revenue rights applications,'' \emph{IEEE Transactions on Power Systems}, vol.~19, no.~2, pp. 802--810, 2004.

\bibitem{8782819}
M.~Khorasany, Y.~Mishra, and G.~Ledwich, ``A decentralized bilateral energy trading system for peer-to-peer electricity markets,'' \emph{IEEE Transactions on Industrial Electronics}, vol.~67, no.~6, pp. 4646--4657, 2020.

\bibitem{9102274}
A.~Paudel, L.~P. M.~I. Sampath, J.~Yang, and H.~B. Gooi, ``Peer-to-peer energy trading in smart grid considering power losses and network fees,'' \emph{IEEE Transactions on Smart Grid}, vol.~11, no.~6, pp. 4727--4737, 2020.

\end{thebibliography}
\noindent
\begin{appendices}
    \appendices

\iffalse
\textit{Assumption 1}: For every $t$, the cost function $f_t(\mathbf{x})$ and the time-varying fairness constraint $g_t(\mathbf{x})$ are convex.\\
\textit{Assumption 2}: For every $t$, $f_t(\mathbf{x})$ has bounded gradient on $\mathcal{X}$, $|| \nabla f_t(\mathbf{x}) ||\leq G, \forall \mathbf{x} \in \mathcal{X}$; and ${g}_t(\mathbf{x})$ is bounded, $|g_t(\mathbf{x})|\leq M$. \\
\textit{Assumption 3}: The radius of the convex feasible set $\mathcal{X}$ is bounded, $||\mathbf{x}-\mathbf{y}||\leq R,\forall \mathbf{x},\mathbf{y}\in \mathcal{X}$. \\% \blue{The radius of $g_t(\mathbf{x})$ is bounded, $||g_t(\mathbf{x})-g_t(y) || \leq S$.}\\
\textit{Assumption 4}: There exists a constant $\epsilon>0$, and two interior points $\mathbf{\tilde{\mathbf{x}}}_t$ and $\tilde{\tilde{\mathbf{x}}}_t$ such that $g_t(\tilde{\mathbf{x}}_t) \leq -\epsilon$ and $g_t(\tilde{\tilde{\mathbf{x}}}_t) \geq \epsilon,\forall t$. \\
\textit{Assumption 5}: The constant $\epsilon$ in Assumption 4 satisfies $\epsilon > \bar{V}(g)$, $\bar{V}(g) = \max_{t\in T} \max_{\mathbf{x} \in \mathcal{X}} || g_{t+1}(\mathbf{x}) - g_t(\mathbf{x}) ||$.
\fi

\section{Proof of Theorem 1} \label{proof_theorem1}
Before moving on to proof, we first state the following 2 Lemmas which will be served as the stepstone for the proof of Theorem 1.

\textbf{Lemma 1 }
The difference between the dual variables of two successive iterates $|\lambda_{1,t+1} - \lambda_{2,t+1}|$ can be upper bounded by 
\begin{align}
    |\lambda_{1,t+1} - \lambda_{2,t+1}| \leq   \bar{\lambda} := 2 \mu M + \frac{2GR+\frac{R^2}{2\alpha} + \mu M^2}{\epsilon-\bar{V}(g)} \label{lambda_upperbound}%\\ | \lambda_{1,t+1}|
    %| \lambda_{2,t+1}| \leq  \bar{\lambda}_2 \leq \bar{\lambda} = \mu M + \frac{2GR+\frac{R^2}{2\alpha} + \mu M^2}{\epsilon-\bar{V}(g)} \label{lambda2_upperbound}
\end{align}
{Proof: Appendeix \ref{appendix_lemma1}}

\textbf{Lemma 2: } Let $\bar{\delta} := \max_{t \in \{ 1,\cdots,T \}} |\lambda_{1,t+1} - \lambda_{2,t+1}|$, the following equality holds
\begin{align}
     \bar{\delta}  = \max(\bar{\lambda}_1, \bar{\lambda}_2) \label{lemma2}
\end{align}
which implies that $\bar{\lambda}_1\leq \bar{\delta} \leq \bar{\lambda}$, and $ \bar{\lambda}_2 \leq \bar{\delta} \leq \bar{\lambda}$. 
$\bar{\lambda}_1,\bar{\lambda}_2$ denote the max value of $\lambda_{1,t},\lambda_{2,t}$ across the whole time horizon, i.e., $\bar{\lambda}_1 \coloneqq \max_{t \in \{ 1,\cdots,T \}} \lambda_{1,t}  $, $\bar{\lambda}_2 \coloneqq \max_{t \in \{ 1,\cdots,T \}} \lambda_{2,t}  $.
\\
{Proof: Appendeix \ref{appendix_lemma2}}
\\

Using the recursion in (\ref{1_update}) and (\ref{2_update}), we have 
\begin{align}
\lambda_{1,T+1}\geq \lambda_{1,T} + \mu g_T(\mathbf{x}_T) \geq \lambda_{1,1} + \sum_{t=1}^\top \mu g_t(\mathbf{x}_t); \\
\lambda_{2,T+1}\geq \lambda_{2,T} - \mu g_T(\mathbf{x}_T) \geq \lambda_{2,1} - \sum_{t=1}^T \mu g_t(\mathbf{x}_t).
\end{align}
\\
Since $\bar{\lambda}_1,\bar{\lambda}_2$ denote the max value of $\lambda_{1,t},\lambda_{2,t}$ across the whole time horizon, i.e., $\bar{\lambda}_1 \coloneqq \max_{t \in \{ 1,\cdots,T \}} \lambda_{1,t}  $, $\bar{\lambda}_2 \coloneqq \max_{t \in \{ 1,\cdots,T \}} \lambda_{2,t}  $, the following inequalities hold %\blue{define $\bar{\lambda}_1$}

%\blue{Using recursion:$\sum_{t=1}^\top \mu g_t(\mathbf{x}_t) \leq 2( \lambda_{1,T+1} -\lambda_{2,T+1} )$} 
\begin{align}
    &\sum_{t=1}^T g_t(\mathbf{x}_t)\leq \frac{\lambda_{1,T+1}}{\mu} - \frac{\lambda_{1,1}}{\mu} \leq \frac{\lambda_{1,T+1}}{\mu} \leq \frac{\bar{\lambda}_1}{\mu} \label{upper_Ft}\\
    &\sum_{t=1}^T g_t(\mathbf{x}_t) \geq -\frac{\lambda_{2,T+1}}{\mu} + \frac{\lambda_{2,1}}{\mu} \geq -\frac{\lambda_{2,T+1}}{\mu} \geq - \frac{\bar{\lambda}_2}{\mu}. \label{lower_Ft}
\end{align}
Furthermore, from \textbf{Lemma 2}, we have $\bar{\lambda}_1 \leq \bar{\lambda}$ and $\bar{\lambda}_2 \leq \bar{\lambda}$. In addition, based on \textbf{Lemma 1}, $ \bar{\lambda} := 2 \mu M + \frac{2GR+\frac{R^2}{2\alpha} + \mu M^2}{\epsilon-\bar{V}(g)}$.
Substituting $\bar{\lambda}$ into \eqref{upper_Ft} and \eqref{lower_Ft} leads to

\begin{align}
    %\sum_{t=1}^T g_t(\mathbf{x}_t) \leq \frac{||\bar{\lambda}_1||}{\mu} = M + \frac{\frac{2GR}{\mu} + \frac{{R^2}{2\alpha\mu}+M^2}{\epsilon-\bar{V}(g)}}
    \sum_{t=1}^T g_t(\mathbf{x}_t) &\leq \frac{\bar{\lambda}_1}{\mu} \leq \frac{\bar{\lambda}}{\mu} = 2M + \frac{\frac{2GR}{\mu}+\frac{R^2}{2\mu \alpha} + M^2}{\epsilon-\bar{V}(g)}\\
    \sum_{t=1}^T g_t(\mathbf{x}_t) &\geq -\frac{\bar{\lambda}_2}{\mu} \geq -\frac{\bar{\lambda}}{\mu} = -2M - \frac{\frac{2GR}{\mu}+\frac{R^2}{2\mu \alpha} + M^2}{\epsilon-\bar{V}(g)}
\end{align}
$\mathbf{\mathcal{F}}_T= \sum_{t=1}^T g_t(\mathbf{x}_t)$, the proof is completed: $-2M - \frac{\frac{2GR}{\mu}+\frac{R^2}{2\mu \alpha} + M^2}{\epsilon-\bar{V}(g)} \leq \mathbf{\mathcal{F}}_T  \leq 2M + \frac{\frac{2GR}{\mu}+\frac{R^2}{2\mu \alpha} + M^2}{\epsilon-\bar{V}(g)}$.

%\noindent
%\textbf{part 2: negative violation}

\section{Proof of Theorem 2} \label{proof_theorem2}
$h_t(\mathbf{x})= \nabla f_{t}(\mathbf{x}_{t})^\top(\mathbf{x}-\mathbf{x}_{t}) + \bm{\lambda}_{t+1}^\top \mathbf{\bar{g}}_{t}(\mathbf{x}) + \frac{||\mathbf{x}-\mathbf{x}_{t} ||^2}{2\alpha}$ is $1/ \alpha$-strongly convex \cite{nesterov2013introductory}, which implies that 
\begin{align*}
    h_t(\mathbf{y})-h_t(\mathbf{x}) \geq \nabla h_{t}(\mathbf{x})^\top (\mathbf{y-x}) + \frac{1}{2\alpha} || \mathbf{y-x}||^2 
\end{align*}
Since $\mathbf{x}_{t+1}$ is the minimizer of the problem $\min_{\mathbf{x}} h_t(\mathbf{x})$:
\begin{align*}
    \nabla h_t(\mathbf{x}_{t+1})^\top (\mathbf{y}-\mathbf{x}_{t+1}) \geq 0, \forall \mathbf{y}
\end{align*}
Set $\mathbf{y}=\mathbf{x}_t^*$ which is the optimal solution at time $t$ as \eqref{dynamic_x*}, $ \mathbf{x}_t^* = \arg \min_{\mathbf{x}_t \in \mathcal{X}_t} f_t(\mathbf{x}_t) ~~ \text{ s.t. } g_t(\mathbf{x}_t) = 0$. Then it leads to
\begin{align}
    h_t(\mathbf{x}_t^*) \geq h_t(\mathbf{x}_{t+1}) + \frac{1}{2\alpha} ||\mathbf{x}_t^*-\mathbf{x}_{t+1} ||^2 \notag
\end{align}
which is equivalent to
\begin{align}
     &\nabla f_{t}(\mathbf{x}_{t})^\top(\mathbf{x}_{t+1}-\mathbf{x}_{t}) + \bm{\lambda}_{t+1}^\top \mathbf{\bar{g}}_{t}(\mathbf{x}_{t+1}) + \frac{||\mathbf{x}_{t+1}-\mathbf{x}_{t} ||^2}{2\alpha} \notag \\
     \leq &\nabla f_{t}(\mathbf{x}_{t})^\top(\mathbf{x}_t^*-\mathbf{x}_{t}) + \bm{\lambda}_{t+1}^\top \mathbf{\bar{g}}_{t}(\mathbf{x}_t^*) \nonumber \\ 
     & + \frac{||\mathbf{x}_t^*-\mathbf{x}_{t} ||^2}{2\alpha} -\frac{||\mathbf{x}_t^*-\mathbf{x}_{t+1} ||^2}{2 \alpha}
\end{align}
Adding $f_t(\mathbf{x}_t)$ on both sides
\begin{align}
    &f_t(\mathbf{x}_t) + \nabla f_{t}(\mathbf{x}_{t})^\top(\mathbf{x}_{t+1}-\mathbf{x}_{t}) \nonumber\\
    &+ \bm{\lambda}_{t+1}^\top \mathbf{\bar{g}}_{t}(\mathbf{x}_{t+1}) + \frac{||\mathbf{x}_{t+1}-\mathbf{x}_{t} ||^2}{2\alpha} \notag \\
    \leq & f_t(\mathbf{x}_t) +\nabla f_{t}(\mathbf{x}_{t})^\top(\mathbf{x}_t^*-\mathbf{x}_{t}) + \bm{\lambda}_{t+1}^\top \mathbf{\bar{g}}_{t}(\mathbf{x}_t^*) \nonumber \\ 
    &+ \frac{||\mathbf{x}_t^*-\mathbf{x}_{t} ||^2}{2\alpha} -\frac{||\mathbf{x}_t^*-\mathbf{x}_{t+1} ||^2}{2 \alpha} \notag
    \\
    \stackrel{(a)}{\leq} & f_t(\mathbf{x}_t^*) + \bm{\lambda}_{t+1}^\top \mathbf{\bar{g}}_t(\mathbf{x}_t^*) + \frac{||\mathbf{x}_t^*-\mathbf{x}_{t} ||^2}{2\alpha} -\frac{||\mathbf{x}_t^*-\mathbf{x}_{t+1} ||^2}{2 \alpha} \notag 
    \\
    \stackrel{(b)}{\leq} & f_t(\mathbf{x}_t^*) + \frac{||\mathbf{x}_t^*-\mathbf{x}_{t} ||^2}{2\alpha} -\frac{||\mathbf{x}_t^*-\mathbf{x}_{t+1} ||^2}{2 \alpha} \label{add_ft}
\end{align}
%\blue{explain}\\
where $(a)$ is due to the convexity of $f_t(\mathbf{x}_t)$ and $(b)$ comes from $g_t(\mathbf{x}_t^*)=0$.
Meanwhile, the term $\nabla f_{t}(\mathbf{x}_{t})^\top(\mathbf{x}_{t+1}-\mathbf{x}_{t})$ by 
\begin{align}
    - \nabla f_{t}(\mathbf{x}_{t})^\top(\mathbf{x}_{t+1}-\mathbf{x}_{t}) &\leq || \nabla f_{t}(\mathbf{x}_{t}) || ||\mathbf{x}_{t+1}-\mathbf{x}_{t} || \notag\\
    &\leq \frac{||\nabla f_{t}(\mathbf{x}_{t})||^2}{2\eta} + \frac{\eta}{2} ||\mathbf{x}_{t+1}-\mathbf{x}_{t} ||^2  \notag\\
    & \stackrel{(c)}{\leq} \frac{G^2}{2\eta} + \frac{\eta}{2} ||\mathbf{x}_{t+1}-\mathbf{x}_{t} ||^2 \label{bound-nablaft(x-x)}
\end{align}
where $\eta$ is an arbitrary positive constant, $(c)$ holds due to assumption 2.
Combining \eqref{add_ft} and \eqref{bound-nablaft(x-x)} and rearranging the terms, we can have
\begin{align}
    & f_t(\mathbf{x}_t) + \bm{\lambda}_{t+1}^\top \mathbf{g}_t(\mathbf{x}_{t+1}) \nonumber \\ 
    \leq & f_t(\mathbf{x}_t^*) + (\frac{\eta}{2} + \frac{1}{2\alpha}) ||\mathbf{x}_{t+1}-\mathbf{x}_{t} ||^2 \notag \\
    & + \frac{1}{2\alpha} (||\mathbf{x}_t^*-\mathbf{x}_{t} ||^2 - ||\mathbf{x}_t^*-\mathbf{x}_{t+1} ||^2) + \frac{G^2}{2\eta}
\end{align}
Setting $\eta =\frac{1}{\alpha}$, we have $\frac{\eta}{2}-\frac{1}{2\alpha}=0$, hence
\begin{align}
     &f_t(\mathbf{x}_t) + \bm{\lambda}_{t+1}^\top \mathbf{g}_t(\mathbf{x}_{t+1}) \nonumber \\ 
     \leq& f_t(\mathbf{x}_t^*) + \frac{1}{2\alpha} (||\mathbf{x}_t^*-\mathbf{x}_{t} ||^2 - ||\mathbf{x}_t^*-\mathbf{x}_{t+1} ||^2) + \frac{\alpha G^2}{2}
\end{align}
Based on $\Delta(\bm{\lambda}_{t}) \leq \mu \bm{\lambda}_t^\top \mathbf{\bar{g}}_t(\mathbf{x}_t) + \mu^2 |g_t(\mathbf{x}_t)|^2$ (\ref{delta_lambda}):
\begin{align}
    &\frac{\Delta(\bm{\lambda}_{t+1})}{\mu} + f_t(\mathbf{x}_t) \nonumber \\ 
    \leq & f_t(\mathbf{x}_t) + \bm{\lambda}_{t+1}^\top \mathbf{\bar{g}}_{t+1}(\mathbf{x}_{t+1}) + \bm{\lambda}_{t+1}^\top \mathbf{\bar{g}}_{t}(\mathbf{x}_{t+1})  \nonumber \\ 
    & -\bm{\lambda}_{t+1}^\top \mathbf{\bar{g}}_{t}(\mathbf{x}_{t+1}) + \mu^2 |g_t(\mathbf{x}_t)|^2 \notag\\
    \leq & f_t(\mathbf{x}_t^*) + \frac{1}{2\alpha}(||\mathbf{x}_t^*-\mathbf{x}_{t} ||^2 - ||\mathbf{x}_t^*-\mathbf{x}_{t+1} ||^2) +  \notag \\
    & \bm{\lambda}_{t+1}^\top(\mathbf{\bar{g}}_{t+1}(\mathbf{x}_{t+1})- \mathbf{\bar{g}}_{t}(\mathbf{x}_{t+1})) + \mu^2 |g_t(\mathbf{x}_t)|^2 + \frac{\alpha G^2}{2} \notag \\
    \stackrel{(a)}{\leq} & f_t(\mathbf{x}_t^*) + \frac{1}{2\alpha}(||\mathbf{x}_t^*-\mathbf{x}_{t} ||^2 - ||\mathbf{x}_t^*-\mathbf{x}_{t+1} ||^2) \nonumber \\ 
    & + |{\lambda}_{1,t+1} - {\lambda}_{2,t+1}|{V}(g) + \mu M^2 + \frac{\alpha G^2}{2} \notag\\
    \stackrel{(b)}{\leq} & f_t(\mathbf{x}_t^*) + \frac{1}{2\alpha}(||\mathbf{x}_t^*-\mathbf{x}_{t} ||^2 - ||\mathbf{x}_t^*-\mathbf{x}_{t+1} ||^2) \nonumber \\ 
    & + |\bar{\lambda}| {V}(g) + \mu M^2 + \frac{\alpha G^2}{2} \label{intermediate_step1}
\end{align}
where ${V}(g)=\max(0,g_{t+1}(\mathbf{x}_{t+1}) - g_{t}(\mathbf{x}_{t+1}) )$, $(a)$ holds because of \eqref{lambda_g_all}, $(b)$ is true due to Lemma 1. % $|\lambda_{1,t+1} - \lambda_{2,t+1}| \leq \bar{\lambda}$.
By adding and subtracting $||\mathbf{x}_t -\mathbf{x}_{t-1}^* ||^2$ in the $||\mathbf{x}_t^*-\mathbf{x}_{t} ||^2 - ||\mathbf{x}_t^*-\mathbf{x}_{t+1} ||^2$, we have that
\begin{align}
    & ||\mathbf{x}_t^*-\mathbf{x}_{t} ||^2 - ||\mathbf{x}_t^*-\mathbf{x}_{t+1} ||^2 \nonumber \\ 
    = &||\mathbf{x}_t^*-\mathbf{x}_{t} ||^2  - ||\mathbf{x}_t -\mathbf{x}_{t-1}^* ||^2 \nonumber \\ 
    & +  ||\mathbf{x}_t -\mathbf{x}_{t-1}^* ||^2  - ||\mathbf{x}_t^*-\mathbf{x}_{t+1} ||^2 \notag \\
    = & ||\mathbf{x}_t^*-\mathbf{x}_{t-1}^* || ||\mathbf{x}_t^* - 2 \mathbf{x}_t \nonumber \\ 
    & + \mathbf{x}_{t-1}^*|| + ||\mathbf{x}_t -\mathbf{x}_{t-1}^* ||^2 - ||\mathbf{x}_t^*-\mathbf{x}_{t+1} ||^2 \notag \\
    \stackrel{(c)}{\leq} & 2R ||\mathbf{x}_t^2-\mathbf{x}_{t-1}^* || + ||\mathbf{x}_t -\mathbf{x}_{t-1}^* ||^2 - ||\mathbf{x}_t^*-\mathbf{x}_{t+1} ||^2 \label{eq44}
\end{align}
where $(c)$ holds since $||\mathbf{x}_t^* - 2 \mathbf{x}_t + \mathbf{x}_{t-1}^*|| \leq || \mathbf{x}_t^* - \mathbf{x}_t|| + ||\mathbf{x}_t - \mathbf{x}_{t-1}^*|| \leq 2R$. Combining  \eqref{eq44} with \eqref{intermediate_step1} can arrives at
\begin{align}
    &\frac{\Delta(\bm{\lambda}_{t+1})}{\mu} + f_t(\mathbf{x}_t) \nonumber \\ 
    \leq & f_t(\mathbf{x}_t^*) + |\bar{\lambda}| {V}(g) +\mu M^2 + \frac{\alpha G^2}{2} \nonumber \\ 
    &+ \frac{1}{2\alpha} (2R ||\mathbf{x}_t^2-\mathbf{x}_{t-1}^* || + ||\mathbf{x}_t -\mathbf{x}_{t-1}^* ||^2 - ||\mathbf{x}_t^*-\mathbf{x}_{t+1} ||^2)
\end{align}
Summing up over $t=1,2,\cdots,T$%\shen{I think the inequality below is missing a + sign?}
\begin{align}
    & \sum_{t=1}^T \frac{\Delta(\bm{\lambda}_{t+1})}{\mu} + \sum_{t=1}^T f_t(\mathbf{x}_t) \notag \\
    \leq & \sum_{t=1}^T f_t(\mathbf{x}_t^*) + \frac{1}{2\alpha} \sum_{t=1}^T (||\mathbf{x}_t -\mathbf{x}_{t-1}^* ||^2 - ||\mathbf{x}_t^*-\mathbf{x}_{t+1} ||^2) \nonumber \\ 
    &+ \frac{R}{\alpha} V(\{ \mathbf{x}_t^* \}_{t=1}^T) + \sum_{t=1}^T |\bar{\lambda}| V({g}_t(\mathbf{x}_t)) +\mu M^2 T + \frac{\alpha G^2 T}{2} \notag \\
    = & \sum_{t=1}^T f_t(\mathbf{x}_t^*) + \frac{1}{2\alpha}(||\mathbf{x}_1-\mathbf{x}_0^* ||^2 - ||\mathbf{x}_T^*-\mathbf{x}_{T+1} ||) \nonumber \\ 
    &+ \frac{R}{\alpha} V(\{ \mathbf{x}_t^* \}_{t=1}^T)  + |\bar{\lambda} | \sum_{t=1}^T V({g}_t(\mathbf{x}_t)) +\mu M^2 T + \frac{\alpha G^2 T}{2} \notag \\
    \leq & \sum_{t=1}^T f_t(\mathbf{x}_t^*) + \frac{1}{2\alpha}(||\mathbf{x}_1-\mathbf{x}_0^* ||^2) + \frac{R}{\alpha} V(\{ \mathbf{x}_t^* \}_{t=1}^T) \nonumber \\ 
    &+ |\bar{\lambda} | V(\{g_t\}_{t=1}^T) +\mu M^2 T + \frac{\alpha G^2 T}{2} \label{intermediate_step2}
\end{align}
Recall the definition of $V(\{ \mathbf{x}_t^*\}_{t=1}^T)$ and  $V(\{\mathbf{\bar{g}}_t\}_{t=1}^T)$as defined in \eqref{var_g} and \eqref{var_x},
% \begin{align*}
%     V(\{ \mathbf{x}_t^*\}_{t=1}^T) &\coloneqq \sum_{t=1}^T V(\mathbf{x}_t^*) = \sum_{t=1}^T ||\mathbf{x}_t^* - \mathbf{x}_{t-1}^*||\\
%      V(\{g_t\}_{t=1}^T) &\coloneqq \sum_{t=1}^T  V({g}_t) = \sum_{t=1}^T \max_{\mathbf{x}} |g_{t+1}(\mathbf{x}) - g_{t}(\mathbf{x}) |
% \end{align*}
and the fact that $\mathbf{\mathcal{R}}_T = \sum_{t=1}^T f_t(\mathbf{x}_t) - \sum_{t=1}^T f_t(\mathbf{x}_t^*)$, by rearranging the terms of \eqref{intermediate_step2}, we  have

\begin{align}
    \mathbf{\mathcal{R}}_T \leq & \frac{R}{\alpha}V(\{ \mathbf{x}_t^*\}_{t=1}^T) + \frac{||\mathbf{x}_1-\mathbf{x}_0^* ||^2}{2\alpha} + |\bar{\lambda} | V(\{g_t\}_{t=1}^T) \notag \nonumber \\ 
    & + \mu M^2 T + \frac{\alpha G^2 T}{2} - \sum_{t=1}^T \frac{\Delta(\bm{\lambda}_{t+1})}{\mu} \nonumber\\
    \leq & \frac{R}{\alpha}V(\{ \mathbf{x}_t^*\}_{t=1}^T) + \frac{||\mathbf{x}_1-\mathbf{x}_0^* ||^2}{2\alpha} + |\bar{\lambda} | V(\{g_t\}_{t=1}^T) \nonumber \\ 
    &+ \mu M^2 T + \frac{\alpha G^2 T}{2} -\frac{||\bm{\lambda}_{T+2} ||^2}{2\mu} + \frac{||\bm{\lambda}_2||^2}{2\mu} \nonumber\\
    \stackrel{(a)}{\leq} & \frac{R}{\alpha}V(\{ \mathbf{x}_t^*\}_{t=1}^T) +\frac{R^2}{2\alpha} + |\bar{\lambda} | V(\{\mathbf{\bar{g}}_t\}_{t=1}^T) \nonumber \\ 
    &+ \mu M^2 T + \frac{\alpha G^2 T}{2} + \frac{\mu M^2}{2}
\end{align}
where (a) holds since $||\mathbf{x}_1-\mathbf{x}_0^* || \leq R$, $|\bm{\lambda}_{T+2} |^2 \geq 0$, $\bm{\lambda}_{1}=\mathbf{0}$ and $||\bm{\lambda}_2||^2 = |{\lambda}_{1,2}|^2 + |{\lambda}_{2,2}|^2 = (\max(0, \mu g_1(\mathbf{x}_1)))^2 + (\max(0, -\mu g_1(\mathbf{x}_1)))^2 \leq \mu^2(g_1(\mathbf{x}_1))^2  \leq \mu^2 M^2$.

\subsection{Upper bounds} \label{bound_proof}
%If the variation constants are known, the primal and dual stepsizes can be chosen as 
%\begin{align*}
%    \alpha = \mu = \sqrt{\frac{ \max \big\{ V(\{ \mathbf{x}_t^*\}_{t=1}^T),V(\{\mathbf{\bar{g}}_t\}_{t=1}^T) \big\} }{T}}
%\end{align*}
%then the regret is bounded by
%\begin{align*}
%    \mathbf{\mathcal{R}}_T & = \mathcal{O} \big(  \max \big\{ \sqrt{V(\{ \mathbf{x}_t^*\}_{t=1}^T) T}, \sqrt{V(\{\mathbf{\bar{g}}_t\}_{t=1}^T) T} \big\} \big)
%\end{align*}
%and the dynamic fairness is bounded as
%\begin{align*}
%    |\mathbf{\mathcal{F}}_T| & = \mathcal{O} \big(  \max \big\{ \frac{T}{V(\{ \mathbf{x}_t^*\}_{t=1}^T)}, \frac{T}{V(\{\mathbf{\bar{g}}_t\}_{t=1}^T)} \big\} \big).
% \end{align*}
% \blue{Replace $||\Bar{\bm{\lambda}}|| \coloneqq \mu M + \frac{2GR+\frac{R^2}{2\alpha} + \mu M^2}{\epsilon-\bar{V}(g)} = \mathcal{O}(\sqrt{\frac{T}{V^{+}}})$  } % in \eqref{theorem2}
Let $V^{+} =  \max \big\{ V(\{ \mathbf{x}_t^*\}_{t=1}^T),V(\{\mathbf{\bar{g}}_t\}_{t=1}^T) \big\}$. With the knowledge of variations, the primal and dual stepsizes can be chosen as $\alpha = \mu = \sqrt{\frac{ V^{+} }{T}}$. Then 
\begin{align*}
    |\Bar{\lambda}| &\coloneqq 2 \mu M + \frac{2GR+\frac{R^2}{2\alpha} + \mu M^2}{\epsilon-\bar{V}(g)} = \mathcal{O}(\sqrt{\frac{T}{V^{+}}})\\
    |\mathbf{\mathcal{F}}_T| &\leq \frac{|\Bar{\lambda}|}{\mu} = \mathcal{O}(\frac{T}{V^{+}}) \\
    &= \mathcal{O} \left(  \max \left\{ \frac{T}{V(\{ \mathbf{x}_t^*\}_{t=1}^T)}, \frac{T}{V(\{\mathbf{\bar{g}}_t\}_{t=1}^T)} \right\} \right)
\end{align*}

Replace the $\bar{\lambda}$ in \eqref{theorem2}
\begin{align*}
    %||\Bar{\bm{\lambda}}|| \coloneqq \sqrt{\frac{ V^{+} }{T}} M + \frac{2GR+\frac{R^2}{2} \sqrt{\frac{T}{V^{+}}} + \sqrt{\frac{ V^{+} }{T}} M^2}{\epsilon-\bar{V}(g)} = \mathcal{O}(\sqrt{\frac{T}{V^{+}}}) \\
    \mathbf{\mathcal{R}}_T \leq& \frac{R}{\alpha}V(\{ \mathbf{x}_t^*\}_{t=1}^T) +\frac{R^2}{2\alpha} + |\bar{\lambda} | V(\{\mathbf{\bar{g}}_t\}_{t=1}^T) \nonumber \\ 
    &+ \mu M^2 T + \frac{\alpha G^2 T}{2} + \frac{\mu M^2}{2} \\
    \leq& \frac{R}{\alpha}V(\{ \mathbf{x}_t^*\}_{t=1}^T) +\frac{R^2}{2\alpha} + \mu M^2 T + \frac{\alpha G^2 T}{2} + \frac{\mu M^2}{2}\nonumber \\ 
    &+ (2 \mu M + \frac{2GR+\frac{R^2}{2\alpha} + \mu M^2}{\epsilon-\bar{V}(g)}) V(\{\mathbf{\bar{g}}_t\}_{t=1}^T)  \\
    \leq& R \sqrt{ V^{+} T} +\frac{R^2}{2} \sqrt{\frac{ T }{V^{+}}} + \mathcal{O}(\sqrt{\frac{T}{V^{+}}}) V^{+} \nonumber \\ 
    & + \sqrt{ V^{+} T} M^2  + \sqrt{ V^{+} T} \frac{ G^2}{2} + \sqrt{\frac{ V^{+} }{T}} \frac{ M^2}{2} \\
    \coloneqq& \mathcal{O}(\sqrt{ V^{+} T}) \nonumber \\ 
    =&  \mathcal{O} \left(  \max \left\{ \sqrt{V(\{ \mathbf{x}_t^*\}_{t=1}^T) T}, \sqrt{V(\{\mathbf{\bar{g}}_t\}_{t=1}^T) T} \right\} \right)
\end{align*}

\section{Upperbound of $\mathbf{\mathcal{R}}^{\text{off}}_T$} \label{offline_upperbound}
\begin{align}
    \mathbf{\mathcal{R}}^{\text{off}}_T =& \sum_{t=1}^T f_t(\mathbf{x}_t) - \sum_{t=1}^T f_t(\mathbf{x}_t^{\text{off}}) \notag \\
    =& \sum_{t=1}^T f_t(\mathbf{x}_t)  - \sum_{t=1}^T f_t(\mathbf{x}_t^*) + \sum_{t=1}^T f_t(\mathbf{x}_t^*) - \sum_{t=1}^T f_t(\mathbf{x}_t^{\text{off}}) \notag\\
    =& \mathbf{\mathcal{R}}_T + \sum_{t=1}^T f_t(\mathbf{x}_t^*) - \sum_{t=1}^T f_t(\mathbf{x}_t^{\text{off}}) \label{R_t+T}
\end{align}
The term $ \sum_{t=1}^T f_t(\mathbf{x}_t^*) - \sum_{t=1}^T f_t(\mathbf{x}_t^{\text{off}}) $ represents the difference between the performance of instantaneous minimizers and the offline optimal solutions.

Consider the dual function of the instantaneous primal problem (\ref{one-frame}), which can be expressed by minimizing online Lagrangian in (\ref{online_lagrangian}) at time $t$
\begin{align} \label{dual_func}
    \mathcal{J}_t(\bm{\lambda}) \coloneqq& \min_{\mathbf{x}_t \in \mathcal{X}_t}\mathcal{L}_t(\mathbf{x}_t,\bm{\lambda}) \notag \\
    =&\min_{\mathbf{x}_t \in \mathcal{X}_t} f_{t}(\mathbf{x}_t) + \bm{\lambda}_{t}^\top \mathbf{\bar{g}}_{t-1} \notag \\
    =& \min_{\mathbf{x}_t \in \mathcal{X}_t}  f_{t}(\mathbf{x}_t) + \lambda_{1} g_t(\mathbf{x}_t) + \lambda_{2} (-g_t(\mathbf{x}_t)).
\end{align}
And the dual function of (\ref{objective}) over the entire horizon is 
\begin{align}
    \mathcal{J}(\bm{\lambda}) \coloneqq& \min_{\mathbf{x}_t \in \mathcal{X}_t, \forall t} \sum_{t=1}^T \mathcal{L}_t(\mathbf{x}_t,\bm{\lambda}) \notag \\
    = & \sum_{t=1}^T \min_{\mathbf{x}_t \in \mathcal{X}_t} \mathcal{L}_t(\mathbf{x}_t,\bm{\lambda}) = \sum_{t=1}^T \mathcal{J}_t(\bm{\lambda}) .
\end{align}
Since the problems (\ref{one-frame}) and (\ref{objective}) are both convex, assumption 4 implies that strong duality holds. Accordingly, $ \sum_{t=1}^T f_t(\mathbf{x}_t^*) - \sum_{t=1}^T f_t(\mathbf{x}_t^{\text{off}}) $ in \eqref{R_t+T} can be written as
\begin{align}
    \sum_{t=1}^T f_t(\mathbf{x}_t^*) - \sum_{t=1}^T f_t(\mathbf{x}_t^{\text{off}}) = \sum_{t=1}^T \max_{\bm{\lambda}_t} \mathcal{J}_t(\bm{\lambda}_t)  - \max_{\bm{\lambda}} \sum_{t=1}^T \mathcal{J}_t(\bm{\lambda}) \label{step_F1}
\end{align}
% which is the difference between the dual objective of the static best solution and the instantaneous best solution.
Then we establish  $ \sum_{t=1}^T f_t(\mathbf{x}_t^*) - \sum_{t=1}^T f_t(\mathbf{x}_t^{\text{off}}) $ can be bounded by the variation of the dual function.

Define the variation of the dual function (\ref{dual_func}) from time $t$ to $t+1$ as
\begin{align}
    V(\mathcal{J}_t) \coloneqq \max_{\bm{\lambda}} |\mathcal{J}_{t+1}(\bm{\lambda}) - \mathcal{J}_t(\bm{\lambda}) | \label{var_dual}
\end{align}
and the total variation over the time horizon $T$ as 
\begin{align}
   V(\{ \mathcal{J}_t \}_{t=1}^T) \coloneqq \sum_t^{T} V(\mathcal{J}_t) \label{total_var_dual}
\end{align}
Let $\tilde{t}$ denote any time slot in $\{1,\cdots,T\}$, we have
\begin{align}
   &\sum_{t=1}^T \max_{\bm{\lambda}_t} \mathcal{J}_t(\bm{\lambda}_t)  - \max_{\bm{\lambda}} \sum_{t=1}^T \mathcal{J}_t(\bm{\lambda}) \nonumber \\ 
   \leq& \sum_t^{T} (\mathcal{J}_t(\bm{\lambda}_t^*) - \mathcal{J}_t(\bm{\lambda}_{\tilde{t}}^*)) \nonumber \\ 
   \leq& T \max_{t} \{ \mathcal{J}_t(\bm{\lambda}_t^*) - \mathcal{J}_t(\bm{\lambda}_{\tilde{t}}^*) \}. \label{step_F2}
\end{align}
where $\bm{\lambda}_t^* \in \arg\max_{\bm{\lambda}} \mathcal{J}_t(\bm{\lambda})$ which is the instantaneous best solution for \eqref{dual_func}. 

Then we will show that 
\begin{align*}
    \max_{t} \{ \mathcal{J}_t(\bm{\lambda}_t^*) - \mathcal{J}_t(\bm{\lambda}_{\tilde{t}}^*) \} \leq 2 V(\{ \mathcal{J}_t \}_{t=1}^T).
\end{align*}
Assume there exists a slot $t_0$ that $\mathcal{J}_{t_0}(\bm{\lambda}_{t_0}^*) - \mathcal{J}_{t_0}(\bm{\lambda}_{\tilde{t}}^*) > 2 V(\{ \mathcal{J}_t \}_{t=1}^T)$. Then it implies that
\begin{align*}
    \mathcal{J}_{\tilde{t}}(\bm{\lambda}_{\tilde{t}}^*) \stackrel{(a)}{\leq}& \mathcal{J}_{t_0}(\bm{\lambda}_{\tilde{t}}^*) + V(\{ \mathcal{J}_t \}_{t=1}^T) \nonumber \\ 
    <& \mathcal{J}_{t_0}(\bm{\lambda}_{t_0}^*) -  V(\{ \mathcal{J}_t \}_{t=1}^T) \nonumber \\ 
    \stackrel{(b)}{\leq}&  \mathcal{J}_{\tilde{t}}(\bm{\lambda}_{t_0}^*), ~~ \forall \tilde{t}\in T.
\end{align*}
where $(a)$ and $(b)$ come from $\max_{t_1,t_2} |\mathcal{J}_{t_1}(\bm{\lambda}) - \mathcal{J}_{t_2}(\bm{\lambda})| \leq V(\{ \mathcal{J}_t \}_{t=1}^T)$ since $V(\{ \mathcal{J}_t \}_{t=1}^T)$ is the accumulated variation over $T$. 
$\mathcal{J}_{\tilde{t}}(\bm{\lambda}_{\tilde{t}}^*) \leq \mathcal{J}_{\tilde{t}}(\bm{\lambda}_{t_0}^*)$ contradict that that $\bm{\lambda}_{\tilde{t}}^*$ is the maximizer of $\mathcal{J}_{\tilde{t}}(\bm{\lambda})$. Therefore we have $\max_{t} \{ \mathcal{J}_t(\bm{\lambda}_t^*) - \mathcal{J}_t(\bm{\lambda}_{\tilde{t}}^*) \} \leq 2 V(\{ \mathcal{J}_t \}_{t=1}^T)$. Substituting to  leads to \eqref{step_F1} and \eqref{step_F2}
\begin{align*}
    \sum_{t=1}^T f_t(\mathbf{x}_t^*) - \sum_{t=1}^T f_t(\mathbf{x}_t^{\text{off}}) \leq 2 V(\{ \mathcal{J}_t \}_{t=1}^T).
\end{align*}
Replacing the term $\sum_{t=1}^T f_t(\mathbf{x}_t^*) - \sum_{t=1}^T f_t(\mathbf{x}_t^{\text{off}})$ in \eqref{R_t+T} leads to
\begin{align*}
    \mathbf{\mathcal{R}}^{\text{off}}_T = \mathbf{\mathcal{R}}_T + \sum_{t=1}^T f_t(\mathbf{x}_t^*) - \sum_{t=1}^T f_t(\mathbf{x}_t^{\text{off}}) \leq \mathbf{\mathcal{R}}_T +  2 V(\{ \mathcal{J}_t \}_{t=1}^T).
\end{align*}

\section{Flow constraint and Transmission utilization fee} \label{flow-constraint}
The appropriate P2P market design should make it possible to have different prices for different transaction which implies that all market clients may account for preferences, different valuations of electricity and differentiated network charges through the market. Hence, the impact of each transaction on line flow constraints and cost are considered.
Line flow constraints are added as a constraint of the objective function to model physical network in the energy trading.
\begin{align}
    -\rho_l^{\max} \leq \rho_l \leq \rho_l^{\max}
\end{align}
where $\rho_l^{\max}$ is the maximum capacity of line $l$.\\
An method to specify share of each transaction in the line flow constraint is employing Power transfer distribution factor(PTDF). 
% The concept of using PTDF to evaluate impact of each transaction on the line flow is presented in \cite{8782819}.
{PTDF is defined as ``linear approximation of the first order sensitivity of the active power flow and represents change in active power flow over certain line, caused by change in active power generation in certain node of electric power system” \cite{liu2004role}.}
PTDF is used to calculate flow of lines and indicate lines that used for power transfer in each individual transaction \cite{8782819},\cite{9102274}.
PTDF for line $l$ shown by $\phi^{ij}_l$ and indicates part of the generated energy by seller $i$, which is transferred to buyer $j$ through line $l$. $\phi^{ij}_l$ can be obtained using 
\begin{align}
    \phi^{ij}_l=\phi^{i}_l - \phi^{j}_l
\end{align}
where $\phi^{i}_l , \phi^{j}_l$ are injection shift factors(ISF) in line $l$ for node $i$ and $j$ respectively.
{ISF is an approximation of the sensitivity matrix and quantifies the redistribution of power through each branch following a change in generation or load in a particular node.} 
The ISF matrix is shown as $\Phi \in \mathbb{R}^{N_l \times N}$ where $N_l$ denotes the number of lines.
This matrix can be obtained by having diagonal branch susceptance matrix($B'$), branch to node incidence matrix($A$) and reduced nodal susceptance matrix($C$).
PTDF matrix depends on the network structure and is independent of the power flows through the network .\\
\begin{align}
    & \Phi = B' A C^{-1}\\
    & B' = diag[b_1,\dots, b_{N_l}] \in \mathbb{R}^{N_l \times N_l}\\
    & A= [a_1,a_2,\cdots,a_{N_l}]^\top \in \mathbb{R}^{N_l \times N}\\
    & C = A^T B' A \in \mathbb{R}^{N \times N}
\end{align}
In the matrix $A$, $a_l^\top \stackrel{e.g.}{=} [1,-1,0,\cdots,0]$ is the $l$th row, in which a line exist between the first and second node.
\\
By having PTDF, the power transfer distance between consumer $j$ and producer $i$ can be obtained as
\begin{align}
    \mathbf{D}^{ij} = \sum_{l \in N_l} \phi^{ij}_l, \quad \mathbf{D} \in \mathbb{R}^{N \times N}
\end{align}
With traded energy decision($\mathbf{X}_t$), the power flow in line $l$ can be obtained by
\begin{align}
    \rho_l(\mathbf{X}_t) = \sum_{i\in w_t^p} \sum_{j\in w_t^c} \phi^{ij}_{l}\mathbf{x}_t^{ij} % = \sum_{i\in w_c} \sum_{j\in w_p} \phi_{ij}^{l}P_{ij},
\end{align}
and the total line utilization fee can be calculated as
\begin{align*}
     \sum_{i \in w_t^p} \sum_{j \in N} \gamma \mathbf{D}^{ij} \mathbf{X}_t^{ij} = -\sum_{j \in w_t^c} \sum_{i \in N} \gamma \mathbf{D}^{ji} \mathbf{X}_t^{ji}.
\end{align*}
where $w_t^p$ denotes the clients who are producers in time slot $t$ and $\mathbf{X}_t^{ij} \geq 0, \forall i \in w_t^p,j\in N$, $w_t^c$ denotes the clients who are consumers in time slot $t$ and $\mathbf{X}_t^{ji} \geq 0, \forall j \in w_t^p,i\in N$.

%\newpage
%\ % empty page
%\newpage
\section{Proof of Lemma 1} \label{appendix_lemma1}
\textbf{Lemma 1 }
The difference between the dual variables of two successive iterates $|\lambda_{1,t+1} - \lambda_{2,t+1}|$ can be upper bounded by 
\begin{align}
    |\lambda_{1,t+1} - \lambda_{2,t+1}| \leq   \bar{\lambda} := 2 \mu M + \frac{2GR+\frac{R^2}{2\alpha} + \mu M^2}{\epsilon-\bar{V}(g)} %\label{lambda_upperbound}%\\ | \lambda_{1,t+1}|
    %| \lambda_{2,t+1}| \leq  \bar{\lambda}_2 \leq \bar{\lambda} = \mu M + \frac{2GR+\frac{R^2}{2\alpha} + \mu M^2}{\epsilon-\bar{V}(g)} \label{lambda2_upperbound}
\end{align}
% where $\epsilon > \bar{V}(g)$, $\bar{V}(g) = \max_{t\in T} \max_{\mathbf{x} \in \mathcal{X}} || g_{t+1}(\mathbf{x}) - g_t(\mathbf{x}) ||$ due to assumption 5.

\textbf{Proof:}\\ % The upper bound $\bar{\lambda}$ (\ref{bar_lambda}) can be proven by contradiction.\\

Based on the updating rule of \eqref{1_update} and \eqref{2_update}, we have
\begin{align*}
    |\lambda_{1,t+1}|^2 =& | \max( 0,\lambda_{1,t} + \mu g_t(\mathbf{x}_t)) |^2 \nonumber \\
    \leq& | \lambda_{1,t} + \mu g_t(\mathbf{x}_t) |^2 \nonumber \\
    =& |\lambda_{1,t} |^2 + 2 \mu \lambda_{1,t} g_t(\mathbf{x}_t) + \mu^2 |g_t(\mathbf{x}_t)|^2\\
    |\lambda_{2,t+1}|^2 =& | \max( 0,\lambda_{2,t} - \mu g_t(\mathbf{x}_t)) |^2  \nonumber \\
    \leq& | \lambda_{2,t} + \mu g_t(\mathbf{x}_t) |^2 \nonumber \\
    =& |\lambda_{2,t} |^2 - 2 \mu \lambda_{2,t} g_t(\mathbf{x}_t) + \mu^2 |g_t(\mathbf{x}_t)|^2
\end{align*}
Reorganizing the terms leads to
\begin{subequations}
\begin{align}
    \Delta(\lambda_{1,t}) &\coloneqq \frac{|\lambda_{1,t+1}|^2 - |\lambda_{1,t}|^2}{2} \leq \mu\lambda_{1,t} g_t(\mathbf{x}_t) + \frac{\mu^2}{2} |g_t(\mathbf{x}_t)|^2 \\
     \Delta(\lambda_{2,t}) &\coloneqq \frac{|\lambda_{2,t+1}|^2 - |\lambda_{2,t}|^2}{2} \leq -\mu\lambda_{2,t} g_t(\mathbf{x}_t) + \frac{\mu^2}{2} |g_t(\mathbf{x}_t)|^2
\end{align}
\end{subequations}
Combining the two inequalities results in
\begin{align}
    \Delta(\bm{\lambda}_{t})  &\coloneqq \frac{||\bm{\lambda}_{t+1} ||^2 - ||\bm{\lambda}_{t} ||^2}{2} \leq \mu \bm{\lambda}_t^\top \mathbf{\bar{g}}_t(\mathbf{x}_t) + \mu^2 |g_t(\mathbf{x}_t)|^2 \label{delta_lambda}
\end{align}

where $\bm{\lambda}_{t}=[\lambda_{1,t}, \lambda_{2,t}]^\top$ and $\mathbf{\bar{g}}_{t}(\mathbf{x})=[g_t(\mathbf{x}),-g_t(\mathbf{x})]^\top$.
%\textit{Proof:} $|\lambda_{1,t+1}|^2 = | \max( 0,\lambda_{1,t} + \mu g_t(\mathbf{x}_t)) |^2 \leq | \lambda_{1,t} + \mu g_t(\mathbf{x}_t) |^2 = |\lambda_{1,t} |^2 + 2 \mu \lambda_{1,t} g_t(\mathbf{x}_t) + \mu^2 |g_t(\mathbf{x}_t)|^2$. $ \Delta(\lambda_{1,t}) \leq \mu\lambda_{1,t} g_t(\mathbf{x}_t) + \frac{\mu^2}{2} |g_t(\mathbf{x}_t)|^2$.\\
%Similarly, $ \Delta(\lambda_{2,t}) \leq -\mu\lambda_{2,t} g_t(\mathbf{x}_t) + \frac{\mu^2}{2} |g_t(\mathbf{x}_t)|^2$. $\Delta(\bm{\lambda}_{t}) = \Delta(\lambda_{1,t})+\Delta(\lambda_{2,t}) \leq \mu \bm{\lambda}_t^\top \mathbf{\bar{g}}_t(\mathbf{x}_t) + \mu^2 |g_t(\mathbf{x}_t)|^2$.
%Recall that $\mathbf{x}_{t+1} = \min_{\mathbf{x}} \nabla f_{t}(\mathbf{x}_{t})^\top(\mathbf{x}-\mathbf{x}_{t}) + \bm{\lambda}_{t}^\top \mathbf{\bar{g}}_{t}(\mathbf{x}) + \frac{||\mathbf{x}-\mathbf{x}_{t} ||^2}{2\alpha}$. 
Based on the update in \eqref{primal_update}, i.e., $\mathbf{x}_{t+1} = \min_{\mathbf{x}} \nabla f_{t}(\mathbf{x}_{t})^\top(\mathbf{x}-\mathbf{x}_{t}) + \bm{\lambda}_{t}^\top \mathbf{\bar{g}}_{t}(\mathbf{x}) + \frac{||\mathbf{x}-\mathbf{x}_{t} ||^2}{2\alpha}$, the following inequality hold for $\tilde{\mathbf{x}}$
\begin{align}
  &\nabla f_t(\mathbf{x}_t)^\top (\mathbf{x}_{t+1} - \mathbf{x}_t) + \bm{\lambda}_{t+1}^\top \mathbf{\bar{g}}_t(\mathbf{x}_{t+1}) + \frac{1}{2\alpha} ||\mathbf{x}_{t+1}-\mathbf{x}_t||^2  \notag  \\
  {\leq}& \nabla f_t(\mathbf{x}_t)^\top (\tilde{\mathbf{x}}_{t} - \mathbf{x}_t) + \bm{\lambda}_{t+1}^\top \mathbf{\bar{g}}_t(\tilde{\mathbf{x}}_{t}) + \frac{1}{2\alpha} ||\tilde{\mathbf{x}}_{t}-\mathbf{x}_t||^2 \notag 
%&\nabla f_t(\mathbf{x}_t)^\top (\mathbf{x}_{t+1} - \mathbf{x}_t) + \bm{\lambda}_{t+1}^\top \mathbf{\bar{g}}_t(\mathbf{x}_{t+1}) + \frac{1}{2\alpha} ||\mathbf{x}_{t+1}-\mathbf{x}_t||^2     \leq \nabla f_t(\mathbf{x}_t)^\top (\tilde{\tilde{\mathbf{x}}}_{t} - \mathbf{x}_t) + \bm{\lambda}_{t+1}^\top \mathbf{\bar{g}}_t(\tilde{\tilde{\mathbf{x}}}_t) + \frac{1}{2\alpha} ||\tilde{\tilde{\mathbf{x}}}_{t}-\mathbf{x}_t||^2 \notag   
\end{align}

In the case when $\lambda_{1,t+1} \geq \lambda_{2,t+1}$, the  right-hand side  can be equivalently written and bounded as follows

\begin{align}
    & \nabla f_t(\mathbf{x}_t)^\top (\tilde{\mathbf{x}}_{t} - \mathbf{x}_t) + \bm{\lambda}_{t+1}^\top \mathbf{\bar{g}}_t(\tilde{\mathbf{x}}_{t}) + \frac{1}{2\alpha} ||\tilde{\mathbf{x}}_{t}-\mathbf{x}_t||^2 \notag 
 %   \\
  %  & = \nabla f_t(\mathbf{x}_t)^\top (\tilde{\mathbf{x}}_{t} - \mathbf{x}_t) + ( \lambda_{1,t+1} g_t(\tilde{\mathbf{x}}_{t}) + \lambda_{2,t+1}(-g_t(\tilde{\mathbf{x}}_{t}))) + \frac{1}{2\alpha} ||\tilde{\mathbf{x}}_{t}-\mathbf{x}_t||^2 \notag 
    \\
    =& \nabla f_t(\mathbf{x}_t)^\top (\tilde{\mathbf{x}}_{t} - \mathbf{x}_t) + ( \lambda_{1,t+1} - \lambda_{2,t+1}) g_t(\tilde{\mathbf{x}}_{t}) \nonumber \\
    &+ \frac{1}{2\alpha} ||\tilde{\mathbf{x}}_{t}-\mathbf{x}_t||^2 \notag 
    \\
    %& \stackrel{a}{\leq} \nabla f_t(\mathbf{x}_t)^\top (\tilde{\mathbf{x}}_{t} - \mathbf{x}_t) +   \lambda_{1,t+1} g_t(\tilde{\mathbf{x}}_{t})  + \frac{1}{2\alpha} ||\tilde{\mathbf{x}}_{t}-\mathbf{x}_t||^2 \notag 
    %\\
     \stackrel{(a)}{\leq}& \nabla f_t(\mathbf{x}_t)^\top (\tilde{\mathbf{x}}_{t} - \mathbf{x}_t) - \epsilon   (\lambda_{1,t+1}-\lambda_{2,t+1})  + \frac{1}{2\alpha} ||\tilde{\mathbf{x}}_{t}-\mathbf{x}_t||^2 \notag
    \\
    % & \stackrel{b}{=} \nabla f_t(\mathbf{x}_t)^\top (\tilde{\mathbf{x}}_{t} - \mathbf{x}_t) -\epsilon (||\lambda_{1,t+1}|| + ||\lambda_{2,t+1}||) + \frac{1}{2\alpha} ||\tilde{\mathbf{x}}_{t}-\mathbf{x}_t||^2 \notag \\
    =& \nabla f_t(\mathbf{x}_t)^\top (\tilde{\mathbf{x}}_{t} - \mathbf{x}_t) - \epsilon |\lambda_{1,t+1} - \lambda_{2,t+1}| + \frac{1}{2\alpha} ||\tilde{\mathbf{x}}_{t}-\mathbf{x}_t||^2 \label{lambda1>2}
\end{align}
where  (a) is true do to assumption 4.
% where (a) follows by $\lambda_{1,t+1} \geq \lambda_{2,t+1}, \lambda_{1,t+1}-\lambda_{2,t+1} \geq 0, |\lambda_{1,t+1}-\lambda_{2,t+1}| =\lambda_{1,t+1}-\lambda_{2,t+1} $, .
Rearranging terms in \eqref{lambda1>2},
\begin{align} \label{lambda_g_1>2}
    &\bm{\lambda}_{t+1}^\top \mathbf{\bar{g}}_t(\mathbf{x}_{t+1})\nonumber \\ 
    \leq & \nabla f_t(\mathbf{x}_t)^\top (\tilde{\mathbf{x}}_{t} - \mathbf{x}_t) - \nabla f_t(\mathbf{x}_t)^\top (\mathbf{x}_{t+1} - \mathbf{x}_t)   \nonumber \\ 
    & - \epsilon |\lambda_{1,t+1} - \lambda_{2,t+1}| + \frac{1}{2\alpha} ||\tilde{\mathbf{x}}_{t}-\mathbf{x}_t||^2 - \frac{1}{2\alpha} ||\mathbf{x}_{t+1}-\mathbf{x}_t||^2 \notag \\
    \stackrel{(a)}{\leq} &  ||\nabla f_t(\mathbf{x}_t)|| ||\tilde{\mathbf{x}}_{t} - \mathbf{x}_t|| - ||\nabla f_t(\mathbf{x}_t)|| ||\mathbf{x}_{t+1} - \mathbf{x}_t|| \nonumber \\ 
    & - \epsilon |\lambda_{1,t+1} - \lambda_{2,t+1}| + \frac{1}{2\alpha} ||\tilde{\mathbf{x}}_{t}-\mathbf{x}_t||^2 - \frac{1}{2\alpha} ||\mathbf{x}_{t+1}-\mathbf{x}_t||^2 \notag\\
    %\bm{\lambda}_{t+1}^\top \mathbf{\bar{g}}_t(\mathbf{x}_{t+1}) 
    \stackrel{(b)}{\leq} & 2GR - \epsilon |\lambda_{1,t+1} - \lambda_{2,t+1}| + \frac{R^2}{2\alpha} %||\bm{\lambda}_{t+1}||
\end{align}
where $(a)$ comes from Cauchy–Schwarz inequality and $(b)$ results from assumptions 2 and 3. 

Similarly, in the case when $\lambda_{1,t+1} \leq \lambda_{2,t+1}$, the following holds
\begin{align}
    &\nabla f_t(\mathbf{x}_t)^\top (\mathbf{x}_{t+1} - \mathbf{x}_t) + \bm{\lambda}_{t+1}^\top \mathbf{\bar{g}}_t(\mathbf{x}_{t+1}) + \frac{1}{2\alpha} ||\mathbf{x}_{t+1}-\mathbf{x}_t||^2  \notag \\
    \leq & \nabla f_t(\mathbf{x}_t)^\top (\tilde{\tilde{\mathbf{x}}}_{t} - \mathbf{x}_t) + \bm{\lambda}_{t+1}^\top \mathbf{\bar{g}}_t(\tilde{\tilde{\mathbf{x}}}_t) + \frac{1}{2\alpha} ||\tilde{\tilde{\mathbf{x}}}_{t}-\mathbf{x}_t||^2 \notag 
   % \\
   % & = \nabla f_t(\mathbf{x}_t)^\top (\tilde{\tilde{\mathbf{x}}}_t - \mathbf{x}_t) + ( \lambda_{1,t+1} g_t(\tilde{\mathbf{x}}_{t}) + \lambda_{2,t+1}(-g_t(\tilde{\tilde{\mathbf{x}}}_t))) + \frac{1}{2\alpha} ||\tilde{\tilde{\mathbf{x}}}_t-\mathbf{x}_t||^2 \notag 
    \\
    =& \nabla f_t(\mathbf{x}_t)^\top (\tilde{\tilde{\mathbf{x}}}_{t} - \mathbf{x}_t) + (  \lambda_{2,t+1} -\lambda_{1,t+1}) (-g_t(\tilde{\tilde{\mathbf{x}}}_{t})) \nonumber \\ 
    & + \frac{1}{2\alpha} ||\tilde{\tilde{\mathbf{x}}}_{t}-\mathbf{x}_t||^2 \notag 
    \\
    \leq& \nabla f_t(\mathbf{x}_t)^\top (\tilde{\tilde{\mathbf{x}}}_{t} - \mathbf{x}_t) - \epsilon   (\lambda_{2,t+1}-\lambda_{1,t+1})  + \frac{1}{2\alpha} ||\tilde{\tilde{\mathbf{x}}}_{t}-\mathbf{x}_t||^2 \notag
    \\
    \leq& \nabla f_t(\mathbf{x}_t)^\top (\tilde{\tilde{\mathbf{x}}}_{t} - \mathbf{x}_t) - \epsilon |\lambda_{1,t+1} - \lambda_{2,t+1}| + \frac{1}{2\alpha} ||\tilde{\tilde{\mathbf{x}}}_{t}-\mathbf{x}_t||^2 \label{lambda2>1}
\end{align}
Rearranging terms in \eqref{lambda2>1}
\begin{align} \label{lambda_g_2>1}
    &\bm{\lambda}_{t+1}^\top \mathbf{\bar{g}}_t(\mathbf{x}_{t+1}) \notag\\
    \leq & \nabla f_t(\mathbf{x}_t)^\top (\tilde{\tilde{\mathbf{x}}}_{t} - \mathbf{x}_t) - \nabla f_t(\mathbf{x}_t)^\top (\mathbf{x}_{t+1} - \mathbf{x}_t)  \nonumber \\ 
    &- \epsilon |\lambda_{1,t+1} - \lambda_{2,t+1}| + \frac{1}{2\alpha} ||\tilde{\tilde{\mathbf{x}}}_{t}-\mathbf{x}_t||^2 - \frac{1}{2\alpha} ||\mathbf{x}_{t+1}-\mathbf{x}_t||^2 \notag \\
    \leq &  ||\nabla f_t(\mathbf{x}_t)|| ||\tilde{\tilde{\mathbf{x}}}_{t} - \mathbf{x}_t|| - ||\nabla f_t(\mathbf{x}_t)|| ||\mathbf{x}_{t+1} - \mathbf{x}_t|| \nonumber \\ 
    & - \epsilon |\lambda_{1,t+1} - \lambda_{2,t+1}| + \frac{1}{2\alpha} ||\tilde{\tilde{\mathbf{x}}}_{t}-\mathbf{x}_t||^2 - \frac{1}{2\alpha} ||\mathbf{x}_{t+1}-\mathbf{x}_t||^2 \notag\\
    %\bm{\lambda}_{t+1}^\top \mathbf{\bar{g}}_t(\mathbf{x}_{t+1}) \shen{Idon't understand why you need this part on the left}
    \leq & 2GR - \epsilon |\lambda_{1,t+1} - \lambda_{2,t+1}| + \frac{R^2}{2\alpha} %||\bm{\lambda}_{t+1}||
\end{align}
Combining \eqref{lambda_g_1>2} and \eqref{lambda_g_2>1} leads to
\begin{align} \label{lambda_g_all}
\bm{\lambda}_{t+1}^\top \mathbf{\bar{g}}_t(\mathbf{x}_{t+1})\leq 2GR - \epsilon |\lambda_{1,t+1} - \lambda_{2,t+1}| + \frac{R^2}{2\alpha}.
\end{align}
%\shen{where does the next inequality come from? Never start a equation without any explanation}
Hence, \eqref{delta_lambda} can be re-written as follows by adding and subtracting $\mu\bm{\lambda}_{t+1}^\top \mathbf{\bar{g}}_t(\mathbf{x}_{t+1})$

%\textbf{otherwise($g_{t}(\mathbf{x}_t) < 0$), $x_{t+1} = \min_{x} \nabla f_{t}(\mathbf{x}_{t})^\top(\mathbf{x}-x_{t}) - \lambda_{2,t}^\top g_{t}(\mathbf{x}) + \frac{||x-x_{t} ||^2}{2\alpha}$ :}
%\begin{align}
%    &\nabla f_t(\mathbf{x}_t)^\top (\mathbf{x}^*_{t+1} - \mathbf{x}_t) + \lambda_{1,t+1}^\top g_t(\mathbf{x}^*_{t+1}) + \frac{1}{2\alpha} ||x^*_{t+1}-\mathbf{x}_t||^2 \notag\\
%    \leq & \nabla f_t(\mathbf{x}_t)^\top (\mathbf{x}_{t+1} - \mathbf{x}_t) + \lambda_{1,t+1}^\top g_t(\mathbf{x}_{t+1}) + \frac{1}{2\alpha} ||x_{t+1}-\mathbf{x}_t||^2
%\end{align}

%\begin{align}
%    \lambda_{1,t+1}^\top (g_t(\mathbf{x}^*_{t+1}) - g_t(\mathbf{x}_{t+1})) \leq &  \nabla f_t(\mathbf{x}_t)^\top (\mathbf{x}_{t+1} - \mathbf{x}_t) - \nabla f_t(\mathbf{x}_t)^\top (\mathbf{x}^*_{t+1} - \mathbf{x}_t) \notag \\
%    &+ \frac{1}{2\alpha} ||x_{t+1}-\mathbf{x}_t||^2 - \frac{1}{2\alpha} ||x^*_{t+1}-\mathbf{x}_t||^2\\
%    & \leq 2GR + \frac{R^2}{2\alpha}
%\end{align}
%\noindent
%\blue{With new assumption 4:}
%\begin{align}
%    &\lambda_{1,t+1}^\top( g_t(\mathbf{x}_{t+1})-g_t(\mathbf{x}^*_{t+1})) \leq || \lambda_{1,t+1} ||S\\
%    &\lambda_{1,t+1}^\top g_t(\mathbf{x}_{t+1}) \leq2GR - \epsilon ||\lambda_{1,t+1}|| + \frac{R^2}{2\alpha}+|| \lambda_{1,t+1}||S
%\end{align}

\begin{align} \label{delta_lambda1}
    \Delta(\bm{\lambda}_{t+1}) \leq & \mu\bm{\lambda}_{t+1}^\top \mathbf{\bar{g}}_{t+1}(\mathbf{x}_{t+1}) -\mu  \bm{\lambda}_{t+1}^\top \mathbf{\bar{g}}_t(\mathbf{x}_{t+1}) \nonumber \\ 
    & + \mu\bm{\lambda}_{t+1}^\top \mathbf{\bar{g}}_t(\mathbf{x}_{t+1}) + \mu^2 |g_{t+1}(\mathbf{x}_{t+1})|^2 \notag\\
    \stackrel{(a)}{\leq} & \mu\bm{\lambda}_{t+1}^\top (\mathbf{\bar{g}}_{t+1}(\mathbf{x}_{t+1}) - \mathbf{\bar{g}}_t(\mathbf{x}_{t+1})) \nonumber \\ 
    & + \mu(2GR - \epsilon |\lambda_{1,t+1} - \lambda_{2,t+1}| + \frac{R^2}{2\alpha}) + \mu^2 M^2 \notag 
    \\
    = & \mu ({\lambda}_{1,t+1} - {\lambda}_{2,t+1} ) (g_{t+1}(\mathbf{x}_{t+1}) -{g}_t(\mathbf{x}_{t+1}) )  \nonumber \\ 
    & -\epsilon  \mu |\lambda_{1,t+1} - \lambda_{2,t+1}| + 2\mu GR + \frac{\mu R^2}{2\alpha} + \mu^2 M^2 \notag
    \\
    %\Delta(\bm{\lambda}_{t+1})  
    \stackrel{(b)}{\leq}& \mu \bar{V}(g) |\lambda_{1,t+1} - \lambda_{2,t+1}| -\mu\epsilon   |\lambda_{1,t+1} - \lambda_{2,t+1}| \nonumber \\ 
    & + 2\mu GR + \frac{\mu R^2}{2\alpha} + \mu^2 M^2 % \mu \bar{V}(g) -> 2\mu \bar{V}(g) ?
\end{align}
where $(a)$ because of \eqref{lambda_g_all} and $(b)$ holds due to Cauchy-Schwarz inequality and assumption 5.

Note that Lemma 1 holds at $t=1$ since $\bm{\lambda}_1 = \mathbf{0}$.
Assume that (\ref{lambda_upperbound}) holds for all time slots up till $t+1$, and $t+2$ is the first time slot that  (\ref{lambda_upperbound}) does not hold. Then we have
\begin{subequations}
\begin{align}
    |\lambda_{1,t+1} - \lambda_{2,t+1}| &\leq 2\mu M + \frac{2GR+\frac{R^2}{2\alpha} + \mu M^2}{\epsilon-\bar{V}(g)} \\
    |\lambda_{1,t+2} - \lambda_{2,t+2}| &\geq  2\mu M + \frac{2GR+\frac{R^2}{2\alpha} + \mu M^2}{\epsilon-\bar{V}(g)} \label{upperbound_vio}
\end{align}      
\end{subequations}

Hence, it holds that
\begin{align*}
    &|\lambda_{1,t+1} - \lambda_{2,t+1}| \\
    \geq& |\lambda_{1,t+2} - \lambda_{2,t+2}| - |(\lambda_{1,t+1} - \lambda_{2,t+1}) -(\lambda_{1,t+2} - \lambda_{2,t+2})|\\
    \geq& |\lambda_{1,t+2} - \lambda_{2,t+2}| - \nonumber |({\lambda_{1,t+1}}+\mu {g}_{t+1}(\mathbf{x}_{t+1})-{\lambda_{1,t+1}}) \nonumber \\ 
    &- ({\lambda_{2,t+1}}-\mu {g}_{t+1}(\mathbf{x}_{t+1})-{\lambda_{2,t+1}} )|\\
    =& |\lambda_{1,t+2} - \lambda_{2,t+2}| - |2 \mu {g}_{1,t+1}(\mathbf{x}_{t+1})|\\
    \stackrel{(a)}{\geq}& \frac{2GR+\frac{R^2}{2\alpha} + \mu M^2}{\epsilon-\bar{V}(g)}
\end{align*}
where $(a)$ holds due to \eqref{upperbound_vio} and assumption 2.
Substituting it into (\ref{delta_lambda1}) and combining with assumption 5 arrives at $\Delta(\bm{\lambda}_{t+1})=\frac{\lambda_{1,t+2}^2 + \lambda_{2,t+2}^2 - \lambda_{1,t+1}^2 - \lambda_{2,t+1}^2}{2}  \leq 0$, %which contradicts $|\lambda_{1,t+1} - \lambda_{2,t+1}| \leq \bar{\lambda} \leq |\lambda_{1,t+2} - \lambda_{2,t+2}|$.
%Hence, we have $\lambda_{1,t+2}^2 + \lambda_{2,t+2}^2 - \lambda_{1,t+1}^2 - \lambda_{2,t+1}^2 \leq 0$ 
which can be written as
% \begin{align*}
%     %& \lambda_{1,t+2}^2 + \lambda_{2,t+2}^2 - \lambda_{1,t+1}^2 - \lambda_{2,t+1}^2 \leq 0 \\
%     &(\lambda_{1,t+2} + \lambda_{1,t+1})(\lambda_{1,t+2} - \lambda_{1,t+1}) + \lambda_{2,t+2} + \lambda_{2,t+1})(\lambda_{2,t+2} - \lambda_{2,t+1}) \leq 0
% \end{align*}
% After rearranging terms, it follows that
\begin{align}
    &(\lambda_{1,t+2} + \lambda_{1,t+1})(\lambda_{1,t+2} - \lambda_{1,t+1}) \nonumber \\ 
    \leq& (\lambda_{2,t+2} + \lambda_{2,t+1})(\lambda_{2,t+1} - \lambda_{2,t+2} ). \label{delta_lambda<1}
\end{align}
%\shen{revised till here, cannot follow the following any more, needs some words to explain}
%\shen{fix the cross-refrencing}
If $g_{t+1}(\mathbf{x}_{t+1}) >0$, the updating rule in \eqref{1_update} leads to $\lambda_{1,t+2} - \lambda_{1,t+1} = \mu g_{t+1}(\mathbf{x}_{t+1})$ and \eqref{2_update} implies $ 0 \leq  \lambda_{2,t+1} -\lambda_{2,t+2}\leq \mu g_{t+1}(\mathbf{x}_{t+1})$. Substituting them into \eqref{delta_lambda<1} leads to
\begin{align*}
    %&(\lambda_{1,t+2} + \lambda_{1,t+1})(\lambda_{1,t+2} - \lambda_{1,t+1}) \leq (\lambda_{2,t+2} + \lambda_{2,t+1})(\lambda_{2,t+1} - \lambda_{2,t+2} ) 
    %\\
    %&(\lambda_{1,t+2} + \lambda_{1,t+1})(\lambda_{1,t+2} - \lambda_{1,t+1}) = (\lambda_{1,t+2} + \lambda_{1,t+1}) \mu g_{t+1}(\mathbf{x}_{t+1}) \\
    %&(\lambda_{2,t+2} + \lambda_{2,t+1})(\lambda_{2,t+1} - \lambda_{2,t+2} ) \leq (\lambda_{2,t+2} + \lambda_{2,t+1}) \mu g_{t+1}(\mathbf{x}_{t+1})\\
    &(\lambda_{1,t+2} + \lambda_{1,t+1}) \mu g_{t+1}(\mathbf{x}_{t+1}) \nonumber \\ 
    \leq&(\lambda_{2,t+2} + \lambda_{2,t+1}) \mu g_{t+1}(\mathbf{x}_{t+1}) 
\end{align*}
which is equivalent to $\lambda_{1,t+2} + \lambda_{1,t+1} \leq \lambda_{2,t+2} + \lambda_{2,t+1}$. Furthermore, 
since % $\lambda_{1,t+2} - \lambda_{1,t+1} = \mu g_{t+1}(\mathbf{x}_{t+1})$, 
$\lambda_{1,t+2} + \lambda_{1,t+1} = 2 \lambda_{1,t+1} + \mu g_{t+1}(\mathbf{x}_{t+1})$ and % $\lambda_{2,t+1} -\lambda_{2,t+2}  \leq \mu g_{t+1}(\mathbf{x}_{t+1}) $, 
$\lambda_{2,t+1} + \lambda_{2,t+2} \leq  2\lambda_{2,t+2} + \mu g_{t+1}(\mathbf{x}_{t+1})$,
we have
\begin{align*}
    2 \lambda_{1,t+1} + \mu g_{t+1}(\mathbf{x}_{t+1}) \leq 2\lambda_{2,t+2} + \mu g_{t+1}(\mathbf{x}_{t+1})
\end{align*}
which implies $\lambda_{1,t+1} \leq \lambda_{2,t+2}$ and henceforth  $ \lambda_{1,t+2} \leq \lambda_{2,t+1}$. % Then $\lambda_{1,t+2} = \lambda_{1,t+1} + \mu g_{t+1}(\mathbf{x}_{t+1}) \leq \lambda_{2,t+2} +\mu g_{t+1}(\mathbf{x}_{t+1}) \leq \lambda_{2,t+1}$. 
\\
Due to $\lambda_{1,t+1} \leq \lambda_{2,t+2} \leq \lambda_{2,t+1}$, $\lambda_{1,t+1} \leq \lambda_{1,t+2} \leq \lambda_{2,t+1}$ and non-negativity of $\lambda_{1,t+1}, \lambda_{2,t+1} ,\lambda_{1,t+2}, \lambda_{2,t+2}$, we have 
\begin{align*}
    |\lambda_{1,t+1} - \lambda_{2,t+1}| \geq |\lambda_{1,t+2} - \lambda_{2,t+2}| 
\end{align*}
which contradicts the $|\lambda_{1,t+1} - \lambda_{2,t+1}| \leq  |\lambda_{1,t+2} - \lambda_{2,t+2}|$.

Similarly, if $g_{t+1}(\mathbf{x}_{t+1}) <0$, we have 
$\lambda_{2,t+1} \leq \lambda_{1,t+2} \leq \lambda_{1,t+1}$ and $\lambda_{2,t+1} \leq \lambda_{2,t+2} \leq \lambda_{1,t+1}$. 
Due to the non-negativity of $\lambda_{1,t+1}, \lambda_{2,t+1} ,\lambda_{1,t+2}, \lambda_{2,t+2}$, we can have $|\lambda_{1,t+1} - \lambda_{2,t+1}| \geq |\lambda_{1,t+2} - \lambda_{2,t+2}|$ which contradicts the $|\lambda_{1,t+1} - \lambda_{2,t+1}| \leq  |\lambda_{1,t+2} - \lambda_{2,t+2}|$. To sum up, Lemma 1 can be proved by contradiction. 

Proof of \textbf{Lemma 1} is completed.

\section{Proof of Lemma 2} \label{appendix_lemma2}
\textbf{Lemma 2: } Let $\bar{\delta} := \max_{t \in \{ 1,\cdots,T \}} |\lambda_{1,t+1} - \lambda_{2,t+1}|$, the following equality holds
\begin{align}
     \bar{\delta}  = \max(\bar{\lambda}_1, \bar{\lambda}_2) %\label{lemma2}
\end{align}
which implies that $\bar{\lambda}_1\leq \bar{\delta} \leq \bar{\lambda}$, and $ \bar{\lambda}_2 \leq \bar{\delta} \leq \bar{\lambda}$. 
$\bar{\lambda}_1,\bar{\lambda}_2$ denote the max value of $\lambda_{1,t},\lambda_{2,t}$ across the whole time horizon, i.e., $\bar{\lambda}_1 \coloneqq \max_{t \in \{ 1,\cdots,T \}} \lambda_{1,t}  $, $\bar{\lambda}_2 \coloneqq \max_{t \in \{ 1,\cdots,T \}} \lambda_{2,t}  $.
%Since $\lambda_{1,t+1},\lambda_{2,t+1} \geq 0$, \blue{$\bar{|\lambda_{1,t+1} - \lambda_{2,t+1}|} = \max(\bar{\lambda}_1,\bar{\lambda}_2)=\bar{\lambda}$.} % = \mu M + \frac{2GR+\frac{R^2}{2\alpha} + \mu M^2}{\epsilon-\bar{V}(g)}

%\shen{revised till here}

\textbf{Proof of Lemma 2}: by  mathematical induction % \blue{$\lambda_{d,t} \rightarrow \delta_t$}

\textit{Proposition}: for time slot $t$, $\bar{\delta}_t = \max_{t} |\lambda_{1,t+1} - \lambda_{2,t+1}| = \max(\bar{\lambda}_{1,t}, \bar{\lambda}_{2,t})$. $\bar{\lambda}_{1,t}$ and $\bar{\lambda}_{2,t}$ denote the maximum value of $\lambda_{1,t+1}$ and $\lambda_{2,t+1}$ at time $t$ and before time $t$.

\textit{Base case:} since $\bm{\lambda}$ is initialized as $\mathbf{0}$, $\bar{\delta}_t = \max(\bar{\lambda}_1, \bar{\lambda}_2)$ holds for $t =1,2$.

\textit{Inductive step:} \\
Assume
$ \tau$ is the last time 
$\lambda_{1,\tau+1}=0$ or $\lambda_{2,\tau+1}=0$ holds, meaning $\lambda_{1,\tilde{t}}>0$ and $\lambda_{2,\tilde{t}}>0$ for all $\tau<\tilde{t}<t$. Hence, $\bar{\lambda}_{1,\tau} = \max_{t \in \{ 1,\cdots,\tau \}} \lambda_{1,t+1}$ and $\bar{\lambda}_{2,\tau} = \max_{t \in \{ 1,\cdots,\tau \}} \lambda_{2,t+1}$ which are the maximum value of $\lambda_{1,t}$ and $\lambda_{2,t}$ at time $\tau$ and before time $\tau$. 
We also have $\bar{\delta}_\tau = \max_{t \in \{ 1,\cdots,\tau \}} |\lambda_{1,t+1} - \lambda_{2,t+1}| = \max(\bar{\lambda}_{1,\tau}, \bar{\lambda}_{2,\tau})$.
{Below, we will discuss 3 different cases: $\lambda_{2,t+1}=0$, $\lambda_{1,t+1}=0$ and $\lambda_{1,t+1}, \lambda_{2,t+1} >0$.} %\shen{what are the different cases you discussed below?}

Case 1: %For $t$ that 
$\lambda_{1,t+1} >0$ and $\lambda_{2,t+1} >0$, then the maximum operations in \eqref{1_update} and \eqref{2_update} would not be activated.
So $\lambda_{1,t+1}  = \lambda_{1,\tau+1} + \sum_{\tilde{t}=\tau+1}^t \mu g_{\tilde{t}}$ and $\lambda_{2,t+1}  = \lambda_{2,\tau+1} - \sum_{\tilde{t}=\tau+1}^t \mu g_{\tilde{t}}$.

If  $\lambda_{2,\tau+1}=0$, then $\sum_{\tilde{t}=\tau+1}^t \mu g_{\tilde{t}} < 0$ which leads to
%$\lambda_{1,t+1}\leq \lambda_{1,\tau} \leq \bar{\lambda}_{1,\tau} \leq \bar{\lambda}_{d,\tau}$, $\lambda_{2,t+1}\leq \lambda_{1,\tau} \leq \bar{\lambda}_{1,\tau} \leq \bar{\lambda}_{d,\tau}$, and $|\lambda_{1,t+1}-\lambda_{2,t+1}|\leq \lambda_{1,\tau} \leq \bar{\lambda}_{1,\tau} \leq \bar{\lambda}_{d,\tau}$;
\begin{align*}
    &\lambda_{1,t+1}\leq \lambda_{1,\tau+1}, \\%\leq \bar{\lambda}_{1,\tau} \leq \bar{\lambda}_{d,\tau} \\
    &\lambda_{2,t+1}\leq \lambda_{1,\tau+1}. %\leq \bar{\lambda}_{1,\tau} \leq \bar{\lambda}_{d,\tau} \\
    %&|\lambda_{1,t+1}-\lambda_{2,t+1}|\leq \lambda_{1,\tau} \leq \bar{\lambda}_{1,\tau} \leq \bar{\lambda}_{d,\tau}.
\end{align*}
Since $\lambda_{1,\tau+1} \leq \bar{\lambda}_{1,\tau}$ and $\bar{\lambda}_{1,\tau} \leq \bar{\delta}_\tau$, we can have 
\begin{align*}
    \lambda_{1,t+1} \leq \bar{\delta}_\tau,  \lambda_{2,t+1}\leq \bar{\delta}_\tau.
\end{align*}
Due to their nonegativity, 
\begin{align*}
    |\lambda_{1,t+1}-\lambda_{2,t+1}|\leq \bar{\delta}_\tau.
\end{align*}

Similarly, if $\lambda_{1,\tau+1}=0$, then $\sum_{t=\tau+1}^t \mu g_t > 0$ which leads to %$\lambda_{1,t+1}\leq \lambda_{2,\tau} \leq \bar{\lambda}_{2,\tau} \leq \bar{\lambda}_{d,\tau}$, $\lambda_{2,t+1}\leq \lambda_{2,\tau} \leq \bar{\lambda}_{2,\tau} \leq \bar{\lambda}_{d,\tau}$, and $|\lambda_{1,t+1}-\lambda_{2,t+1}|\leq \lambda_{2,\tau} \leq \bar{\lambda}_{2,\tau} \leq \bar{\lambda}_{d,\tau}$.
\begin{align*}
    & \lambda_{1,t+1}\leq  \lambda_{2,\tau+1} \\% \leq \bar{\lambda}_{2,\tau} \leq \bar{\lambda}_{d,\tau}\\
    & \lambda_{2,t+1}\leq \lambda_{2,\tau+1} % \leq \bar{\lambda}_{2,\tau} \leq \bar{\lambda}_{d,\tau}\\
    %& |\lambda_{1,t+1}-\lambda_{2,t+1}|\leq \lambda_{2,\tau} \leq \bar{\lambda}_{2,\tau} \leq \bar{\lambda}_{d,\tau}
\end{align*}
which lead to $\lambda_{1,t+1} \leq \bar{\delta}_\tau,  \lambda_{2,t+1}\leq \bar{\delta}_\tau$. $|\lambda_{1,t+1}-\lambda_{2,t+1}|\leq \bar{\delta}_\tau$ also holds due to the non-negativity.

So $\bar{\delta}_t = \bar{\delta}_\tau= \max(\bar{\lambda}_{1,t}, \bar{\lambda}_{2,t})$ holds for $t$ while $\lambda_{1,t+1} >0$ and $\lambda_{2,t+1} >0$. Since $\lambda_{1,\tilde{t}+1} >0$ and $\lambda_{2,\tilde{t}+1} >0$ for all $\tau<\tilde{t}<t$, $\bar{\delta}_{\tilde{t}} = \bar{\delta}_\tau$ also holds.

%Case2: %For $t$ that 
%$\lambda_{1,t} =0$ or $\lambda_{2,t} =0$
Case 2: $\lambda_{1,t+1} =0$, % $\bar{\lambda}_{1,t}=\bar{\lambda}_{1,\tau}$, $\bar{\lambda}_{2,t} = \max(\bar{\lambda}_{2,\tau},\lambda_{2,t+1})$. $\bar{\lambda}_{d,\tau} = \max(|\lambda_{1,t+1} - \lambda_{2,t+1}| ,\bar{\lambda}_{d,\tau}) = \max(\lambda_{2,t+1} ,\bar{\lambda}_{d,\tau}) = \max(\lambda_{2,t+1}, \bar{\lambda}_{2,\tau}, \bar{\lambda}_{1,\tau}) = \max(\bar{\lambda}_{1,t}, \bar{\lambda}_{2,t})$.\\
then $\lambda_{1,t+1} \leq \bar{\lambda}_{1,\tau+1}$ due to the non-negativity of $\lambda_1$. So we can have $\bar{\lambda}_{1,t}=\bar{\lambda}_{1,\tau}$
Based on the definition of $\bar{\lambda}_{2,t}$, we can have
\begin{align*}
    %& \bar{\lambda}_{1,t}=\bar{\lambda}_{1,\tau}, ~~
    \bar{\lambda}_{2,t} = \max(\bar{\lambda}_{2,\tau},\lambda_{2,t+1}) \\
\end{align*}
Then $\bar{\delta}_{t}$ can be written as
\begin{align*}
    \bar{\delta}_{t} =& \max(|\lambda_{1,t+1} - \lambda_{2,t+1}| ,\bar{\delta}_\tau) \\  
    \stackrel{(a)}{=}& \max(\lambda_{2,t+1} ,\bar{\delta}_\tau)  \\ 
    \stackrel{(b)}{=}& \max(\lambda_{2,t+1}, \bar{\lambda}_{2,\tau}, \bar{\lambda}_{1,\tau}) \\
    \stackrel{(c)}{=}& \max(\bar{\lambda}_{1,t}, \bar{\lambda}_{2,t})
\end{align*}
where $(a)$ holds due to $\lambda_{1,t+1} =0$, $(b)$ holds due to $\bar{\delta}_\tau =\max (\bar{\lambda}_{2,\tau}, \bar{\lambda}_{1,\tau})$.
$(c)$ holds since $\bar{\lambda}_{2,t} = \max(\bar{\lambda}_{2,\tau},\lambda_{2,t+1}), \bar{\lambda}_{1,t}=\bar{\lambda}_{1,\tau}$.

So $\bar{\delta}_t = \max(\bar{\lambda}_{1,t}, \bar{\lambda}_{2,t})$ holds while $\lambda_{1,t+1} =0$.

Case 3: $\lambda_{2,t+1} =0$, % $\bar{\lambda}_{2,t}=\bar{\lambda}_{2,\tau}$,$\bar{\lambda}_{1,t} = \max(\bar{\lambda}_{1,\tau},\lambda_{1,t+1})$. $\bar{\lambda}_{d,\tau} = \max(|\lambda_{1,t+1} - \lambda_{2,t+1}| ,\bar{\lambda}_{d,\tau}) = \max(\lambda_{1,t+1} ,\bar{\lambda}_{d,\tau}) = \max(\lambda_{1,t+1}, \bar{\lambda}_{1,\tau}, \bar{\lambda}_{2,\tau}) = \max(\bar{\lambda}_{1,t}, \bar{\lambda}_{2,t})$.
similar to case 2, we can have 
\begin{align*}
    \bar{\lambda}_{2,t}=&\bar{\lambda}_{2,\tau}\\
    \bar{\lambda}_{1,t} =& \max(\bar{\lambda}_{1,\tau},\lambda_{1,t+1}) \\
    \bar{\delta}_{t} =& \max(|\lambda_{1,t+1} - \lambda_{2,t+1}| ,\bar{\delta}_\tau) \\
    =& \max(\lambda_{1,t+1} ,\bar{\delta}_\tau) \\
    =& \max(\lambda_{1,t+1}, \bar{\lambda}_{1,\tau}, \bar{\lambda}_{2,\tau}) \\
    =& \max(\bar{\lambda}_{1,t}, \bar{\lambda}_{2,t})
\end{align*}
So $\bar{\delta}_t = \max(\bar{\lambda}_{1,t}, \bar{\lambda}_{2,t})$ holds while $\lambda_{1,t+1} =0$.

Combining the three cases, $\bar{\delta}_t = \max(\bar{\lambda}_{1,t}, \bar{\lambda}_{2,t})$ also holds for $t$.

\textit{Conclusion:} since both the base case and the inductive step holds, the statement $\bar{\lambda}_d = \max(\bar{\lambda}_1, \bar{\lambda}_2)$ holds across the whole time horizon $\{1,\cdots,T\}$. %for every time $t$.

Proof of \textbf{Lemma 2} is completed.
\end{appendices}

\end{document}